%% file: main.tex
\input{_constants}
\cameraready

\pdfoutput=1
\documentclass[10pt,twocolumn,letterpaper]{article}
\input{cvpr_header}
\usepackage{url}
\usepackage{arydshln}
\usepackage{footnote}
\usepackage{multirow}
\usepackage{pifont}

\usepackage[noEnd=false]{algpseudocodex}
\usepackage{algorithm}

\renewcommand{\thefootnote}{\fnsymbol{footnote}}

\usepackage{stmaryrd}

\title{FlowMimic: Mask-free Visual Editing and Generation with Pixel-pair Warped Flow Field for Online Video Editing Data Generation and Modality Mimicry}

\author{
    \normalsize Dingyun Zhang$^{1, \dag}$ \quad Lixue Gong$^{1}$ \quad Wei Liu$^{1, \S, \dag}$\\
    \small $^{1}$ByteDance \\
}
\begin{document}

\twocolumn[{
    \renewcommand\twocolumn[1][]{#1}
    \maketitle
    \vspace*{-2.4em}
    \begin{center}
        \captionsetup{type=figure}
        \includegraphics[width=0.912\linewidth]{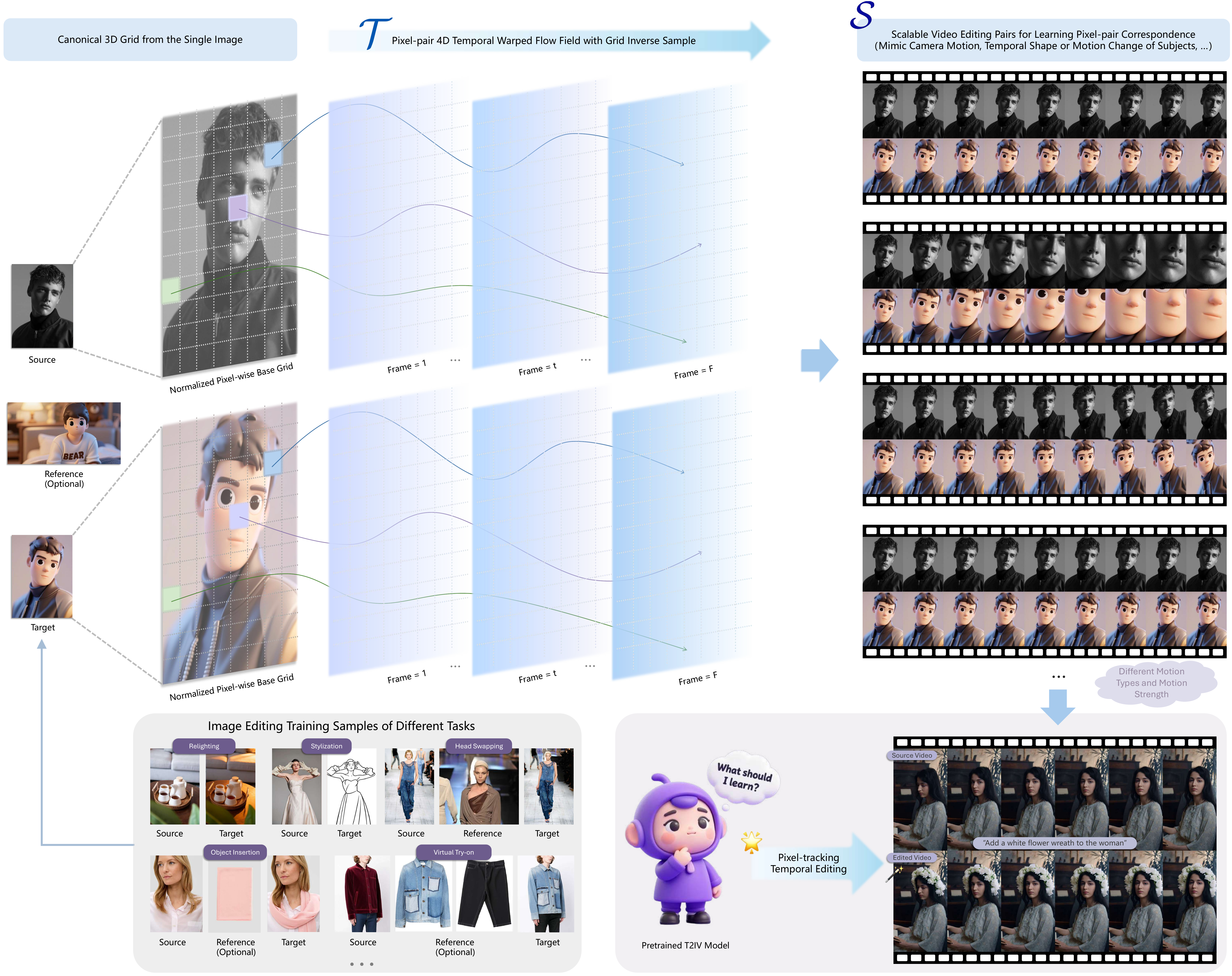} \vspace{-0.8em}
        \captionof{figure}{
            \textbf{Overview of the video editing pair generation paradigm with pixel-pair temporal warped flow field.}
            For any given image editing training sample of a task, we can obtain video editing samples in real time via this field that enable the model to learn the video editing capability for this task. 
            Such video editing samples align with the model's cognitive patterns, eliminating the need to adapt them to conform to human perception.
            In this paper, all video editing tasks of interest are trained jointly only with our online generated video samples; the bottom-right corner illustrates an editing result for object insertion. Please refer to \cref{subsec:qualitative_video_editing_results} for more results.
        }
        \label{fig:Pixel-pair Temporal Warped Flow Field}
    \end{center}
}]

\footnotetext{$^\S$Project lead. $^\dag$Corresponding authors.}
\renewcommand{\thefootnote}{\arabic{footnote}} 

\input{00_abstract}
\input{01_intro}
\input{02_related}
\input{03_method}
\input{04_exp}
\input{10_conclusion}

{
    \small
    \bibliographystyle{unsrt} 
    \bibliography{11_references}
}

\clearpage \appendix \input{12_appendix}

\end{document}

%% file: _constants.tex
\def\paperID{0000}
\def\confName{CVPR}
\def\confYear{2024}

\newif\ifreview 
\newif\ifarxiv 
\newif\ifcamera \newcommand{\cameraready}{\cameratrue}
\newif\ifrebuttal 

%% file: cvpr_header.tex
\ifreview \usepackage[review]{cvpr} \fi
\ifarxiv \usepackage[pagenumbers]{cvpr} \fi
\ifrebuttal \usepackage[rebuttal]{cvpr} \fi
\ifcamera \usepackage{cvpr} \fi

\input{_macros}  

\usepackage{xr-hyper}

\makeatletter
\newcommand*{\addFileDependency}[1]{
  \typeout{(#1)}
  \@addtofilelist{#1}
  \IfFileExists{#1}{}{\typeout{No file #1.}}
}

\makeatother

\definecolor{cvprblue}{rgb}{0.21,0.49,0.74}
\usepackage[pagebackref,breaklinks,colorlinks,citecolor=cvprblue]{hyperref}
\usepackage[capitalize]{cleveref}
\crefname{section}{Sec.}{Secs.}
\crefname{table}{Table}{Tables}
\crefname{figure}{Fig.}{Figs.}

%% file: _macros.tex
\usepackage{graphicx}	
\usepackage{amsmath}	
\usepackage{amssymb}	
\usepackage{booktabs}
\usepackage{times}
\usepackage{microtype}
\usepackage{epsfig}
\usepackage[table,xcdraw,dvipsnames]{xcolor}
\usepackage{caption}
\usepackage{float}
\usepackage{placeins}
\usepackage{color, colortbl}
\usepackage{stfloats}
\usepackage{enumitem}
\usepackage{tabularx}
\usepackage{xstring}
\usepackage{multirow}
\usepackage{xspace}
\usepackage{url}
\usepackage{subcaption}
\usepackage{xcolor}
\usepackage[hang,flushmargin]{footmisc}

\ifcamera \usepackage[accsupp]{axessibility} \fi

% ...

\ifarxiv  \fi

\newcommand{\R}[1]{{%
    \textbf{%
        \ifstrequal{#1}{1}{\textcolor{red}{R#1}}{%
        \ifstrequal{#1}{2}{\textcolor{blue}{R#1}}{%
        \ifstrequal{#1}{3}{\textcolor{magenta}{R#1}}{%
        \ifstrequal{#1}{4}{\textcolor{teal}{R#1}}{%
                           \textcolor{cyan}{R#1}%
        }}}}%
    }%
}}

%% file: 00_abstract.tex
\begin{abstract}
In line with the prevailing direction of vision research, we explore the integration of both generation and editing capabilities for video and image modalities within a single model. Current approaches to collecting video editing data typically depend on labour-intensive, time-consuming curated procedures—involving object mask annotation, the use of error-introducing pair synthesis via I2V model and ControlNet-like guidance, and VLM-based quality filtering or refinement—and demonstrate limited task scalability. As a result, the diversity of editing tasks remains substantially narrower than that available for image editing models. We develop a pixel-pair temporal warped flow field that can directly generate corresponding video editing samples in real time from image editing samples, and we demonstrate across multiple levels of video editing tasks that a model can learn video editing using only such data.
We regard the image modality as a particular form of the video modality. Accordingly, we design a modality mimic generation loss and a modality mimic editing loss to relatively align the capabilities—and thereby the output distributions—of the two modalities through mutual imitation.
Moreover, language-based visual editing entails the comprehension of the editing instruction and the reference visual content, the localization of the region corresponding to that instruction within the reference visual contents, and the modification of that region alone. Existing approaches predominantly rely on external aids, such as fine-tuning an additional MLLM or explicitly supplying a mask sequence as auxiliary input during inference. In contrast, we aspire for the model to internalize this capability. To that end, we introduce sense-related tasks—for instance, referring expression segmentation—along with corresponding editing-region-aware latent-level loss and attention-level loss.
Due to file size constraints, the figures in the arXiv file have been heavily lossy-compressed. Please visit the uncompressed file at: \url{https://huggingface.co/datasets/FlowMimic/Uncompressed/blob/main/main.pdf}.
\end{abstract}

%% file: 01_intro.tex
\section{Introduction}
\label{sec:introduction}
\begin{quote}
    \textit{``Why didn't they ask Evans?''}

    \hfill -- Agatha Christie, \textit{Marple}
    \vspace{-3mm}
\end{quote}
The construction of video editing datasets~\cite{cheng2023consistent,hertz2022prompt,zi2025se,bai2025scaling,wu2025insvie,he2025openve} currently relies on labour-intensive pipelines such as annotating object masks via multimodal large language models (MLLMs)~\cite{yin2024survey} including vision-language models (VLM)~\cite{zhang2024vision,li2025survey} like SAM~\cite{ravisam,carion2025sam}, DINO~\cite{liu2024}, and Qwen-VL~\cite{wang2024qwen2,bai2025qwen2,bai2025qwen3} followed by inpainting, carefully curating video clips to form editing pairs, or generating video sequences from image pairs using image-to-video (I2V) models guided with control signals~\cite{zhang2023adding} in an error-introducing manner.
These paradigms are typically task-dependent, and the generated video editing samples need to be filtered and refined by the manual, VLM-driven, or RL-driven~\cite{yu2025reward} procedures.
This raises two critical questions: (1) Can we acquire video editing representations without such elaborate and poorly scalable processes, and (2) must the video editing training samples consist of human-intelligible temporal content for models to learn how to perform temporal editing according to the text instructions? 
Much like the principle in Agatha Christie's \textit{Why Didn't They Ask Evans?}\footnote{\url{https://www.imdb.com/title/tt1276406/}}, we argue for a simpler, more model-centric direct approach: a scalable paradigm that leverages only existing diverse image editing samples to real-time generate video editing samples suited to model learning. 

The central issue is how models develop temporal editing competence. 
We posit that models do not require video pairs that conform to real-world coherence; rather, temporal consistency of pixel-level editing correspondences—anchored on the first frames of the input and edited video—suffices. Even if objects deform unnaturally in subsequent frames, models can still learn frame-consistent pixel-aligned editing relationships. 
Thus, at inference, given a real source video and the optional reference images, the model yields a naturally coherent editing effect. 

As illustrated in \cref{fig:Pixel-pair Temporal Warped Flow Field}, we take the multi-reference stylization editing task as an example, where the style of a reference image is transferred to the content of the source image. By viewing both the input and target images as sets of points (pixels), our objective is to maintain the pixel-wise editing correspondences consistently across time.
Our naive motivation is to let each spatially corresponding pixel-pair ``travel'' or ``random walk'' freely and synchronizedly through time while maintaining its spatial correspondence in every arriving frame plane. 
Once the model learns this temporal relationship, performing editing on human-intelligible videos during inference becomes a natural outcome.
We construct a 3D grid for both the input and target image, treating each as a canonical image, and build a pixel-pair 4D temporal warped flow field via grid inverse sampling. 
Our underlying intuition is to expose the model to clusters of corresponding point flows, thereby enabling the temporal attention component to perceive how the editing effect from the first frame pair—the image modality—is maintained throughout the video sequence. 
Specific details are deferred to the methodology section in \cref{subsec:Pixel-pair Temporal Warped Flow Field}.
Experiments in \cref{sec:experiments} confirm this approach, showing generalized editing performance across diverse mixed editing tasks based on a pretrained Text-to-Image-Video (T2IV) model~\cite{wan2025wan}.
Our experiments demonstrate that, with a low learning rate warmed up from $5 \times 10^{-6}$ to $1 \times 10^{-5}$, the model begins to exhibit initial responsiveness to these video editing tasks within only a few thousand training steps.

Current methods for joint video and image modelling in generation and editing often rely on a two-stage training paradigm: training first on the image modality and then incorporating video data~\cite{wan2025wan}, or performing post-training on existing T2IV models with additional video and image editing data~\cite{mou2025instructx,chen2026vino}. 
The key distinction between processing these two modalities is the presence of temporal attention. 
An interesting observation is that the synthetic-aware appearance of images generated by some T2IV models is often more pronounced than that of the first frame in a generated video, as exemplified in \cref{fig:flowmimic_fig_2}. 
In other words, the first frame of a multi-frame generated video often exhibits better photorealism compared to a single-frame generated image. 
In fact, we conceptualize an image as a special single-frame video modality. Consequently, we intend to align the output distributions of the video and image modalities for both generation and editing tasks. That is, to make the distribution of Text-to-Image (T2I) closely match that in Text-to-Video (T2V), and similarly, the distribution of Image-to-Image (I2I) closely match that in Video-to-Video (V2V). In other words, we aim to treat the image and video modalities as a unified modality in terms of output distribution. To achieve this alignment, we design two loss functions: the first-frame modality mimic generation loss and the first-frame modality mimic editing loss. The former encourages the T2I generation ability of the model to mimic its photorealistic T2V counterpart, while the latter encourages the V2V editing response of the model to mimic its I2I counterpart, which is an easier task and therefore converges faster, as illustrated in \cref{fig:flowmimic_fig_4}.

Contemporary visual editing approaches frequently incorporate mechanisms to enhance visual-cognitive understanding. 
Some methods~\cite{bian2025videopainter,ye2025unic,gao2025anyportal} provide explicit masks during inference to distinguish editing regions, while some propagation-based methods~\cite{liu2025generative} introduce an auxiliary editing-region-mask predicting module during training. 
Many mainstream works~\cite{mou2025instructx,univideo,chen2026vino} integrate MLLMs with connectors, either frozen or fine-tuned, to refine visual and textual tokens, leveraging MLLMs capabilities to localize the editing region. 
In this work, we aim to enhance the model's capabilities in natural language understanding, visual comprehension, cross-modality matching between textual instructions and visual regions, and editing-region-aware editing. This enables the model to locate the intended editing region within a reference video or image based on an editing instruction and to modify only that specific area. In contrast to approaches that rely on an auxiliary editing-region mask sequence as input during inference, or those that leverage the priors of MLLMs with the trade-off of including large model parameters, we aim for the model to internalize these abilities.
To this end, we introduce sense-related tasks, including the referring expression segmentation task and the sense select task, and corresponding editing-region-aware latent-level and attention-level losses. 

To validate the feasibility of our data paradigm and algorithms, we employ a lightweight model and an online generation manner for video editing samples.
Specifically, we perform post-training on a pretrained T2IV model Wan2.1-T2V-1.3B~\cite{wan2025wan} and adapt the model architecture for editing tasks, including introducing a reference-inject self-attention mask.
To validate our data paradigm, we employ only image editing samples and the video editing samples generated online from image editing samples via the pixel-pair temporal warped flow field during training. In other words, we use no other specially curated video editing data.

Our model, FlowMimic, demonstrates effective editing across multiple categories after mixed editing task training, which is illustrated in \cref{subsec:qualitative_video_editing_results}.
In practice, based on these experiments, whenever a loose layout correspondence exists between the source and target editing images for a specific video editing task, the video editing data from the pixel-wise temporal warped flow field enables the model to learn the associated video editing ability. 
Our experiments in \cref{sec:experiments} also demonstrate that, compared with the pretrained T2IV model, FlowMimic yields more camera-authentic T2I results in the image modality, and the output distributions of its video editing and image editing results are relatively aligned.
Analysis of the cross-attention maps in \cref{fig:flowmimic_fig_27} shows that the model can reliably localize a textual phrase to its corresponding visual region in the reference input. 
Its performance across diverse editing tasks further confirms that it could identify and edit only the intended editing region. 
That is to say, the model has assimilated the essential capacity for editing-region-aware editing to some extent, which is underpinned by the accurate comprehension of the editing instruction and the reference visual input, precise localization of the mentioned subject within the visual input, and the disciplined modification of solely that region.
Besides, as shown in \cref{fig:flowmimic_fig_20}, with referring expression segmentation image editing samples and the video counterparts online generated, FlowMimic achieves a SAM3-like capability of segmenting and temporally tracking the language-described objects in videos, such as ``a girl in yellow'', illustrating the advantage of our pixel-aligned data paradigm for high-level video tasks like referring expression segmentation and semantic segmentation.
The contributions can be summarized as follows:
\begin{itemize}
    \item We abstract the essence of video editing learning as the temporal maintaining of pixel-level editing effect, propose and model a temporal point-flow warped field to scalably generate video editing data in real time from image editing samples alone, demonstrate the effectiveness of this pixel-aligned paradigm towards learning video editing across diverse-level editing tasks, and reveal its promising extensibility to high-level video tasks such as segmentation-related tasks, low-level video tasks such as video deblurring, and controllable video generation such as condition maps to video.
    \item We posit that a model trained concurrently on both image and video modalities for generation and editing tasks should yield relatively aligned output distributions across the two modalities, as the image modality is essentially a special case of a video modality. Accordingly, we design a modality mimic generation loss and a modality mimic editing loss.
    \item The capacity for language-based visual editing fundamentally requires a model to comprehend the instruction and the reference visual inputs, localize the corresponding region in the reference visual input, and modify it exclusively. Rather than relying on fine-tuning additional modules such as VLMs with connector adaptation or assisted mask sequence inputs, we endow the model with this internalized ability through introducing sense-related tasks and corresponding editing-region-aware latent-level and attention-level losses.
\end{itemize}

%% file: 02_related.tex
\section{Related Work}
\subsection{Extensive Image Editing Datasets}
In recent years, the scale of publicly available image editing datasets has continued to expand, with different datasets excelling in distinct aspects.
We briefly introduce several examples below.
MagicBrush~\cite{zhang2023magicbrush} is a manually annotated dataset for instruction-based real image editing. 
It comprises over 10K triplets of source images, editing instructions, and target images.
Pico-Banana-400K~\cite{qian2025pico} is a 400K image dataset for instruction-based image editing.
It is constructed by leveraging Nano Banana\footnote{\url{https://gemini.google/overview/image-generation/}} to generate editing pairs from real photographs. 
It maintains content preservation and instruction faithfulness through MLLM-based quality scoring and curation.
Moreover, GPT-IMAGE-EDIT-1.5M~\cite{wang2025gpt} contains more than 1.5M high-quality editing triplets. 
It leverages GPT-4o\footnote{\url{https://openai.com/index/hello-gpt-4o/}} to unify and refine three existing image editing datasets.
ImgEdit~\cite{ye2025imgedit} comprises 1M curated editing pairs, which contain novel and complex editing tasks.

\subsection{Video Editing Data Curation Paradigms}
In recent years, to advance the field of video editing, researchers have also devoted considerable effort to creating video editing training data.
InsV2V~\cite{cheng2023consistent} constructs a synthetic video editing dataset to overcome the scarcity of naturally paired videos. 
The method adapts the Prompt-to-Prompt~\cite{hertz2022prompt} approach on a video diffusion model, while utilising paired text prompts from both image caption and video caption sources. 
The generated 16-frame video samples are then filtered using CLIP-based~\cite{radford2021learning} metrics.
Señorita-2M~\cite{zi2025se} collects videos online. 
Its dataset is created by orchestrating a pipeline of specialized editing experts—including local and global stylizers, a remover, and an inpainter—applied to the sourced videos. 
The resulting video pairs are then filtered using CLIP-based metrics and trained classifiers to ensure quality.
Furthermore, Ditto~\cite{bai2025scaling} leverages edited key-frames and depth maps to generate edited videos. 
Besides, InsViE-1M~\cite{wu2025insvie} leverages a two-stage editing-filtering pipeline for generating triplets from real-world videos. 
In the first stage, it edits the first frame of each video and screens the best edited sample. 
In the second stage, it edits videos via noise inversion and the I2V model, and filters them with VLMs.
OpenVE-3M~\cite{he2025openve} is constructed via a multi-stage pipeline that relies on orchestrating VLMs (e.g., GPT-4o) and I2V models (e.g., Wan2.1-I2V~\cite{wan2025wan}) to synthesize editing pairs. 
This process involves intricate, category-specific generation workflows followed by VLM-based filtering. 
Using rendered data is also a viable approach for specific sub-tasks. 
During the internship in early 2024, Dingyun Zhang proposed constructing camera-controllable body animation video training data using a 3D game engine and Mixamo\footnote{\url{https://www.mixamo.com/}}. 
This approach was later adopted in constructing training datasets for video relighting~\cite{bian2025relightmaster} or recamera~\cite{bai2025recammaster} tasks.

\begin{figure*}[htpb]
    \centering
    \includegraphics[width=1.0\linewidth, trim=0pt 0pt 0pt 0pt, clip]{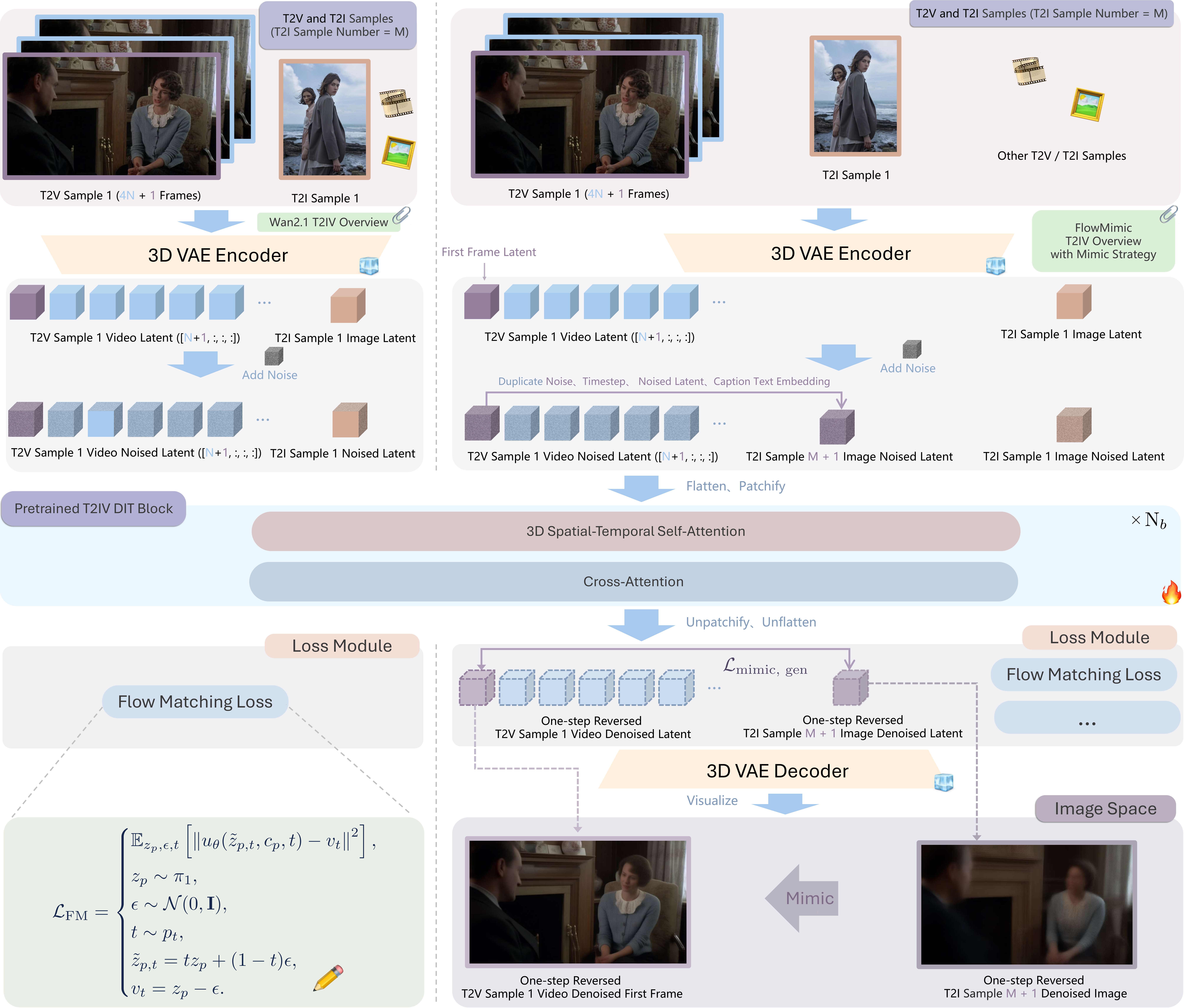} 
    \captionsetup{font=scriptsize}
    \caption{
        \textbf{Overview of the pretrained T2IV model~\cite{wan2025wan} and our modality mimic generation loss.}
        The area to the left of the grey dashed line provides a schematic overview of our adopted pretrained T2IV model, while the area to the right introduces the modality mimic generation loss used in our generation training step. The DiT blocks are shared components across both sides of the dashed line. 
        In the schematic illustrating the modality mimic generation loss, for better clarity, we present the newly created T2I sample as the $(M+1)$-th sample in the current training batch. 
        In our implementation, this new sample replaces the first T2I sample in the current training batch.
        The flame symbol denotes that the parameters in those parts are optimized during training, whereas the ice symbol denotes that the parameters in those parts remain frozen. Please refer to \cref{subsec:preliminary} and \cref{subsec:First Frame Modality Mimic Loss} for details, respectively.
    }
    \label{fig:flowmimic_fig_32}
\end{figure*}

\subsection{Video Editing Paradigms}
Recent video editing paradigms can be broadly classified into training-free and training-based approaches.
For training-free approaches, a prerequisite typically involves mapping the source video to the initial noise of a pretrained video model through an inversion technique. The success of the subsequent editing process critically depends on the quality of this initial noise latent, thereby imposing stringent requirements on the inversion method employed.
AnyV2V~\cite{ku2024anyv2v} applies the existing image editing method on the first source frame, inverts the source video into the initial noise of an I2V model, and injects the source video branch's spatial convolution, spatial attention, and temporal attention features to the editing video branch during DDIM~\cite{song2020denoising} sampling.
Videoshop~\cite{fan2024videoshop} introduces inversion with noise extrapolation to improve the reconstruction quality of the source video from the initial noise.
Furthermore, ContextFlow~\cite{chen2026contextflow} proposes rectified flow~\cite{lipman2022flow} inversion towards object-related video editing.
It employs off-the-shelf image editing tools to edit the first source frame, and a dual-path sampling process to decouple reconstruction and editing.

Training-based video editing paradigms are pursued for upgrading editing capabilities by either fine-tuning the parameters of a video model or incorporating learnable LoRA~\cite{hu2022lora} parameters.
I2VEdit~\cite{ouyang2024i2vedit} and GenProp~\cite{liu2025generative} fine-tune the I2V models by adding LoRAs, and propagate the off-the-shelf first frame editing result to the rest of the video like AnyV2V during inference.
VACE~\cite{jiang2025vace} trains a plugged-and-play context adapter to inject task concepts into the pretrained T2V model.
VideoPainter~\cite{bian2025videopainter} introduces an ID-Adapter to enhance the temporal identity consistency. 
Moreover, UNIC~\cite{ye2025unic} fine-tunes a pretrained DiT-based~\cite{peebles2023scalable} video model and employs in-context learning.
It introduces task-dependent RoPE~\cite{su2024roformer} and condition bias to differentiate the editing tasks.
Specifically, GenProp trains a mask predictor to identify the editing region. VACE, VideoPainter, and UNIC leverage the editing-region mask sequence provided during inference to locate the editing region.

\subsection{Task-specific Editing Models}
Many methods focus on specific editing tasks, and we briefly introduce some of them.
Light-A-Video~\cite{zhou2025light} is a  training-free video relighting method with a consistent light attention module and a progressive light fusion strategy.
ANYPORTAL~\cite{gao2025anyportal} is a method for video relighting and background replacement.
It is divided into three stages: background generation, light harmonization, and consistency enhancement. 
VFace~\cite{baliah2026vface} proposes a plug-and-play method for video face swapping.
It introduces target structure guidance and frequency spectrum attention interpolation for identity preservation.
KeyTailor~\cite{he2025devil} proposes a keyframe-driven details injection strategy for video virtual try-on.
It also introduces a high-definition video try-on dataset across various garment styles.
Some approaches employ additional conditional signals to enable controllable editing, typically based on ControlNet-like~\cite{zhang2023adding} or adapter-like modules. 
For example, portrait animation method Rig-Your-Portrait~\cite{zhang2025rigyourportrait} firstly combines explicit 3D guidance (e.g., personalized mesh and normal maps) and an identity-expression-illumination decoupling strategy with the video generation model, which enables identity-preserving transferring of facial expressions, head poses, or illumination from a driving image or video to a reference portrait in a decoupling manner.

\subsection{Unified IV Editing and Generation Methods}
For training-based methods, an active line of recent research focuses on developing unified models for multiple tasks or for multiple modalities—in particular, image and video. 
EditVerse~\cite{ju2025editverse} adopts a full self-attention transformer architecture, in which all text and visual inputs are tokenized and concatenated into a single interleaved sequence. 
The method also introduces a video editing dataset constructed via existing video editing models and VLMs.
InstructX~\cite{mou2025instructx} provides a unified framework for image and video editing. 
It uses a VLM to encode the editing instruction and the source visual input; the resulting queries are fed through a connector and then replace the text embeddings in the DiT. 
UniVideo~\cite{univideo} aims to unify video understanding, generation, and editing. 
It likewise relies on a VLM for visual understanding, coupled with a multimodal DiT (MMDiT) for generation. 
In the same vein, VINO~\cite{chen2026vino} performs both image and video generation and editing within one framework, also employing a VLM with an MMDiT where multimodal inputs are encoded as interleaved conditioning tokens and fed together with source visual tokens to guide the diffusion process. 
Moreover, Omni-video 2~\cite{yang2026omni} focuses on video generation and editing, and similarly depends on a VLM to interpret the instruction and source video into a target caption.

%% file: 03_method.tex
\section{Methodology}
\label{sec:method}
FlowMimic builds upon the publicly available Wan2.1-T2V-1.3B model~\cite{wan2025wan}, a pretrained T2IV model with attractive foundational video generation capabilities. 
We employ this relatively light model to conduct our experiments. 
This section details our method.
We begin with a brief overview of Wan2.1's core architecture and its flow matching loss~\cite{lipman2022flow} in \cref{subsec:preliminary}. 
We then detail our data paradigm centered on a pixel-pair temporal warped flow field in \cref{subsec:Pixel-pair Temporal Warped Flow Field}. 
This is followed by the introduction of the first frame modality mimic losses in \cref{subsec:First Frame Modality Mimic Loss}, which treats video and image modalities as a unified form and fosters mutual imitation via a modality mimic generation loss and a modality mimic editing loss. 
To enhance the model's cognitive understanding of cross-modality correspondence between the editing region specified in the instruction and reference visual regions, and to improve its editing-region-aware capability, we introduce a referring expression segmentation~\cite{ding2025multimodal} task with corresponding latent-level editing-region-aware and attention-level losses in \cref{subsec:Sense Losses}.
In \cref{subsec:Model Architecture Adaptation to Editing Task}, we present the adjustments made to the model architecture for the editing tasks.
Moreover, we present additional details pertaining to the editing tasks in \cref{subsec:Miscellaneous editing-task specifics}.
Finally, in \cref{subsec:From stylized T2I to stylized T2V}, we briefly introduce the common strategy of employing stylized T2I training data to enable the model to learn stylized T2V generation.

\subsection{Preliminary of Wan2.1 and Flow Matching Loss}
\label{subsec:preliminary}
As illustrated on the left of \cref{fig:flowmimic_fig_32}, Wan2.1 is a T2IV model built on Diffusion Transformers (DiT)~\cite{peebles2023scalable}. 
Taking a video input as an example, a video sample $S_{V} \in \mathbb{R}^{(4N + 1) \times H \times W \times C}$ is encoded by a 3D spatio-temporal VAE into a latent representation $z \in \mathbb{R}^{(N + 1) \times \lfloor \frac{H}{8} \rfloor \times \lfloor \frac{W}{8} \rfloor \times 16}$, which is then flattened and patchified into visual tokens $z_p$. 
The first frame is only spatially compressed by the VAE encoder.
During training, given the timestep $t$ sampled from a logit normal distribution $p_t$, the latents are perturbed with Gaussian noise $\epsilon$, yielding the noisy visual tokens $\tilde{z}_{p,t}$, and pass through 3D spatio-temporal full self-attention~\cite{vaswaniattention} layers.
Meanwhile, the text caption $c_p$ is tokenized by the T5~\cite{raffel2020exploring} tokenizer and encoded into text embeddings by the T5 text encoder.
The noisy visual tokens $\tilde{z}_{p,t}$ are subsequently denoised under the guidance of the text embeddings via cross attention.

Following the flow matching~\cite{lipman2022flow} framework, the DiT backbones learn to predict a velocity field $\hat{v}_t$ that progressively guides noisy samples towards the clean data manifold. 
The core objective is to model a vector field that defines straight line trajectories—the shortest and most efficient paths—to transport samples from a simple noise distribution $\mathcal{\pi}_0$ (e.g., $\mathcal{N}(0,\mathbf{I})$) to the target data distribution $\mathcal{\pi}_1$. 
This formulation provides a stable and efficient training target compared to stochastic diffusion paths.
In practice, for a given data sample (i.e., a video latent) $z_p$, a noise latent $\epsilon \sim \mathcal{N}(0,\mathbf{I})$ and a timestep $t \in [0,1]$ are sampled. 
Based on the flow matching assumption, a linear interpolation constructs a straight path between them:
\begin{equation}
 \tilde{z}_{p,t} = (1-t) z_p + t \epsilon .
\label{eq:xt}
\end{equation}
The derivative of this path defines the target velocity:
\begin{equation}
v_t = \frac{d\tilde{z}_{p,t}}{dt} = \epsilon - z_p.
\label{eq:vt}
\end{equation}
The T2IV model $u_\theta$, conditioned on a caption $c_p$, is trained to predict this velocity:
\begin{equation}
\hat{v}_t = u_\theta(\tilde{z}_{p,t}, c_p, t),
\label{eq:u}
\end{equation}
where $\theta$ denotes the model parameters. 
The training objective minimizes the mathematical expectation:
\begin{equation}
 \mathcal{L}_{\text{FM}} = \mathbb{E}_{z_p \sim \pi_1, \epsilon \sim \mathcal{N}(0,\mathbf{I}), t \sim p_t} \left[\bigl\| u_\theta(\tilde{z}_{p,t}, c_p, t) - v_t \bigr\|^2\right].
\label{eq:loss}
\end{equation}
Critically, the target $v_t = \epsilon - z_p$ is constant for any $t$, providing a simple, time-invariant learning signal that encourages better convergence and higher sampling quality. 

\subsection{Pixel-pair Temporal Warped Flow Field}
\label{subsec:Pixel-pair Temporal Warped Flow Field}
As shown in \cref{fig:Pixel-pair Temporal Warped Flow Field}, we introduce a scalable paradigm for generating video editing training samples from a single image editing sample. 
The core idea is that an image editing pair, including the source and target images, can be viewed as two spatially corresponding sets of points (i.e., pixels), while a video editing pair is the ``synchronous walk'' of these two sets of points (i.e., point flows) along the temporal axis. 
Regardless of the time plane, points at the same spatial location maintain editing correspondences.
The learning essence of video editing for a model lies in realizing and maintaining pixel-level editing correspondences across the temporal axis.
To this end, we define a shared, time-varying deformation field—a pixel-pair temporal warped flow field—that applies identical spatial transformations to both the source and target images, thereby producing a temporally coherent video pair in which every frame maintains the same pixel-wise editing relationship.

We construct video editing pairs online during training. 
We first collect image editing pairs for tasks of interest, such as relighting, stylization, head swapping, object insertion, virtual try-on, and so on. 
During training, we randomly sample from these pairs, for instance, selecting a stylization sample with a reference image.

Let \((I_{\text{ref}_1}, I_{\text{ref}_2}, \dots, I_{\text{ref}_n}, I_{\text{tar}})\) denote the normalized image editing sample.
Specifically, \(I_{\text{\text{ref}}_1}\) is the source image, \(\{ I_{\text{ref}_i} \}_{i=2, \dots,n}\) are the optional reference images, \(I_{\text{tar}}\) is the edited image, and $I_{\text{ref}_1}$, $I_{\text{tar}} \in \mathbb{R}^{3 \times H \times W}$. 
In the example of \cref{fig:Pixel-pair Temporal Warped Flow Field}, \(I_{\text{ref}_1}\) is a photo of a young man, \(I_{\text{ref}_2}\) is a style reference image in the style of a 3D animation, and \(I_{\text{tar}}\) is a 3D animated style photo of the young man referenced from \(I_{\text{ref}_1}\). 

We first construct the canonical sampling grids $G^{\text{ref}_1}_{\text{base}}, G^{\text{tar}}_{\text{base}} \in [-1,1]^{H \times W \times 2}$ that represent the normalized coordinate space of the image editing pair \((I_{\text{ref}_1}, I_{\text{tar}})\) respectively. 
As shown in \cref{fig:Pixel-pair Temporal Warped Flow Field}, we take three spatially corresponding pixel-pairs from the canonical base grid of the source image and the canonical base grid of the target image as illustrative examples, namely blue, green, and purple pixel-pairs. 
We aim to ensure that, on each subsequent temporal frame plane, the intersection points of the trajectories of these pixel pairs with that plane remain synchronised and spatially aligned.
To this end, for a given video length $F$, we generate the 4D temporal deformation grids $G^{\text{ref}_1}, G^{\text{tar}} \in [-1,1]^{F \times H \times W \times 2}$ with $G^{\text{ref}_1}_t=G^{\text{ref}_1}[t]$ and $G^{\text{tar}}_t=G^{\text{tar}}[t]$ for $t \in {0,1,\dots,F-1}$ by applying a composition of randomized, time-dependent geometric transformations. 
Crucially, the same deformation sequence is applied to source and target images, guaranteeing that the pixel-wise editing mapping established in the first frame is preserved throughout the sequence.

Given the base grid $G_{\text{base}}$, the specific motion type $\mathcal{M}$, the frame number $t$, the motion strength hyperparameter mapping function $f_{\alpha}(\mathcal{M})$ depending on the motion type, and sample-dependent random seed $S$, the deformation grid at frame number $t$ is defined as:
\begin{equation}
G_t = \mathcal{W}\bigl(G_{\text{base}}, \mathcal{M}, t, f_{\alpha}(\mathcal{M}), S \bigr).
\end{equation}
The deformation grid depends on the random seed $S$ because some parameters undergo sample-wise uniform sampling within prescribed intervals defined by $f_{\alpha}(\mathcal{M})$, thereby enhancing the diversity in the motion strength of the generated videos.
Besides, we ensure that the same random seed $S$ is supplied for both the source and target images of a given sample, thereby guaranteeing that an identical deformation sequence is applied to the pair.
We use the term \textit{forward motion type} to denote constructing a video editing pair starting from the original image pair, and \textit{reverse motion type} to indicate that the generated sequence ends with the original image pair. 
Formally, let \(d(\mathcal{M}) \in \{0,1\}\) be a direction function that indicates whether the motion type is forward (d=0) or reverse (d=1), then:
\begin{equation}
\left\{
 \begin{array}{lr}
 G_0 = G_{\text{base}}, \text{if } d(\mathcal{M}) = 0, & \\
 G_{F-1} = G_{\text{base}}, \text{if } d(\mathcal{M}) = 1. &
 \end{array}
\right.
\end{equation}
Subsequently, given the $F$-frame grid sequence, we generate the corresponding $F$ normalized video frames via a bilinear grid inverse sampling operation. 
Formally, for each frame index $t$, the source and target frames are computed as:
\begin{equation}
\left\{
 \begin{array}{lr}
 I^{\text{ref}_1}_t = \mathcal{G}(I_{\text{ref}_1}, G^{\text{ref}_1}_t), & \\
 I^{\text{tar}}_t = \mathcal{G}(I_{\text{tar}}, G^{\text{tar}}_t), &
 \end{array}
 \right.
\end{equation}
where $\mathcal{G}$ denotes the bilinear grid inverse sampling operator.
The operation samples the input image at the coordinates specified by the deformation grid, producing an output frame of identical spatial dimensions.
When sampled coordinates fall outside the interval $[-1,1]$, we assign values reflected across the border to those out-of-bounds grid locations.

We briefly introduce the warping functions $\mathcal{W}$ for a subset of motion types and their combinations in the following paragraphs; indeed, motion types are fully customizable, and can be specified or combined freely.
Intuitively, we use pan-based, zoom-based, and rotation-based motions to mimic camera movements, while stretch-type and elastic-type deformations simulate random shape, layout, or motion variations of the editing objects or the entire scene.
\paragraph{Pan Motion Type.}
As shown on the right side of \cref{fig:Pixel-pair Temporal Warped Flow Field}, the pan motion transformation translates the point sets along a specified direction. 
For a given base grid $G_{\text{base}} \in [-1,1]^{H \times W \times 2}$, a translation distance $\alpha = f_{\alpha}(\mathcal{M}) > 0$, and a normalized temporal blending factor \(\tau = \frac{t}{F-1} \in [0,1]\), the warping function for rightward translation is defined as:
\begin{equation}
G_t = \mathcal{W}\bigl(G_{\text{base}}, \mathrm{pan\ right}, t, \alpha, S \bigr) = G_{\text{base}} + \tau\begin{pmatrix} \alpha \\ 0 \end{pmatrix}.
\end{equation}
Explicitly, for each spatial position \((h,w)\) in the grid:
\begin{equation}
G_t[h,w] = G_{\text{base}}[h,w] + \begin{pmatrix} \tau\cdot \alpha \\ 0 \end{pmatrix}.
\end{equation}
The blending factor \(\tau\) increases linearly from $0$ at the first frame ($t = 0 $) to $1$ at the final frame ($t = F - 1$), thereby generating a smooth change across the sequence. 
Similarly, the other three cardinal pan directions can be formulated as: 
\begin{equation}
 \left\{
 \begin{array}{lr}
 \mathcal{W}\bigl(G_{\text{base}}, \mathrm{pan\ left}, t, \alpha, S \bigr) = G_{\text{base}} - \tau\begin{pmatrix} \alpha \\ 0 \end{pmatrix}, & \\
 \mathcal{W}\bigl(G_{\text{base}}, \mathrm{pan\ up}, t, \alpha, S \bigr) = G_{\text{base}} - \tau\begin{pmatrix} 0 \\ \alpha \end{pmatrix}, & \\
 \mathcal{W}\bigl(G_{\text{base}}, \mathrm{pan\ down}, t, \alpha, S \bigr) = G_{\text{base}} + \tau\begin{pmatrix} 0 \\ \alpha \end{pmatrix}. & \\
 \end{array}
 \right.
\end{equation}
Other pan-based directions can be defined analogously.
The reverse motion type is obtained by reversing the generated grid sequence. 
In particular, for pan motion types, when sampled coordinates lie outside $[-1,1]$, we use the border values for out-of-bound grid locations, which is intended to account for cases in which the editing region is not yet fully visible in the first reference frame.

\paragraph{Zoom Motion Type.}
The zoom motion transformation scales the point sets uniformly over time, simulating camera zoom-in or zoom-out effects. 
Given the base grid $G_{\text{base}} \in [-1,1]^{H \times W \times 2}$, scaling bounds \(\alpha = (s_{\min}, s_{\max})\) obtained from \(f_{\alpha}(\mathcal{M})\), and a normalized blending factor \(\tau = \frac{t}{F-1} \in [0,1]\), the warping function for a uniform zooming is defined as:
\begin{equation}
G_t = \mathcal{W}\bigl(G_{\text{base}}, \mathcal{M}, t, \alpha, S \bigr) = \frac{G_{\text{base}}}{s(t)},
\end{equation}
where the time-dependent scaling factor \(s(t)\) is given by:
\begin{equation}
s(t) = 
\begin{cases}
s_{\min} + \tau\cdot (s_{\max} - s_{\min}), & \text{for }\mathcal{M}=\mathrm{zoom\text{-}in},\\[6pt]
s_{\max} - \tau\cdot (s_{\max} - s_{\min}), & \text{for }\mathcal{M}=\mathrm{zoom\text{-}out}.
\end{cases}
\end{equation}
Thus, a zoom-in motion linearly increases the scale from $s_{\min}$ to $s_{\max}$, effectively enlarging the image content as the coordinates are divided by a progressively larger number. 
Conversely, a zoom-out motion decreases the scale from $s_{\max}$ to $s_{\min}$, shrinking the apparent content. 
In this paper, we set $s_{\min} = 1.0$.
The same scaling factor is applied to both spatial dimensions, preserving the aspect ratio. 

\paragraph{Rotation Motion Type.}
The rotation motion transformation rotates the point sets about the origin over time, simulating camera rotation effects. 
Given the base grid $G_{\text{base}} \in [-1,1]^{H \times W \times 2}$, a maximum rotation angle $\alpha = \theta_{\max} > 0$ (in degrees) obtained from \(f_{\alpha}(\mathcal{M})\), and a normalized blending factor \(\tau = \frac{t}{F - 1} \in [0,1]\), the warping function for a uniform rotation is defined as: 
\begin{equation}
G_t = \mathcal{W}\bigl(G_{\text{base}}, \mathcal{M}, t, \alpha, S \bigr) = R\bigl(\theta(t)\bigr) \; G_{\text{base}},
\end{equation}
where the time-dependent rotation angle (in radians) is:
\begin{equation}
\theta(t) = 
\begin{cases}
\phantom{-}\tau\cdot \theta_{\max} \cdot \frac{\pi}{180}, & \text{for }\mathcal{M}=\mathrm{anticlockwise\ rotation},\\[6pt]
-\,\tau\cdot \theta_{\max} \cdot \frac{\pi}{180}, & \text{for }\mathcal{M}=\mathrm{clockwise\ rotation},
\end{cases}
\end{equation}
and \(R(\theta)\) denotes the standard 2D rotation matrix:
\begin{equation}
R(\theta) = 
\begin{pmatrix}
\cos\theta & -\sin\theta \\[4pt]
\sin\theta & \cos\theta
\end{pmatrix}.
\end{equation}
Explicitly, for each spatial position \((h,w)\) in the grid, the transformed coordinates are obtained via matrix-vector multiplication:
\begin{equation}
G_t[h,w] = R\bigl(\theta(t)\bigr) \; G_{\text{base}}[h,w].
\end{equation}
Thus, an anticlockwise rotation linearly increases the angle from 0 to $\theta_{\max}$ (converted to radians), whereas a clockwise rotation decreases the angle from 0 to $-\theta_{\max}$. 
The rotation is applied uniformly to all points, preserving relative distances and orientations within the image plane.

\paragraph{Stretch Motion Type.}
The \textbf{stretch-local} motion transformation applies spatially localized, time-varying deformations to simulate shape or layout changes of the editing region. 
Given the base grid $G_{\text{base}} \in [-1,1]^{H \times W \times 2}$, we sample a set of $k$ deformation centres, where each centre $j$ is defined by its sampled location \(\mathbf{c}_j=(x_j, y_j)\in[-0.8,0.8]^2\), an influence radius $\rho_j\in[\rho_{\min}, \rho_{\max}]$, and a strength $\sigma_j\in[\sigma_{\min}, \sigma_{\max}]$; $\rho_{\min}, \rho_{\max}, \sigma_{\min}, \sigma_{\max}$ are drawn from the hyperparameter function \(f_{\alpha}(\mathcal{M})\) and $k$ is sampled from the predefined interval. 
Let \(\tau = \frac{t}{F - 1} \in [0,1]\) be the normalized blending factor. 
For a grid point at spatial index \((h,w)\) with coordinate \(\mathbf{p}=(x, y)^\top=G_{\text{base}}[h,w]\), the displacement induced by the $j$-th centre is:
\begin{equation}
D_j(\mathbf{p}) =
\begin{pmatrix}
(x - x_j) \cdot \gamma_j(r_j) \cdot \sigma_j \cdot \tau \\[4pt]
(y - y_j) \cdot \gamma_j(r_j) \cdot \sigma_j \cdot \tau
\end{pmatrix},
\end{equation}
where $r_j=\bigl\|\mathbf{p} - \mathbf{c}_j\bigl\|_2$ and \(\gamma_j(r_j)=\max\bigl(0, 1 - \frac{r_j}{\rho_j}\bigr)\) is a linear attenuation factor that vanishes beyond the radius $\rho_j$. 
The total displacement at \((h,w)\) is the sum of contributions from all centres, and the deformation grid at frame $t$ is therefore:
\begin{equation}
G_t[h,w] = G_{\text{base}}[h,w] + \sum_{j=1}^{k} D_j\bigl(G_{\text{base}}[h,w]\bigr).
\end{equation}
Writing the whole transformation as a warping function gives:
\begin{equation}
\begin{aligned}
G_t &= \mathcal{W}\bigl(G_{\text{base}}, \mathrm{stretch\ local}, t, \alpha, S\bigr) \\
&= G_{\text{base}} + \sum_{j=1}^{k} D_j,
\end{aligned}
\end{equation}
where $D_j$ denotes the pointwise displacement field generated by the $j$-th centre. 
This stretch-local function produces smooth, radially decaying deformations that mimic natural shape changes; because the identical centres, radii, strengths, and time progression are applied to both the source and target images, the pixel-wise editing correspondence is preserved exactly throughout the generated video sequence.

The \textbf{stretch-global} motion transformation applies a piecewise-linear, anisotropic scaling to the entire image plane over time, thereby simulating smooth changes in the overall layout of the scene. 
Given the base grid $G_{\text{base}} \in [-1,1]^{H \times W \times 2}$, we first partition the temporal sequence of length $F$ into $m$ consecutive segments, where $m$ is randomly drawn from the interval $[m_{\min}, m_{\max}]$, where $m_{\min}, m_{\max}$ are obtained from the hyperparameter function $f_{\alpha}(\mathcal{M})$. 
Each segment $i$ (for $i=1,\dots,m$) contains $f_i$ frames, with $\sum_{i=1}^m f_i = F$. 

For every segment $i$, we independently sample a target stretch factor $\mathbf{s}_i = (s_{x,i},\; s_{y,i})$, where $s_{x,i} \in [s_{x,\min}, s_{x,\max}]$ and $s_{y,i} \in [s_{y,\min}, s_{y,\max}]$.
$s_{x,\min}, s_{x,\max}, s_{y,\min}, s_{y,\max}$ are obtained from $f_{\alpha}(\mathcal{M})$. 
Let $\mathbf{s}_0 = (1,1)$ denote the initial (identity) stretch factor. 
Within segment $i$, the stretch factor evolves linearly from $\mathbf{s}_{i-1}$ to $\mathbf{s}_i$. 
For the $j$-th frame inside that segment $(0 \le j < f_i)$, the intra-segment progress is defined as:
\begin{equation}
\tau = 
\begin{cases}
\dfrac{j}{f_i-1}, & \text{if } f_i > 1, \\[8pt]
0, & \text{if } f_i = 1.
\end{cases}
\end{equation}
The current stretch factor at that frame is therefore:
\begin{equation}
\begin{pmatrix} s_x(t) \\[4pt] s_y(t) \end{pmatrix}
= \mathbf{s}_{i-1} + \tau \bigl( \mathbf{s}_i - \mathbf{s}_{i-1} \bigr).
\end{equation}

The deformation grid at the corresponding global frame index $t$ is obtained by scaling the base grid coordinates independently in the two spatial dimensions:
\begin{equation}
G_t[h,w] = 
\begin{pmatrix}
s_x(t) \cdot x \\[4pt]
s_y(t) \cdot y
\end{pmatrix},
\end{equation}
where $(x,y)^\top = G_{\text{base}}[h,w]$.
Writing the whole transformation as a warping function yields:
\begin{equation}
\begin{aligned}
G_t &= \mathcal{W}\bigl(G_{\text{base}}, \mathrm{stretch\ global}, t, \alpha, S\bigr) \\
&= 
\begin{pmatrix}
s_x(t) \cdot G_{\text{base}}[...,0] \\[4pt]
s_y(t) \cdot G_{\text{base}}[...,1]
\end{pmatrix}.
\end{aligned}
\end{equation}
The identical piecewise linear scaling trajectory is applied to both the source and target images, and therefore, the pixel-wise editing correspondence is also preserved throughout the generated video sequence.

\paragraph{Elastic Motion Type.}
The elastic motion transformation generates a sequence of non-rigid, fluid-like displacement fields that evolve stochastically over time, producing dynamic video distortions. 
Given the base grid $G_{\text{base}} \in [-1,1]^{H \times W \times 2}$, for each frame index $t$, we sample an independent noise field $N^{(t)} \in [-1,1]^{H \times W \times 2}$ with independent, uniformly distributed random values via the seed $S$. 
A smooth displacement field $D^{(t)} \in \mathbb{R}^{H \times W \times 2}$ for frame $t$ is then obtained by convolving each channel of $N^{(t)}$ with a 2D Gaussian kernel $\mathcal{G}_{\sigma}$ of standard deviation $\sigma$:
\begin{equation}
D^{(t)} = \alpha \cdot \bigl( N^{(t)} \ast \mathcal{G}_{\sigma} \bigr),
\end{equation}
where $\ast$ denotes convolution with zero-padding, and $\alpha = \frac{\alpha_0}{\max(H, W)}$ scales the displacement amplitude inversely with the image size. 
$\sigma$ and $\alpha_0$ are the hyperparameters drawn from $f_{\alpha}(\mathcal{M})$. 
The convolution imposes spatial coherence within each frame, ensuring neighbouring points move in a correlated manner, while the independent noise for each frame produces a time-varying distortion pattern.

Let $\tau = \frac{t}{F-1} \in [0,1]$ be the normalized blending factor. 
The deformation grid at frame $t$ is obtained by adding a scaled version of the frame-specific displacement field to the base grid:
\begin{equation}
\begin{aligned}
G_t &= \mathcal{W}\bigl(G_{\text{base}}, \mathrm{elastic}, t, (\alpha_0, \sigma), S\bigr) \\
&= G_{\text{base}} + \tau \cdot D^{(t)}.
\end{aligned}
\end{equation}
Explicitly, for each spatial position $(h,w)$,
\begin{equation}
G_t[h,w] = G_{\text{base}}[h,w] + \tau \cdot D^{(t)}[h,w].
\end{equation}
Thus, for each frame, an independent random displacement field is generated and then scaled linearly from zero at the first frame ($\tau=0$) to the full amplitude at the last frame ($\tau=1$), producing a smooth, globally coherent yet temporally varying elastic distortion. 
Because the same random seed $S$ is used for both the source and target images, the identical sequence of displacement fields $\{D^{(t)}\}_{t=0}^{F-1}$ is applied, guaranteeing that pixel-wise editing correspondences are preserved exactly throughout the generated video sequence.

\paragraph{Composite Motion Type.}
Composite motion types can be constructed by composing the warping functions of two or more basic motion types to the base grid at each frame index $t$. 
We illustrate this using the simplest case of two motion types.
Given a base grid $G_{\text{base}} \in [-1,1]^{H \times W \times 2}$ and two warping functions $\mathcal{W}_1$ and $\mathcal{W}_2$ corresponding to motion types $\mathcal{M}_1$ and $\mathcal{M}_2$ respectively, the composite deformation is defined as:
\begin{equation}
G_t = \mathcal{W}_2 \left( \mathcal{W}_1 \left( G_{\text{base}}, \mathcal{M}_1, t, \alpha_1, S \right),\; \mathcal{M}_2, t, \alpha_2, S \right),
\end{equation}
where $\alpha_1, \alpha_2$ are the hyperparameter sets of the respective motion types and $S$ is the sample-dependent random seed. 
Because each individual transformation is continuous in time, the composite warp is also continuous, yielding a temporally smooth motion that combines the effects of both components.

As a concrete example, consider a composite motion formed by combining a rightward pan with a local stretching effect.
Given the base grid $G_{\text{base}} \in [-1,1]^{H \times W \times 2}$, we first generate the stretched grid sequence $\{G_t^{\text{stretch local}}\}_{t=0}^{F-1}$ via the stretch-local warping function $\mathcal{W}_{\text{stretch local}}$ with its corresponding hyperparameters. 
Let $\alpha_p > 0$ be the pan distance and $\tau = \frac{t}{F-1} \in [0,1]$ the temporal blending factor. 
The composite deformation grid at frame $t$ is then obtained by adding the linear pan displacement to the stretched grid:
\begin{equation}
\begin{aligned}
G_t &= \mathcal{W}\bigl(G_{\text{base}}, \mathrm{pan\ right\&stretch\ local}, t, (\alpha_p, \alpha_s), S\bigr) \\
&= \mathcal{W}_{\text{stretch local}}\bigl(G_{\text{base}}, t, \alpha_s, S\bigr) + \tau \begin{pmatrix} \alpha_p \\ 0 \end{pmatrix},
\end{aligned}
\end{equation}
where $\alpha_s$ denotes the hyperparameters of the stretch-local component. 
Explicitly, for each spatial position $(h,w)$,
\begin{equation}
G_t[h,w] = G_t^{\text{stretch local}}[h,w] + \begin{pmatrix} \tau \cdot \alpha_p \\ 0 \end{pmatrix}.
\end{equation}
Because both $\mathcal{W}_{\text{stretch local}}$ and the linear pan term are continuous in $t$, the composite warp is also continuous, producing a smooth, combined motion that simultaneously exhibits local shape deformation and global translation. 

We can freely define single motion types and then combine them arbitrarily.
The parameters for each motion type can be configured freely; for instance, the direction of a pan can be arbitrary. 
Motion types can be combined either by composing a selected subset to be applied simultaneously as described above, or by dividing the sequence temporally into multiple segments, each employing a distinct motion type, e.g., ``zoom random'' is composed of segments that are randomly sampled from ``zoom-in'' and ``zoom-out'' motion types. 
For subsequent segments, the initial zoom scale is set equal to the final zoom scale of the preceding segment, which ensures the smooth temporal transition of the whole video sequence.
In practice, we find that even performing pairwise combinations on only a subset of these motion types enables the model to learn video editing.

\paragraph{Text captions for video editing samples.}
For a video editing pair derived from the source and target images of an image editing sample, the same description related to the type of motion is appended to the original source and target captions. For instance, the motion caption corresponding to ``pan left'' includes ``The camera pans left.'', and that for ``zoom random'' includes ``The camera zooms randomly.'', thereby incorporating a description of the motion into the final caption. During the inference procedure, the original source and target captions are used without such motion-type appendages.
Concerning the editing captions, except for the first frame propagation task introduced in \cref{para:First Frame Propagation.}, all other tasks retain the image editing captions; that is, we have not deliberately distinguished between ``video'' and ``image'' in the editing captions. Besides, regarding the cross-attention process, the output caption corresponding to the target element is formed by the string concatenation of the editing caption and the target caption.

\paragraph{Influence of Motion Strength.}
The motion strength hyperparameters mapped by $f_{\alpha}(\mathcal{M})$ can be freely adjusted. 
When the motion strength is gentle, the generated video exhibits very smooth inter-frame transitions, resembling a high frame rate effect. 
Conversely, when the motion strength is increased, the inter-frame motion variation becomes more pronounced, analogous to a low frame rate effect.
We find that the strength of the motion parameters influences both the convergence speed and the editing effect. 
Specifically, when the strength is increased, the model converges faster and tends to perform rigid editing, such as appearance editing; conversely, when the strength is decreased, the model becomes more capable of non-rigid editing on the object's shape.
In this paper, the motion strength parameters are set to a moderate level.

\paragraph{What does the model see?}
When video pixels are flattened into visual tokens in the diffusion training space, the model observes the source and target videos that maintain pixel-wise editing correspondence across time, a readily discernible regularity. 
Our hypothesis that the model can thereby learn pixel-tracking temporal editing is confirmed in \cref{sec:experiments}.

\subsection{First Frame Modality Mimic Loss}
\label{subsec:First Frame Modality Mimic Loss}
\begin{figure}[htpb]
 \centering
 \includegraphics[width=1.0\linewidth, trim=0pt 0pt 0pt 0pt, clip]{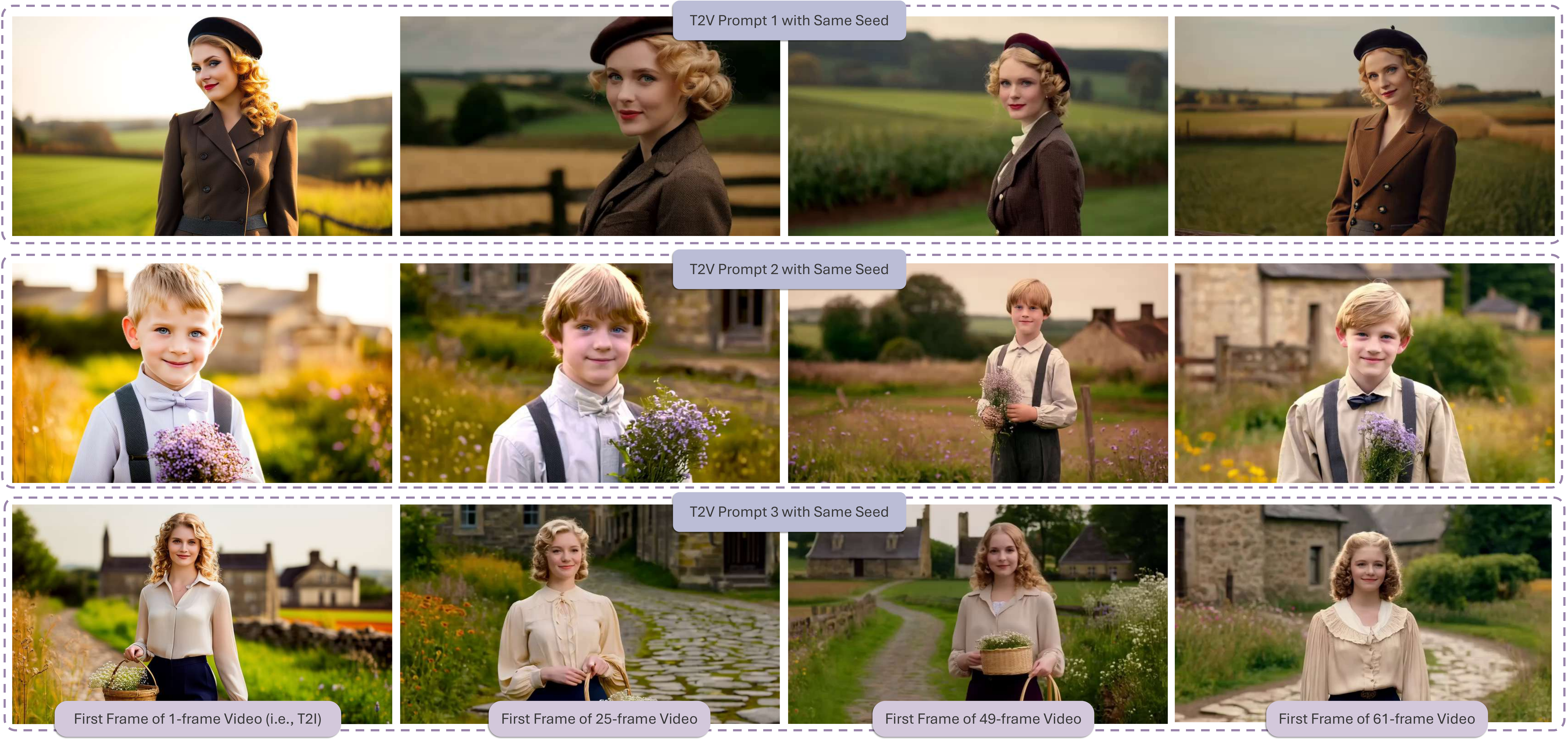} 
 \captionsetup{font=scriptsize}
 \caption{
 \textbf{T2V first frame results from the pretrained T2IV model under increasing inference frame counts, using identical prompts and random seeds, at a resolution of $832\times 480$.}
 When the inference frame count is 1, the output corresponds to the T2I result. It can be observed that the T2I result exhibits a stronger synthetic or artificial appearance, with weaker cinematic realism compared to the first frame result from temporal multi-frame inference.
 }
 \label{fig:flowmimic_fig_2}
\end{figure}

\begin{figure*}[htpb]
 \centering
 \includegraphics[width=1\linewidth, trim=0pt 2pt 0pt 2pt, clip]{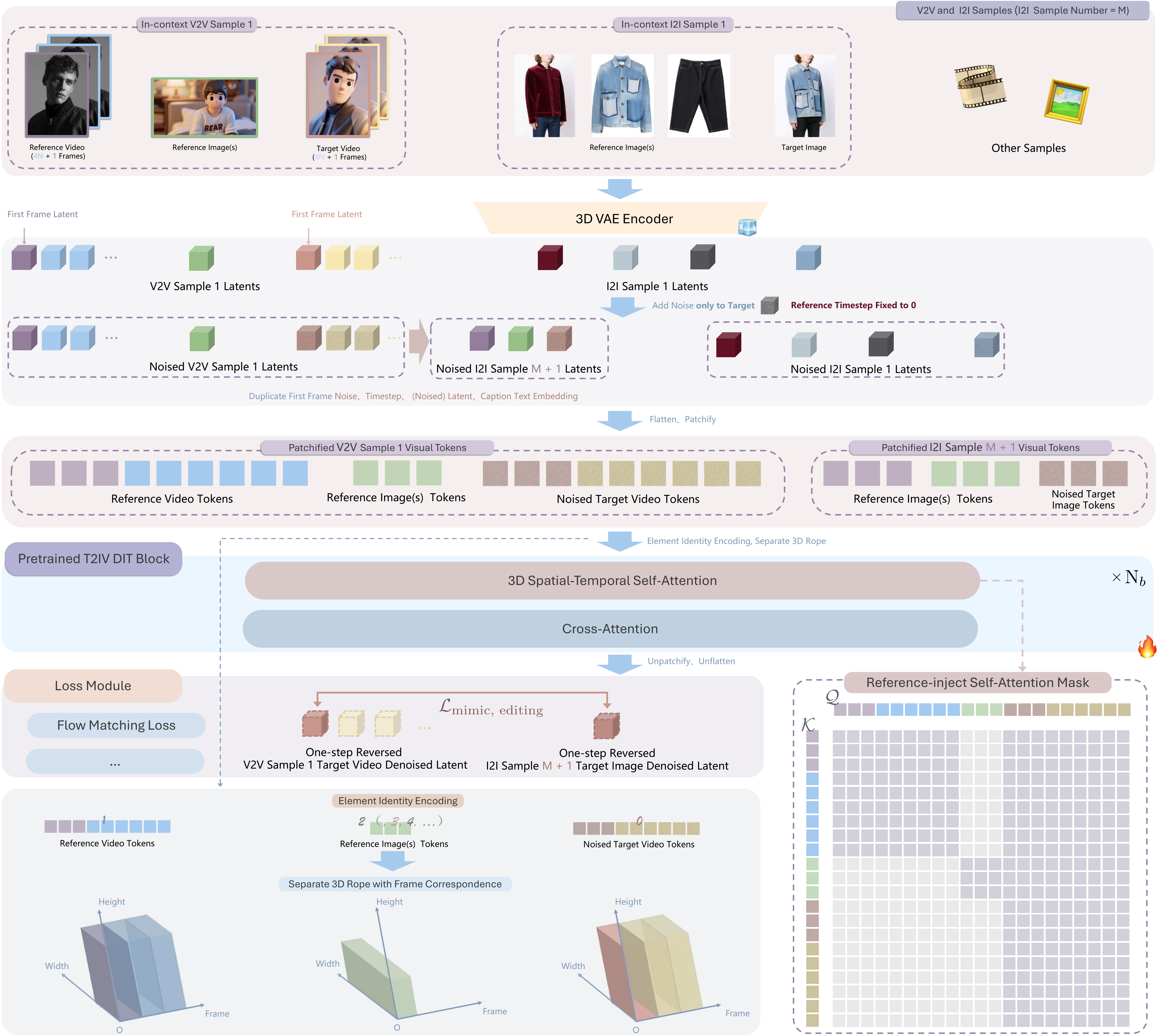} 
 \captionsetup{font=scriptsize}
 \caption{
 \textbf{Overview of the modality mimic editing loss in editing training step and our model architecture adaptations to editing task, including reference-inject self-attention mask, separate 3D rope with frame correspondence, and element identity encoding.}
 In the schematic illustrating the modality mimic editing loss, for better clarity, we present the newly created I2I sample as the $(M+1)$-th sample in the current training batch. 
 In our implementation, this new sample replaces the first I2I sample in the current training batch.
 Please refer to \cref{subsec:First Frame Modality Mimic Loss} and \cref{subsec:Model Architecture Adaptation to Editing Task} for details, respectively.
 }
 \label{fig:flowmimic_fig_3}
\end{figure*}

\begin{figure*}[htpb]
 \centering
 \includegraphics[width=1\linewidth, trim=0pt 2pt 0pt 2pt, clip]{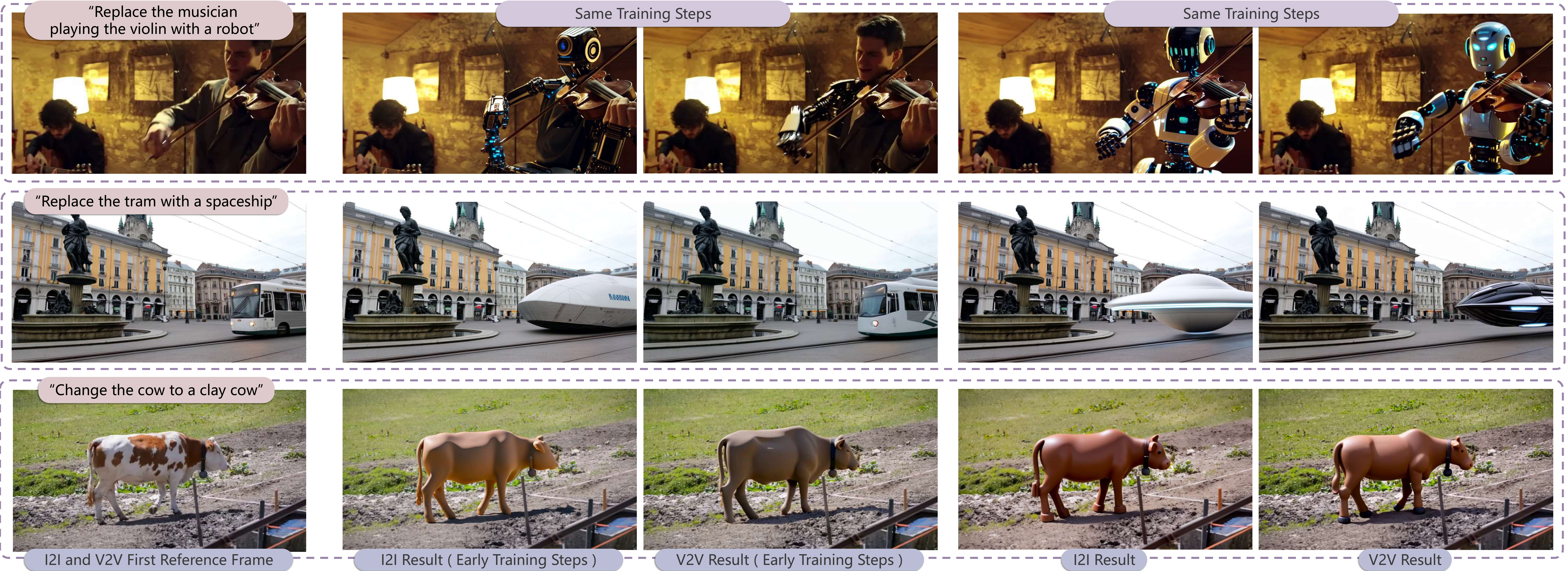} 
 \captionsetup{font=scriptsize}
 \caption{
 \textbf{Convergence speed comparision between I2I and V2V.}
 Given a reference video, I2I inference is applied to its first frame, while V2V inference is conducted on the complete video. A comparison is then performed between the first frames of the I2I and V2V inference outputs. The leftmost column presents the first frames of videos from FiVE-Bench~\cite{2025five}. The two central columns respectively exhibit the first frame comparative results between I2I and V2V inference at identical training steps during the nascent training phase; conversely, the two rightmost columns display analogous comparisons at the same training steps during the advanced training phase. It is discernible that during the early training stage, I2I exhibits faster convergence relative to V2V.
 }
 \label{fig:flowmimic_fig_4}
\end{figure*}

The image and video modalities can be regarded as a single modality, where an image is a special case of a single-frame video.
The pretrained T2IV model Wan2.1 is trained on a mixture of video and image data.
Naturally, we would expect the generative quality of T2I to closely match that of T2V, for instance, the distribution of its first frame, since the former can be viewed as a special instance of the latter.
As exemplified in \cref{fig:flowmimic_fig_2}, we perform inference under identical captions and seeds for 1, 25, 49, and 61 frames, respectively, with resolution $832\times 480$, and observe that the single-frame results exhibit a markedly more synthetic appearance than multi-frame outputs.
This indicates that simply mixing the data during training is insufficient to align the generative distributions of the image and video modalities closely.

Some research~\cite{xi2025sparse} analyzes that 3D self-attention heads in diffusion transformers naturally specialize, with some focusing on spatial relations and others on temporal coherence. 
Consequently, a key difference between single-frame T2I and multi-frame T2V is that the latter, by virtue of its temporal dimension, is inherently more suited to such an entangled, spatio-temporal full self-attention architecture; whereas the former may exhibit unnatural, synthetic artifacts due to the presence of temporal attention heads. 

A natural idea is to make the distribution of the model's denoising outcome for the first frame of a T2V sample—when treated as a non-temporal image sample—closely match the distribution when that same frame is denoised as the first frame of a multi-frame video sequence. 
In other words, the T2I capability is encouraged to mimic the T2V capability. 
This imitation of denoising outcome distributions serves to unify the generative distributions of the two temporally distinct modalities.

\paragraph{Modality Mimic Generation Loss.}
Our training procedure is divided into generation steps and editing steps, the latter distinguished by the inclusion of reference images or videos in their training samples. 
We first introduce the mimic strategy and mimic latent generation loss applied at regular intervals during generation steps. 
Specifically, as shown on the right of \cref{fig:flowmimic_fig_32}, on a generation step divisible by $n_{\text{mimic,gen}}$ (e.g., 5), we take a single T2V sample from the current batch, denoted as \(s^1_{\text{t2v}} \in \mathbb{R}^{(4N+1) \times H \times W \times 3}\).
The sample $s^1_{\text{t2v}}$ is subsequently encoded into the latent space by the 3D VAE encoder $\mathcal{E}$: \(z^1_{\text{t2v}} = \mathcal{E}(s^1_{\text{t2v}}) \in \mathbb{R}^{(N+1) \times \lfloor \frac{H}{8} \rfloor \times \lfloor \frac{W}{8} \rfloor \times 16}\), where the $4N$ frames after the first frame are rank-reduced along the temporal dimension to $N$ frames.
A Gaussian noise latent \(\epsilon^1_{\text{t2v}} \sim \mathcal{N}(0, \mathbf{I})\) of the same shape as $z^1_{\text{t2v}}$ is sampled, and a timestep \(t \sim p(t)\) is drawn. 
The noisy latent $\tilde{z}^1_{\text{t2v, t}}$ is then obtained via the forward diffusion process. 
The first frame of this noisy latent, \(\tilde{z}^1_{\text{t2i, t}} = \tilde{z}^1_{\text{t2v, t}}[:1] \in \mathbb{R}^{1 \times \lfloor \frac{H}{8} \rfloor \times \lfloor \frac{W}{8} \rfloor \times 16}\), is then treated as a new noisy T2I sample. 
Correspondingly, the first frame original latent $z^1_{\text{t2i}} = z^1_{\text{t2v}}[:1]$, the first frame noise $\epsilon^1_{\text{t2i}} = \epsilon^1_{\text{t2v}}[:1]$, the same timestep $t$, and the original first frame caption's text embeddings $c^1_{\text{t2v-f}}$ are assigned to this newly constructed T2I sample, which is then added to the training batch.
We also use the original first frame caption's text embeddings as the text embeddings for this T2V sample.
In this way, the newly added T2I sample and the first frame of the T2V sample share identical configurations before entering the DiT blocks, the sole distinction being that the latter is not an independent sample but is denoised jointly with the subsequent frames as a temporal sequence.
Both $\tilde{z}^1_{\text{t2v, t}}$ and $\tilde{z}^1_{\text{t2i, t}}$ are then flattened and patchified before being passed through $N_b$ identical DiT blocks that perform 3D spatial-temporal self-attention and cross-attention, followed by an unpatchify and unflatten operation to recover the original latent space shape: 
\begin{equation}
 \left\{
 \begin{aligned}
 \hat{v}^1_{\text{t2v}, t} &= \mathcal{F}^{-1} \circ \mathcal{P}^{-1} \bigl(\mathcal{D} \bigl( \mathcal{P} \circ \mathcal{F} (\tilde{z}^1_{\text{t2v, t}}), c^1_{\text{t2v-f}}, t\bigr) \bigr) , \\
 \hat{v}^1_{\text{t2i}, t} &= \mathcal{F}^{-1} \circ \mathcal{P}^{-1} \bigl( \mathcal{D} \bigl( \mathcal{P} \circ \mathcal{F} (\tilde{z}^1_{\text{t2i, t}}), c^1_{\text{t2v-f}}, t\bigr) \bigr) ,
 \end{aligned}
 \right.
\label{eq:30}
\end{equation}
where $\hat{v}_{t}$ denotes the predicted velocity field at timestep $t$, $\mathcal{F}$ flattening, $\mathcal{P}$ patchification, $\mathcal{D}$ the stack of DiT blocks, $\mathcal{P}^{-1}$ and $\mathcal{F}^{-1}$ the corresponding inverse operations, and $\circ$ the composition function operator.
Indeed, the order of noising and flattening can be interchanged arbitrarily, and whether flattening or not does not affect the computation of the loss in this paper.

The subsequent one-step reverse computation to obtain the estimated original latents, $\hat{z}^1_{\text{t2v},0}$ and $\hat{z}^1_{\text{t2i},0}$, is based on the flow matching assumption. 
The linear diffusion schedule is defined as \(x_t = A(t) x_0 + B(t) x_T\), with the velocity defined as the straight-line displacement $v_t = x_T - x_0$, where $x_0$ denotes the clean data, and $x_T$ represents the noise.
Given the noisy latent $x_t$ at timestep $t$ and the model's predicted velocity $\hat{v}_{t}$, the estimate for the original latent $\hat{x}_0$ is derived. 
Specifically, substituting $x_T = v_t + x_0$ into the forward equation yields: 
\begin{equation}
\begin{aligned}
x_t &= A(t) x_0 + B(t)(v_t + x_0) \\
&= (A(t)+B(t))x_0 + B(t)v_t.
\end{aligned}
\end{equation} 
Solving for $x_0$ gives the reverse formula:
\begin{equation}
\hat{x}_0 = \frac{x_t - B(t) \cdot \hat{v}_{t}}{A(t) + B(t)}.
\end{equation}
We employ a linear interpolation schedule configured with a maximum discrete timestep of $T=1000$. 
In this schedule, the coefficients are \(A(t) = 1 - \frac{t}{T}\) and \(B(t) = \frac{t}{T}\). 
Let $\tau = \frac{t}{T} \in [0,1]$, then we have \(A(\tau) = 1 - \tau\) and \(B(\tau) = \tau\), and their sum \(A(\tau) + B(\tau) = 1\). 
This simplification leads to the efficient one-step reverse formula:
\begin{equation}
\hat{x}_0 = x_t - \tau \cdot \hat{v}_{t}.
\label{eq:x0}
\end{equation}
Similarly, solving for $x_T$ yields:
\begin{equation}
\hat{x}_T = x_t + (1 - \tau) \cdot \hat{v}_{t}.
\label{eq:xT}
\end{equation}
Applying this core relationship \cref{eq:x0} to the video and image pathways, the estimated original latents are:
\begin{equation}
\left\{
\begin{aligned}
\hat{z}^1_{\text{t2v},0} &= \tilde{z}^1_{\text{t2v}, t} - \tau \cdot \hat{v}^1_{\text{t2v}, t}, \\
\hat{z}^1_{\text{t2i},0} &= \tilde{z}^1_{\text{t2i}, t} - \tau \cdot \hat{v}^1_{\text{t2i}, t}.
\end{aligned}
\right.
\label{eq:35}
\end{equation}

Let $\hat{z}^1_{\text{t2v-f},0} = \hat{z}^1_{\text{t2v},0}[:1]$ denote the estimated original latent of the first frame of our selected T2V sample.
In fact, $\hat{z}^1_{\text{t2v-f},0}$ and $\hat{z}^1_{\text{t2i},0}$ are the estimates derived from the same single-frame visual tokens under identical conditions—the sole distinction being that the former is processed within a temporal context (undergoing temporal attention alongside subsequent frames) while the latter is processed as an independent image. 
To encourage the T2I capability to mimic the T2V capability, we first compute the probability distributions of these two estimates. 
Each flattened latent $x \in \mathbb{R}^{D}$ is transformed into a probability distribution via the softmax function, defined for the $j$-th element as:
\begin{equation}
\operatorname{softmax}(x)_j = \frac{\exp(x_j)}{\sum\limits_{k=1}^{D} \exp(x_k)}, \quad j = 1, \dots, D,
\end{equation}
where $D = \lfloor \frac{H}{8} \rfloor \times \lfloor \frac{W}{8} \rfloor \times 16$.
Applying this to the respective latents yields the discrete probability distributions:
\begin{equation}
\begin{cases}
p^1_{\text{t2v-f,0}} = \operatorname{softmax}\bigl(\mathcal{F}(\hat{z}^1_{\text{t2v-f},0})\bigr), \\[6pt]
p^1_{\text{t2i,0}} = \operatorname{softmax}\bigl(\mathcal{F}(\hat{z}^1_{\text{t2i},0})\bigr).
\end{cases}
\label{eq:37}
\end{equation}

The Kullback-Leibler (KL) divergence~\cite{joyce2025kullback} is then computed between the two resulting probability distributions to quantify their discrepancy. 
The KL divergence from the image distribution $p^1_{\text{t2i,0}}$ to the target video first frame distribution $p^1_{\text{t2v-f,0}}$ is defined as:
\begin{equation}
D_{\text{KL}}\bigl(p^1_{\text{t2v-f,0}} \,\big\|\, p^1_{\text{t2i,0}}\bigr) = \sum_{j=1}^{D} p^1_{\text{t2v-f,0}}(j) \log \frac{p^1_{\text{t2v-f,0}}(j)}{p^1_{\text{t2i,0}}(j)},
\label{eq:38}
\end{equation}
where the sum runs over all $D$ feature dimensions. 
This non-negative measure is zero if and only if the two distributions are identical almost everywhere. 
The first frame modality mimic generation loss $\mathcal{L}_{\text{mimic, gen}}$ is defined as this KL divergence:
\begin{equation}
\mathcal{L}_{\text{mimic, gen}} = D_{\text{KL}}\bigl(p^1_{\text{t2v-f,0}} \,\big\|\, p^1_{\text{t2i,0}}\bigr).
\label{eq:39}
\end{equation}
We consider that the distributional shapes of $p^1_{\text{t2v-f,0}}$ and $p^1_{\text{t2i,0}}$ respectively reflect the visual perceptual characteristics of the estimated original latents $\hat{z}^1_{\text{t2v-f},0}$ and $\hat{z}^1_{\text{t2i},0}$.
Minimizing $\mathcal{L}_{\text{mimic, gen}}$ encourages the denoising outcome of the standalone image to become statistically indistinguishable from that of the corresponding frame when it is processed as part of a temporal sequence, thereby directly realizing the objective of making the T2I capability mimic the T2V capability. 
This distribution alignment loss between the estimated original latents is added to the overall training objective on the sampled generation steps, steering the model parameters towards a unified generative distribution for the two modalities.

As visualized on the right of \cref{fig:flowmimic_fig_32}, we decode the estimated original latents, $\hat{z}^1_{\text{t2v-f},0}$ and $\hat{z}^1_{\text{t2i},0}$, back to the image space via the 3D VAE decoder $\widetilde{\mathcal{D}}$ at the very beginning of training. 
It can be seen that the resulted \(\widetilde{\mathcal{D}}(\hat{z}^1_{\text{t2v-f},0})\) is markedly clearer and more coherent than \(\widetilde{\mathcal{D}}(\hat{z}^1_{\text{t2i},0})\). 
This validates that our motivation to mimic the T2V capability is meaningful.

\begin{figure*}[htpb]
 \centering
 \includegraphics[width=1\linewidth, trim=0pt 2pt 0pt 2pt, clip]{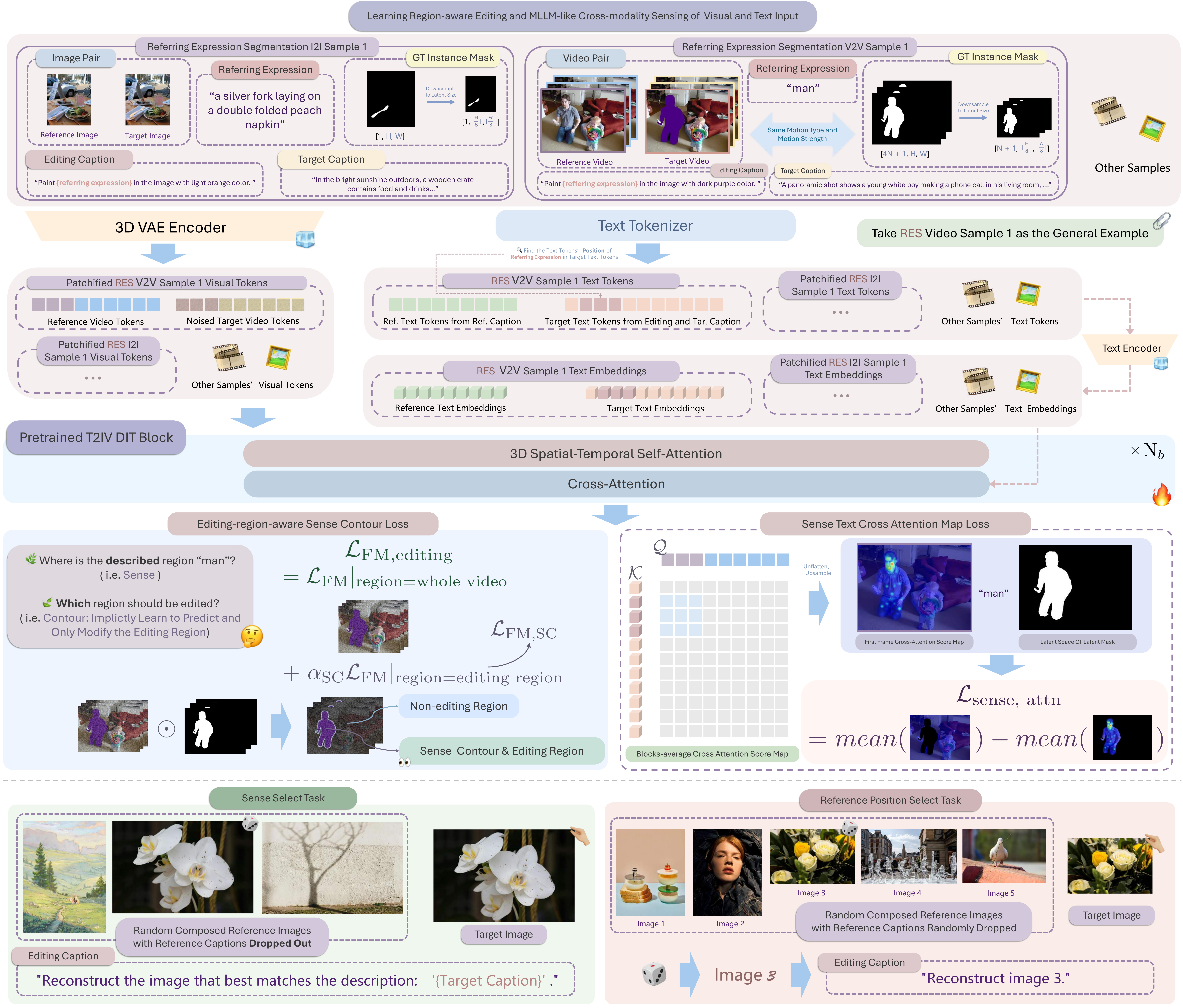} 
 \captionsetup{font=scriptsize}
 \caption{
 \textbf{Overview of sense-related tasks and losses.}
 Additionally, the reference position select task is introduced in the bottom-right corner. Please refer to \cref{subsec:Sense Losses} and \cref{subsec:Model Architecture Adaptation to Editing Task} for details, respectively.
 }
 \label{fig:flowmimic_fig_5}
\end{figure*}

\paragraph{Modality Mimic Editing Loss.}
\label{para:Modality Mimic Editing Loss.}
As illustrated in \cref{fig:flowmimic_fig_3}, during editing steps, we also aim for the editing capabilities of the two modalities to be unified through mutual imitation, and therefore we impose a similarly motivated modality-mimic editing loss. 
Empirically, as shown in \cref{fig:flowmimic_fig_4}, I2I convergence is typically observed earlier than V2V. Once converged on the same task, V2V generally achieves better editing quality. 
Therefore, we encourage V2V to mimic I2I in order to accelerate its convergence, and I2I to mimic V2V so as to attain higher-quality results. 

In line with prevailing image editing~\cite{chen2025unireal,xiao2025omnigen} and video editing~\cite{ye2025unic,univideo,mou2025instructx,chen2026vino} models that adopt visual in-context learning inspired by the counterparts in large language models, our editing samples are arranged in a visual in-context fashion: the reference inputs and the target are concatenated along the temporal dimension and flattened as a single token sequence to the DiT blocks.
It should be noted that we regard the reference elements as conditional inputs, which, like the caption text embeddings, should provide guidance throughout the denoising process and therefore remain unaffected by noise. 
Accordingly, noise is applied only to the target element, while the timesteps for all reference elements are fixed at zero.
The flow matching loss is computed solely for the target element.

On the editing step divisible by $n_{\text{mimic,editing}}$ (e.g., 3), we take a single V2V sample from the current batch, denoted as \(s^1_{\text{v2v}}=\{V_{\text{ref}^1_1}, I_{\text{ref}^1_2}, \dots, I_{\text{ref}^1_{n}}, V_{\text{tar}^1}\}\), where $V_{\text{ref}^1_1} \in \mathbb{R}^{(4N+1) \times H \times W \times 3}$ and $V_{\text{tar}^1} \in \mathbb{R}^{(4N+1) \times H \times W \times 3}$ are the source and edited videos respectively, and $\{I_{\text{ref}^1_i}\}_{i=2,3,\dots,n}$ are the optional reference images.
The sample elements are subsequently encoded into the latent space by the 3D VAE encoder $\mathcal{E}$: $\{\mathcal{E}(V_{\text{ref}^1_1}), \mathcal{E}(I_{\text{ref}^1_2}), \dots, \mathcal{E}(V_{\text{tar}^1})\} = \{z^1_{\text{ref}_1}, z^1_{\text{ref}_2}, \dots, z_{\text{tar}^1}\}$. 
The encoded latents of all sample elements are then flattened and concatenated in order along the temporal dimension, yielding: $z^{1}_{\text{v2v},\mathcal{F}} = z^1_{\text{ref}_1,\mathcal{F}} \oplus z^1_{\text{ref}_2,\mathcal{F}} \oplus \dots \oplus z_{\text{tar}^1,\mathcal{F}}$, where $\oplus$ denotes the concatenation operation.
A Gaussian noise latent $\epsilon^1_{\text{tar}} \sim \mathcal{N}(0, \mathbf{I})$ of the same shape as $z_{\text{tar}^1}$ is sampled and flattened, and a timestep $t \sim p(t)$ is drawn. 
The noisy target latent $\tilde{z}^1_{\text{tar}, t, \mathcal{F}}$ is obtained via the forward diffusion process. 
After corrupting the target latent with noise, we obtain the noisy sample latent $\tilde{z}^1_{\text{v2v,t},\mathcal{F}} = z^1_{\text{ref}_1,\mathcal{F}} \oplus z^1_{\text{ref}_2,\mathcal{F}} \oplus \dots \oplus \tilde{z}^1_{\text{tar}, t,\mathcal{F}}$.
Similarly, the first frame latent is extracted from each element of $\tilde{z}^1_{\text{v2v,t}}$ to form a new I2I sample noisy latent: 
\begin{equation}
 \tilde{z}^1_{\text{i2i},\text{t},\mathcal{F}} = \mathcal{F}(z^1_{\text{ref}_1}[:1]) \oplus \mathcal{F}(z^1_{\text{ref}_2}[:1]) \oplus \mathcal{F}(\tilde{z}^1_{\text{tar}, t}[:1]).
\end{equation}
Correspondingly, the first frame original latent: 
$$
z^1_{\text{i2i},\mathcal{F}} = \mathcal{F}(z^1_{\text{ref}_1}[:1]) \oplus \mathcal{F}(z^1_{\text{ref}_2}[:1]) \dots \oplus \mathcal{F}(z_{\text{tar}^1}[:1]),
$$
the first frame noise $\epsilon^1_{\text{i2i},\mathcal{F}} = \mathbf{0} \oplus \mathbf{0} \oplus \dots \oplus \mathcal{F}(\epsilon^1_{\text{tar}}[:1])$, where $\mathbf{0}$ denotes zero noise for the reference elements, the same timestep $t$ for the target portion (with timesteps for reference portions fixed at zero), and the original caption's text embeddings without motion descriptions $c^1_{\text{v2v,ori}}$ are assigned to this newly constructed I2I sample, which is then added to the training batch. 

In this way, the newly added I2I sample and the first frames of the corresponding elements in the V2V sample share identical configurations before entering the DiT blocks.
The sole distinction is that the first target frame of the latter, within the 3D spatial-temporal self-attention mechanism, interacts with reference and target elements that possess a full temporal structure. 
We assume the first frame of the target element to focus more on editing the first frame of the reference element—that is, to be more concentrated on spatial attention. 
To this end, the model's editing capability for the first frame of the reference elements in the V2V task should mimic its capability in the I2I task.

Both $\tilde{z}^1_{\text{v2v,t}}$ and $\tilde{z}^1_{\text{i2i,t}}$ are processed by the DiT blocks to obtain the predicted target velocity fields.
Following the same flow matching derivation and one-step reverse computation detailed in the generation loss section, and applying the normalized timestep $\tau = \frac{t}{T}$, we obtain the estimated original latents for the target elements:
\begin{equation}
\left\{
\begin{aligned}
\hat{z}^1_{\text{v2v,tar},0} &= \tilde{z}^1_{\text{tar}, t} - \tau \cdot \hat{v}^1_{\text{v2v,tar}, t}, \\
\hat{z}^1_{\text{i2i,tar},0} &= \tilde{z}^1_{\text{tar}, t}[:1] - \tau \cdot \hat{v}^1_{\text{i2i,tar}, t}.
\end{aligned}
\right.
\end{equation}
Similar to the modality mimic generation loss, we extract the first frame of the target element. 
Let $\hat{z}^1_{\text{v2v,tar-f},0} = \hat{z}^1_{\text{v2v,tar},0}[:1]$ denotes the estimated original latent of the first target frame from the V2V sample, we compute the probability distributions of the estimates for $\hat{z}^1_{\text{v2v,tar-f},0}$ and $\hat{z}^1_{\text{i2i,tar},0}$.
Applying the softmax function yields the discrete probability distributions:
\begin{equation}
\begin{cases}
p^1_{\text{v2v,tar-f,0}} = \operatorname{softmax}\bigl(\mathcal{F}(\hat{z}^1_{\text{v2v,tar-f},0})\bigr), \\[6pt]
p^1_{\text{i2i,tar,0}} = \operatorname{softmax}\bigl(\mathcal{F}(\hat{z}^1_{\text{i2i,tar},0})\bigr).
\end{cases}
\end{equation}
The modality mimic editing loss is then defined as the KL divergence between these two distributions:
\begin{equation}
\mathcal{L}_{\text{mimic, editing}} = D_{\text{KL}}\bigl(p^1_{\text{v2v,tar-f,0}} \,\big\|\, p^1_{\text{i2i,tar,0}}\bigr).
\end{equation}

Although the KL divergence possesses a directional (i.e., asymmetric) nature, the two input distributions involved are, in fact, produced by different pathways of the same model. 
Consequently, minimizing this loss does not simply drive one distribution to match a fixed target; rather, it facilitates a bidirectional distribution alignment, encouraging the two distributions to mutually mimic during training.
In fact, during the earlyfocused exploration phase, we incorporated additional measures—the Jensen-Shannon (JS) divergence and Hellinger distance—as loss functions, and observed a moderate improvement in convergence speed and performance. 
In this paper, for the simplicity of the loss functions, we employ only the KL divergence as the measure. 
This choice does not alter the core nature of the method.
Furthermore, we contend that the motivation and methodology of the modality mimicry section can be generalized into different paradigms from different perspectives, which we discuss in \cref{sec:The Modality Mimicry Paradigms}.

\subsection{Sense Losses}
\label{subsec:Sense Losses}
When performing an editing task, the model should first perceive the visual content of the source image or source video, then correctly localize the corresponding region within it based on the textual description of the object to be edited in the instruction, and finally apply changes exclusively to that region without affecting other, non-editing areas. 
We refer to this combined capability of visual perception and instruction-guided localization as ``sense''. 
This sense capability naturally relates to the established task of referring expression segmentation, while the required region-awareness during editing can be steered by a region-specific loss. 
Furthermore, because training data for referring expression segmentation provides readily available, off-the-shelf segmentation masks, these masks can be employed directly as supervision for the editing-region-aware loss, eliminating the need for additional, labour-intensive annotation. 

It is worth discussing that a straightforward method is to use vision-language models (VLMs) such as SAM2~\cite{ravisam}, SAM3~\cite{carion2025sam}, and Qwen-vl~\cite{wang2024qwen2,bai2025qwen2,bai2025qwen3} during both training and inference to annotate region masks, which are then fed as auxiliary conditional inputs—for instance, via in-context learning. The primary aim of this work, however, is to explore the construction of a model that internalizes this sense capability. 
We posit that training on multiple editing tasks can implicitly enhance the model's sense, while this capability can be further refined explicitly through the inclusion of referring expression segmentation. 
Fundamentally, we aim for the model to possess this ability to a sufficient degree intrinsically, so that the acquired sense benefits the model's overall editing performance without compelling it to inherit the potential limitations of the external tools on which mask-conditioned approaches predominantly rely.

\paragraph{Referring Expression Segmentation.}
\label{para:Referring Expression Segmentation.}
We first briefly introduce the referring expression segmentation task.
Multimodal referring expression segmentation addresses the task of segmenting target objects in visual inputs—images or videos—guided by a natural-language referring expression. 
As illustrated in \cref{fig:flowmimic_fig_5}, given the textual referring expression “a silver fork laying on a double folded peach napkin'', the model is required to produce pixel-wise masks that segment the described objects in the image.
Formally, referring expression segmentation (RES) takes an image $I \in \mathbb{R}^{H \times W \times 3}$ and a referring expression $R$ as inputs. 
Its output is a set of binary masks \(\{M^{(k)}\}_{k=1}^{K}\) with each \(M^{(k)} \in \{0,1\}^{H \times W}\), which collectively and precisely mark all regions in $I$ corresponding to the target objects in $R$; here $K$ is the number of target objects. 
This formulation poses a core challenge in multimodal understanding, demanding tight alignment between linguistic and visual representations. 
Unlike conventional semantic or instance segmentation, which operates on a fixed set of categories, RES necessitates a deeper comprehension of the linguistic nuances in $R$ and the ability to reason about spatial, attributive, and relational cues among objects in the scene. 

Referring video object segmentation (RVOS)~\cite{wu2022language,pan2025semantic} extends this setting to the video domain. 
Given a video sequence $V = \{V_t\}_{t=1}^{F}$ (each frame $V_t \in \mathbb{R}^{H \times W \times 3}$) and a referring expression $R$, the goal is to generate a set of temporally consistent binary mask sequences \(\{M^{(k)}\}_{k=1}^{K}\), where each \(M^{(k)} = \{M^{(k)}_t\}_{t=1}^{F}\) and \(M^{(k)}_t \in \{0,1\}^{H \times W}\) denotes the segmentation of the $k$-th referred object in frame $V_t$. 
Beyond the challenges present in RES, RVOS introduces additional difficulties such as maintaining temporal coherence for each object's mask sequence, handling appearance changes of the referred objects over time, and robustly tracking them throughout the video.

\paragraph{Editing-region-aware Sense Contour Loss.}
\label{para:Editing-region-aware Sense Contour Loss.}
We employ referring expression datasets including the single-target datasets RefCOCO~\cite{yu2016modeling}, RefCOCOg~\cite{yu2016modeling}, RefCOCO+~\cite{yu2016modeling}, and RefClef~\cite{kazemzadeh2014referitgame}, as well as the multi-target (i.e., expression that
specifies multiple objects in the image) portion of the gRefCOCO~\cite{liu2023gres} dataset. 
Specifically, RefCOCO's expressions allow both location-based and appearance-based references in images, RefCOCO+'s expressions focus on appearance-based expressions,
RefCOCOg's expressions contain longer and more complex expressions, and RefClef's expressions are simple and focus on real-world expressions.
We adapt the referring expression segmentation task into a region-coloring formulation that is more amenable to in-context visual learning: the editing task is to overlay the region in the source visual inputs that corresponds to the textual expression with a specified, uniform color.
Leveraging our temporal pixel-pair warped flow paradigm alongside the RES image editing data, we also perform online generation of RVOS video editing data.

As shown in \cref{fig:flowmimic_fig_5}, during editing steps, a fixed proportion of RES I2I and RES V2V (i.e., RVOS) samples are included in the training batch. 
We hereby use a RES V2V sample as the example; the RES I2I sample is treated as its single-frame special case, and the sense losses are computed in an identical manner.

Given a RES V2V sample $s_{\text{v2v}}=\{V_{\text{ref}_1}, V_{\text{tar}}\}$, where $V_{\text{ref}_1} \in \mathbb{R}^{(4N+1) \times H \times W \times 3}$ and $V_{\text{tar}} \in \mathbb{R}^{(4N+1) \times H \times W \times 3}$ are the source and edited videos respectively, the sample elements are encoded into the latent space by the 3D VAE encoder $\mathcal{E}: z_{\text{v2v}} = \{\mathcal{E}(V_{\text{ref}_1}), \mathcal{E}(V_{\text{tar}})\} = \{z_{\text{ref}_1}, z_{\text{tar}}\}$.

To maintain consistency with the pretrained T2IV model, each reference element in the editing sample is assigned a corresponding textual caption. In particular, the caption for the target element is formed by concatenating the editing instruction with the original textual caption of the target. Within the cross-attention layers, the visual tokens of each element attend exclusively to the textual embeddings derived from its own caption.
In general, we denote the set of captions for a V2V sample as $c_{p,\text{v2v}}=\{c_{p,\text{ref}_1}, c_{p,\text{ref}_2}, \dots, c_{p,\text{ref}_{n}}, c_{p,\text{output}}\}$ , where each $c_{p,\text{ref}_j}$ is the corresponding textual description of the $j$-th reference element, and $c_{p,\text{output}} = c_{p,\text{editing}}\cup c_{p,\text{tar}}$. $c_{p,\text{editing}}$ is the editing instruction, $c_{p,\text{tar}}$ is the original description of the target element and $\cup$ denotes string concatenation.
In the RES task, we apply dropout to all reference captions $\{c_{p,\mathrm{ref}_j}\}_{j=1}^n$ and employ a template-based instruction of the form $c_{p,\text{editing}} = \text{``Paint} \ \llbracket R \rrbracket \ \text{in the image with} \ \llbracket color \rrbracket \text{.''}$ (and similar variants) to construct the editing instruction, where $R$ denotes the referring expression and $color$ specifies the target color. The delimiters $\llbracket \cdot \rrbracket$ here are used to mark positions for phrase insertion, and they are not included in the captions.
Besides, $c_{\text{v2v}}$ is denoted as the concatenated text embeddings of the captions corresponding to each element in $s_{\text{v2v}}$.

When generating a RES V2V sample from its corresponding RES I2I sample, the associated ground-truth referring expression segmentation mask is likewisely processed through the pixel-pair temporal warped flow field. 
This yields a temporally coherent mask sequence \(\{m_{\text{tar},j}\}_{j=1}^{4N+1} \in \mathbb{R}^{(4N+1) \times H \times W}\) that shares the identical frame count, motion type, motion strength, and random seed as the source and target videos, $V_{\text{ref}_1}$ and $V_{\text{tar}}$, thereby ensuring $\{m_{\text{tar},j}\}_{j=1}^{4N+1}$ serves as a pixel-aligned temporal ground-truth object mask sequence for the V2V sample.
This mask sequence is then downsampled to match the size of the latent representation $z_{\text{tar}}$. 
Specifically, to obtain a latent space ground-truth editing region mask sequence of target size \(\mathbb{R}^{(N+1) \times \lfloor \frac{H}{8} \rfloor \times \lfloor \frac{W}{8} \rfloor \times 16}\), the first frame of $\{m_{\text{tar},j}\}_{j=1}^{4N+1}$ is spatially downsampled via bilinear interpolation to \(\mathbb{R}^{1 \times \lfloor \frac{H}{8} \rfloor \times \lfloor \frac{W}{8} \rfloor}\), while the subsequent $4N$ frames are spatiotemporally downsampled via trilinear interpolation to \(\mathbb{R}^{N \times \lfloor \frac{H}{8} \rfloor \times \lfloor \frac{W}{8} \rfloor}\). These components are then concatenated along the temporal dimension, binarized, unsqueezed, and broadcasted to form the final latent space mask sequence $\mathcal{M} = \{\mathcal{M}_{\text{tar},j}\}_{j=1}^{N+1}$.

The sample latent $z_{\text{v2v}} = [z_{\text{ref}_1}, z_{\text{tar}}]$ is subsequently perturbed with gaussian noise, yielding the noisy latent $\tilde{z}_{\text{v2v},t} = [z_{\text{ref}_1}, \tilde{z}_{\text{tar},t}]$, where $t$ denotes the sampled timestep. This noisy latent is then processed by the DiT blocks to obtain the predicted target velocity field: $\hat{v}_{\text{v2v, tar}, t}=\mathcal{F}^{-1} \circ \mathcal{P}^{-1} \bigl(\mathcal{D} \bigl( \mathcal{P} \circ \mathcal{F} (\tilde{z}_{\text{v2v, t}}), c_{\text{v2v}}, t\bigr) \bigr)$.
The original flow matching loss can be formulated as:
\begin{equation}
 \begin{aligned}
 \mathcal{L}_{\text{FM,global}} &= \mathcal{L}_{\text{FM}} |_{\text{region=whole video}} \\
 &= \bigl\| \hat{v}_{\text{v2v,tar}, t} - v_{\text{v2v,tar},t} \bigr\|^2_2 \\
 &= \bigl\| (\hat{z}_{\text{v2v,tar},T} - \hat{z}_{\text{v2v,tar},0}) - ({\epsilon}_{\text{tar}} - z_{\text{tar}}) \bigr\|^2_2,
 \end{aligned}
\end{equation}
where $\hat{z}_{\text{v2v,tar},T}$ and $\hat{z}_{\text{v2v,tar},0}$ can be obtained following \cref{eq:x0} and \cref{eq:xT}, and ${\epsilon}_{\text{tar}}$ is the added target gaussian noise.

The editing-region-aware sense contour loss is motivated by the objective of guiding the model to localize the region corresponding to the textual instruction and to apply changes exclusively within that region. Consequently, we compute a region-aware, local flow matching loss solely over the area defined by the latent editing region mask $\mathcal{M}$:
\begin{equation}
 \begin{aligned}
 \mathcal{L}_{\text{FM,SC}} &= \mathcal{L}_{\text{FM}} |_{\text{region=editing region}} \\
 &= \bigl\| \hat{v}_{\text{v2v,tar}, t} \odot \mathcal{M} - v_{\text{v2v,tar},t} \odot \mathcal{M} \bigr\|^2_2 \\
 &= \bigl\| (\hat{z}_{\text{v2v,tar},T} \odot \mathcal{M} - \hat{z}_{\text{v2v,tar},0} \odot \mathcal{M}) \\
 &\quad - ({\epsilon}_{\text{tar}} \odot \mathcal{M} - z_{\text{tar}} \odot \mathcal{M}) \bigr\|^2_2,
 \end{aligned}
\end{equation}
where ``SC'' represents ``Sense Contour'' and $\odot$ is the Hadamard operation.
Hence, the complete flow matching loss for the editing steps can be expressed as the sum of the original loss over the entire region and the editing-region-aware sense contour loss:
\begin{equation}
\mathcal{L}_{\text{FM,editing}} = \mathcal{L}_{\text{FM,global}} + \alpha_{\text{SC}} \, \mathcal{L}_{\text{FM,SC}},
\end{equation}
where $\alpha_{\text{SC}}$ is the weighting hyperparameter.

\begin{figure*}[htpb]
 \centering
 \includegraphics[width=1\linewidth, trim=0pt 2pt 0pt 2pt, clip]{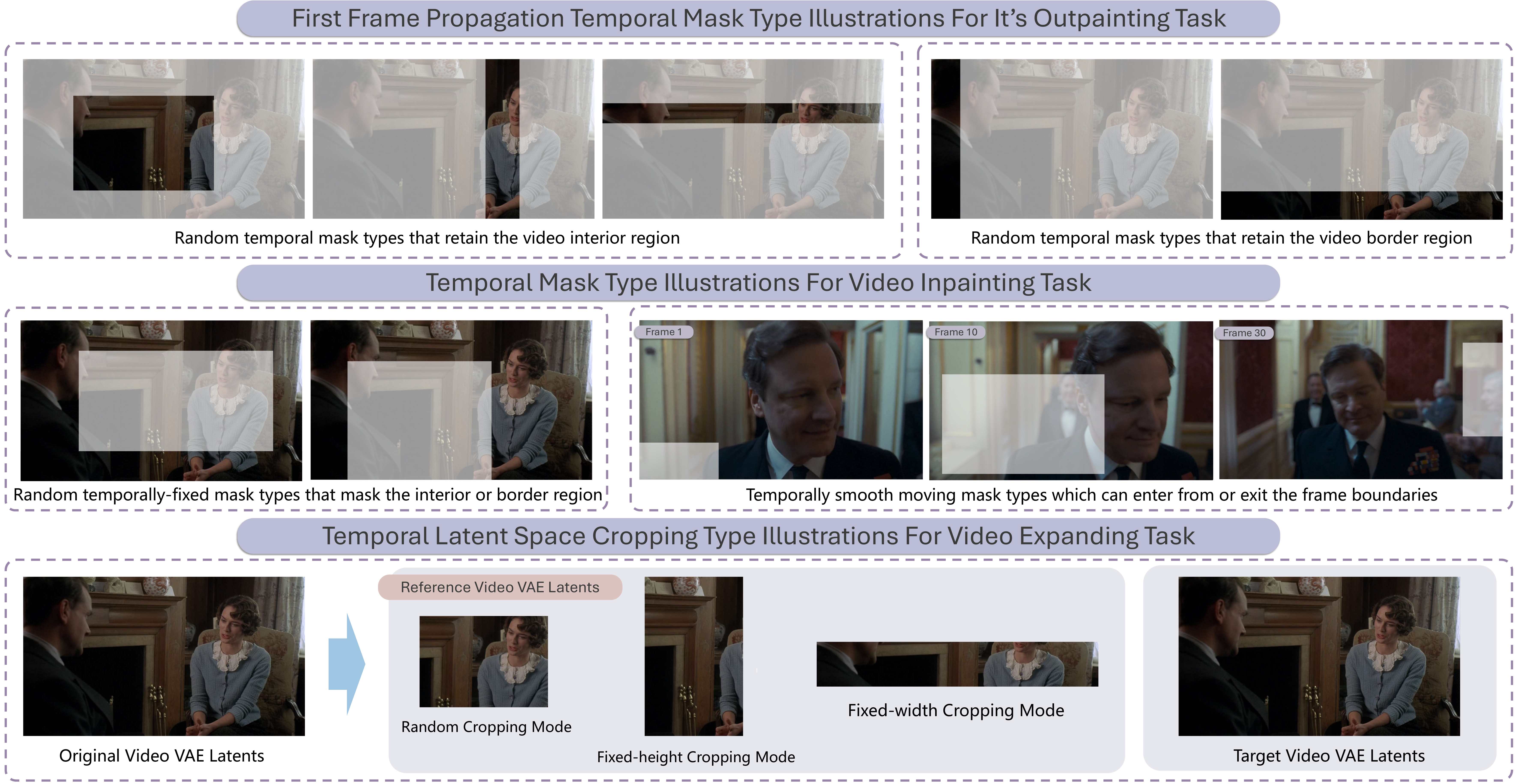} 
 \captionsetup{font=scriptsize}
 \caption{
 \textbf{Temporal mask type illustrations for the outpainting sub-task of the first frame propagation task and the video inpainting task, respectively, and temporal latent space cropping type illustrations for the video expanding task.}
 The light colored regions indicate the portions of the original video frame that are masked. 
 For the outpainting sub-task, temporally-fixed mask type of video inpainting task, and temporal latent space cropping type of video expanding task, we use the first frame here for illustration.
 The masked or cropped region is identical across all frames. 
 For the temporally smooth moving mask type of video inpainting task, we exemplify the corresponding masks for a subset of frames.
 Please refer to \cref{subsec:Miscellaneous editing-task specifics} for details.
 }
 \label{fig:flowmimic_fig_6}
\end{figure*}

\begin{figure*}[htpb]
 \centering
 \includegraphics[width=1\linewidth, trim=0pt 2pt 0pt 2pt, clip]{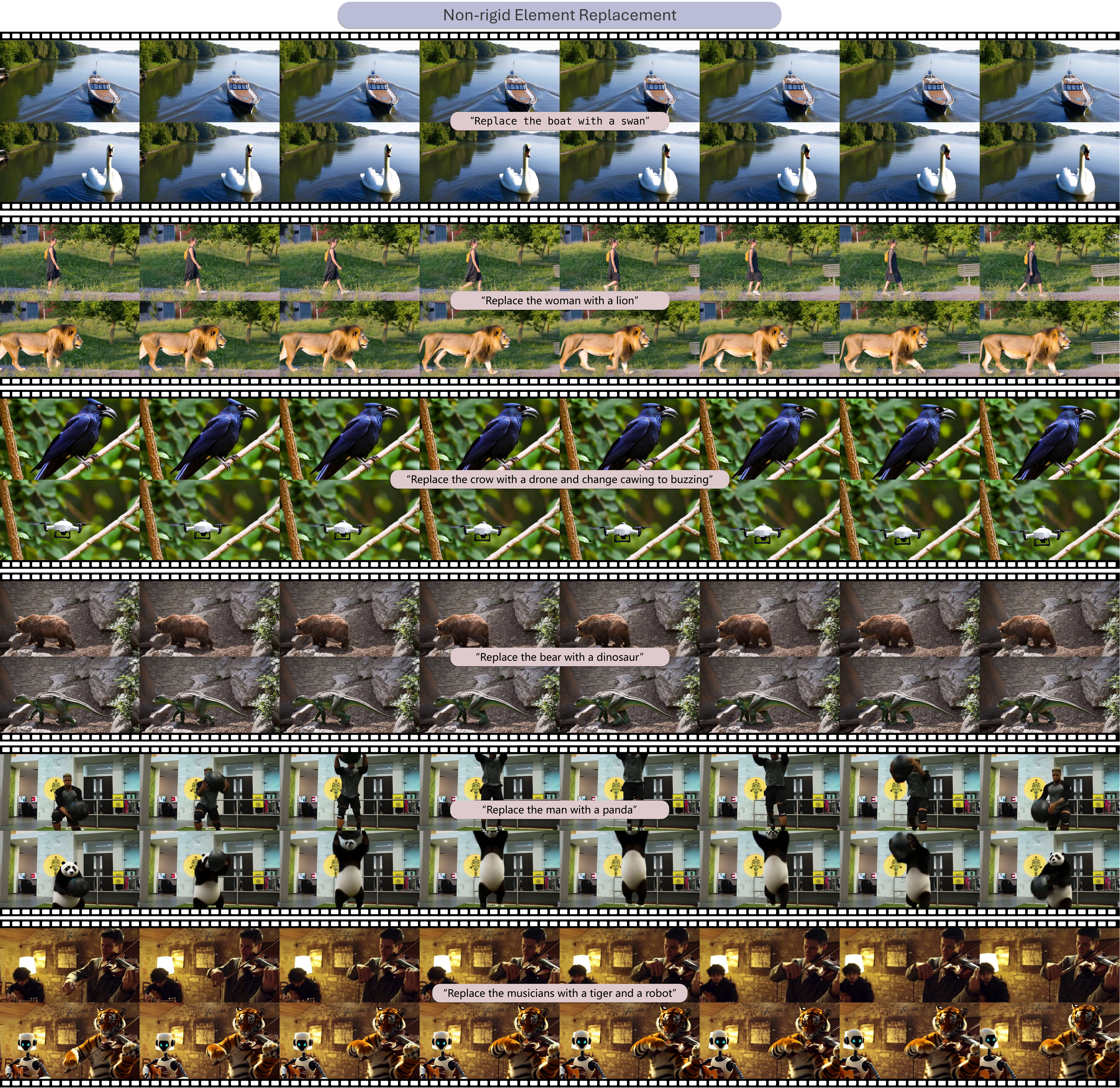} 
 \captionsetup{font=scriptsize}
 \caption{
 \textbf{Single-reference non-rigid element replacement results on FiVE-Bench~\cite{2025five}.}
 Our method yields natural replacements while generating corresponding motion and physical effects, e.g., the walking motion of the lion in the second example and the dinosaur-tail shadow in the fourth. The sixth example demonstrates multi-target editing.
 }
 \label{fig:flowmimic_fig_7}
\end{figure*}

\begin{figure*}[htpb]
 \centering
 \includegraphics[width=1\linewidth, trim=0pt 2pt 0pt 2pt, clip]{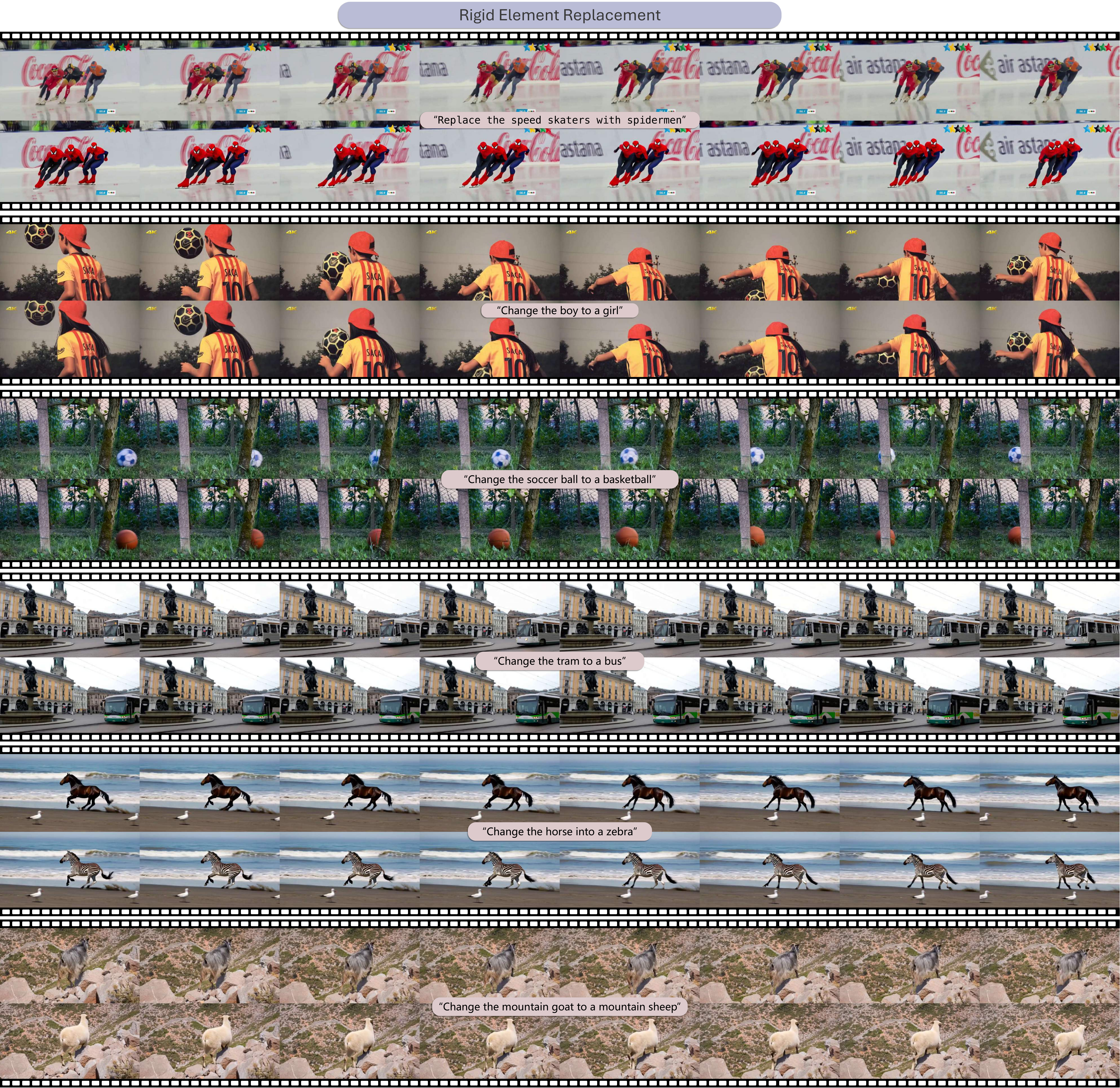} 
 \captionsetup{font=scriptsize}
 \caption{
 \textbf{Single-reference rigid element replacement results on FiVE-Bench~\cite{2025five}.}
 Our editing modifies only the reference region corresponding to the editing instruction, and the edited result exhibits a rigid correspondence with the reference object, as illustrated by the basketball in the third example.
 }
 \label{fig:flowmimic_fig_8}
\end{figure*}

\paragraph{Sense Text Cross Attention Map Loss.}
\label{para:Sense Text Cross Attention Map Loss.}
The sense contour loss $\mathcal{L}_{\text{FM,SC}}$ explicitly steers the model to learn an editing-region-aware editing capability, while the referring expression segmentation task implicitly promotes the alignment between linguistic and visual representations. 
To further explicitly regularise the learning of the latter, we introduce a more direct supervision in the form of a sense text cross attention map loss during the cross-attention process. We begin by briefly discussing the meaning of cross attention maps, which naturally leads to the motivation behind this loss.

The cross-attention conditions the denoising of visual tokens on the text embeddings. 
Specifically, at each cross-attention layer, we utilize the element-wise cross-attention mask to perform the cross-attention procedure separately between the visual tokens of each element in the sample and their corresponding text embeddings. 
The outputs are then concatenated to form the final cross-attention output for the sample.
For each element, the cross-attention procedure first projects the visual tokens and text embeddings into multi-head Query, Key, and Value matrices. 
Let the flattened and patchified visual tokens be \(z_p \in \mathbb{R}^{(f \cdot h_{p} \cdot w_{p}) \times d_z}\) and the text embeddings be $c \in \mathbb{R}^{l_t \times d_c}$, where $f \cdot h_{p} \cdot w_{p}$ is the number of visual tokens, $l_t$ is the length of text embeddings, and $d_z$, $d_c$ are their respective feature dimensions.
For our pretrained T2IV model, $d_z = d_c = n_h \times d_h$.
$n_h$ is the number of attention heads, and $d_h$ is the dimension per head. 
The linear projections yield:
\begin{equation}
Q = z_p W^q,\quad K = c W^k,\quad V = c W^v,
\end{equation}
where \(W^q \in \mathbb{R}^{d_z \times d_z}\), \(W^k, W^v \in \mathbb{R}^{d_c \times d_z}\). 
The projected outputs are then reshaped and the head dimension is separated, resulting in $Q_r$, $K_r$, $V_r \in \mathbb{R}^{n_h \times L \times d_h}$, where $L = f \cdot h_{p} \cdot w_{p}$ for $Q_r$ and $L = l_t$ for $K_r$ and $V_r$. 
The cross-attention layer computes the scaled dot-product attention scores:
\begin{equation}
A = \operatorname{Softmax}\!\left(\frac{Q_r K_r^{\top}}{\sqrt{d_h}}\right) \in [0,1]^{n_h \times (f \cdot h_{p} \cdot w_{p}) \times l_t}.
\end{equation}
$A$ weighted sum over the Value matrix gives the cross-attention output \(O_{\text{cross}} = A V_r\). 
During this process, the cross-attention mechanism “scatters'' textual information across the visual tokens. 
In other words, as described in \cite{xiao2025fastcomposer}, the attention map $A$ connects each visual token to every conditional embedding at each layer, where $A[a_h, i, k]$ quantifies the information flow from the $k$-th text embedding to the $i$-th visual token for $a_h$-th attention head. 
Based on this semantic meaning of the cross-attention score map, in the RES task, when the source visual tokens attend to the text embeddings corresponding to the output caption, the cross attention map associated with the referring expression should exhibit the highest response (and thus the strongest information flow) within the ground-truth object mask, while showing considerably weaker responses outside the mask. 
Therefore, we employ a loss that attenuates the average attention score outside the latent mask while encouraging a higher average score inside it, thereby steering the model's attention to focus more precisely on the visual region described by the editing instruction.

Specifically, as illustrated in \cref{fig:flowmimic_fig_5}, the caption of each sample element, $c_{p,\text{v2v}}=\{c_{p,\text{ref}_1}, c_{p,\text{ref}_2}, \dots, c_{p,\text{ref}_n}, c_{p,\text{output}}\}$, is firstly tokenized by the T5~\cite{raffel2020exploring} tokenizer: ${\text{tok}}_{\text{v2v}}=\{{\text{tok}}_{\text{ref}_1}, {\text{tok}}_{\text{ref}_2}, \dots, {\text{tok}}_{\text{ref}_n}, {\text{tok}}_{\text{output}}\}$. 
This process converts each constituent string into a sequence of subword tokens. 
In the RES task, we identify the subsequence of $\text{tok}_{\text{output}}$ that corresponds to the referring expression. 
Subsequently, the text tokens for each element are mapped to high-dimensional text embeddings by the T5 text encoder and an additional embedding MLP~\cite{wan2025wan}; for the output caption this yields $c_{\text{output}} \in \mathbb{R}^{l_{\text{output}} \times d_c}$, where $l_{\text{output}}$ is the textual sequence length and $d_c$ the embedding dimension. 
The embeddings corresponding to the referring expression maintain their original sequential location within $\text{tok}_{\text{output}}$.

In the cross-attention layer of the $\ell$-th DiT block, where $l \in \{1,2,\dots,N_b\}$ and $N_b$ is the total DiT block count, we first compute the cross-attention score map between the first frame visual tokens of the RES reference element and the text embeddings of the output caption. 
Let the sample visual tokens be denoted as \(z_{p} \in \mathbb{R}^{(2f_p \cdot 2h_p \cdot 2w_p) \times d_z}\) and the concatenated sample caption text embeddings as $c_{\text{v2v}} \in \mathbb{R}^{l_{\text{v2v}} \times d_c}$, where $f_p$, $h_p$ and $w_p$ are the dimensions of the reference patchified visual tokens, $d_z$ and $d_c$ are the visual and textual feature dimensions respectively, and $l_{\text{v2v}}$ is the length of the concatenated text embeddings. 
The projection layers project them into multi-head Query, Key, and Value matrices:
\begin{equation}
 \left\{
\begin{aligned}
Q^{(\ell)} &= z_{p} \, W^{(\ell)}_q, \\
K^{(\ell)} &= c_{\text{v2v}} \, W^{(\ell)}_k, \\
V^{(\ell)} &= c_{\text{v2v}} \, W^{(\ell)}_v,
\end{aligned}
 \right.
\end{equation}
where \(W^{(\ell)}_q \in \mathbb{R}^{d_z \times d_z}\), \(W^{(\ell)}_k, W^{(\ell)}_v \in \mathbb{R}^{d_c \times d_z}\) are the projection weights of the $\ell$-th block. 
As detailed above, the projected outputs are reshaped and the head dimension is separated, yielding \(Q_r^{(\ell)}, K_r^{(\ell)}, V_r^{(\ell)} \in \mathbb{R}^{n_h \times L \times d_h}\) with $L = 2f_p \cdot 2h_p \cdot 2w_p$ for \(Q_r^{(\ell)}\) and $L = l_{\text{v2v}}$ for \(K_r^{(\ell)}\) and \(V_r^{(\ell)}\). 
Denoting the query features corresponding to the first reference frame as \(Q_{r,\text{ref}_1\text{-f}}^{(\ell)} \in \mathbb{R}^{n_h \times (h_p \cdot w_p) \times d_h}\), the key features corresponding to the output caption as \(K_{r,\text{output}}^{(\ell)} \in \mathbb{R}^{n_h \times l_{\text{output}} \times d_h}\), where $l_{\text{output}}$ is the length of output caption's text embbedings .
The scaled dot-product attention scores are then computed as:
\begin{equation}
\begin{aligned}
A_{\text{ref}_1\text{-f}}^{(\ell)} &= \operatorname{Softmax}\!\left(\frac{Q_{r,\text{ref}_1\text{-f}}^{(\ell)} (K_{r,\text{output}}^{(\ell)})^{\top}}{\sqrt{d_h}}\right) \\
& \in [0,1]^{n_h \times (h_p \cdot w_p) \times l_{\text{output}}}.
\end{aligned}
\end{equation}
The resulting \(A_{\text{ref}_1\text{-f}}^{(\ell)}\) is the cross-attention score map for the $\ell$-th block; it quantifies, for each attention head, the association strength between every spatial position of the reference's first frame and every text embedding in the output caption.

Let the start and end indices of the referring expression within the output caption embedding be $s_{\text{ref}}$ and $e_{\text{ref}}$. 
The referring-expression-corresponding average cross-attention score map for the $\ell$-th block and the $j$-th head is obtained by averaging over this specific text embedding span:
\begin{equation}
\bar{A}_{\text{ref}_1\text{-f}}^{(\ell,j)} = \frac{1}{e_{\text{ref}} - s_{\text{ref}} + 1} \sum_{k=s_{\text{ref}}}^{e_{\text{ref}}} A_{\text{ref}_1\text{-f}}^{(\ell)}[j, :, k] \in [0,1]^{h_p \cdot w_p}.
\end{equation}
To obtain a single, consolidated spatial attention map that aggregates information across all blocks and heads, we further average over DiT block and head dimensions:
\begin{equation}
\bar{A}_{\text{ref}_1\text{-f}} = \frac{1}{N_b \cdot n_h} \sum_{\ell=1}^{N_b} \sum_{j=1}^{n_h} \bar{A}_{\text{ref}_1\text{-f}}^{(\ell,j)} \in [0,1]^{h_p \cdot w_p}.
\end{equation}
As discussed in prior studies~\cite{chen2026contextflow,raghu2021vision}, early blocks in transformer architectures typically process spatial and structural information, while deeper blocks handle more abstract semantic concepts. 
Consequently, we here compute the average attention score across all blocks to incorporate attentional information from different semantic levels.
The resulting 2D blocks-average cross-attention score map $\bar{A}_{\text{ref}_1\text{-f}}$ quantifies the average attention strength, aggregated across the entire diffusion process, from the referring expression to each spatial location of the reference's first frame.

The cross-attention score map $\bar{A}_{\text{ref}_1\text{-f}}$ is subsequently unflattened and upsampled to the latent size via bilinear interpolation, yielding $\bar{A}_{\text{ref}_1\text{-f}}^{\text{up}} \in [0,1]^{h_l \times w_l}$, where $h_l = 2h_p$ and $w_l = 2w_p$ denote the upsampled spatial size. 
Using the corresponding ground-truth region mask $\widetilde{\mathcal{M}} = \mathcal{M}_{\text{tar},0} \in \{0,1\}^{h_l \times w_l}$ and its complement $1-\widetilde{\mathcal{M}}$ which serves as the background mask, the upsampled cross attention map is partitioned, and the sense text cross attention map loss for the sample is computed as the difference between the average attention scores in the background and the foreground:
\begin{equation}
\begin{aligned}
\mathcal{L}_{\text{sense, attn}} &= \operatorname{mean}\bigl(\bar{A}_{\text{ref}_1\text{-f}}^{\text{up}} \odot (1 - \widetilde{\mathcal{M}})\bigr) - \operatorname{mean}\bigl(\bar{A}_{\text{ref}_1\text{-f}}^{\text{up}} \odot \widetilde{\mathcal{M}} \bigr) \\
&= \frac{1}{h_l w_l} \sum_{i=1}^{h_l} \sum_{j=1}^{w_l} \bar{A}_{\text{ref}_1\text{-f}}^{\text{up}}[i,j] \odot (1-\widetilde{\mathcal{M}}_{i,j}) \\
&\quad - \frac{1}{h_l w_l} \sum_{i=1}^{h_l} \sum_{j=1}^{w_l} \bar{A}_{\text{ref}_1\text{-f}}^{\text{up}}[i,j] \odot \widetilde{\mathcal{M}}_{i,j}.
\end{aligned}
\end{equation}
This formulation encourages the model to assign higher attention scores within the ground-truth object region and lower scores outside it according to the referring expression, thereby explicitly regularising the cross-attention mechanism to focus on the semantically relevant visual areas.

\begin{figure*}[htpb]
 \centering
 \includegraphics[width=1\linewidth, trim=0pt 2pt 0pt 2pt, clip]{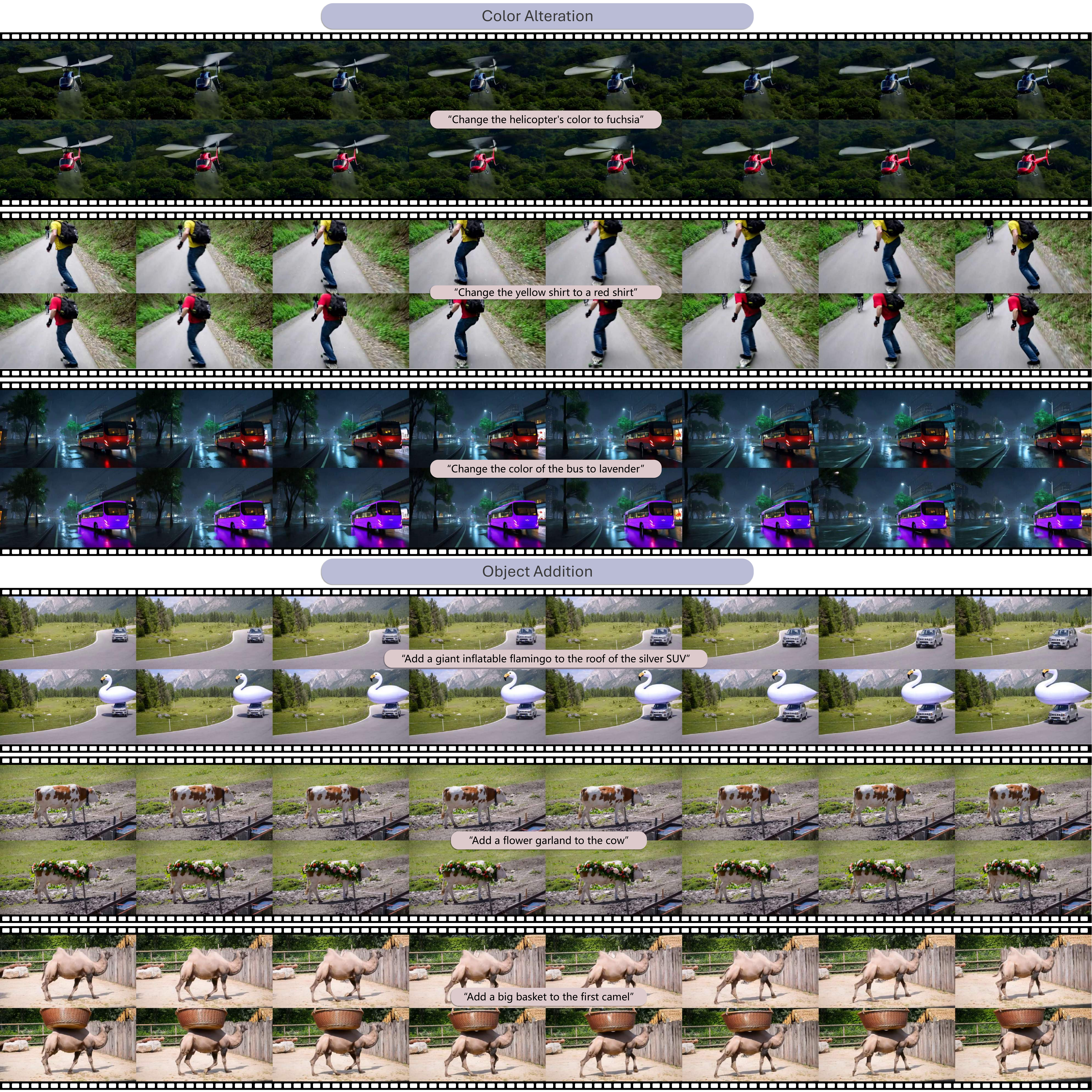} 
 \captionsetup{font=scriptsize}
 \caption{
 \textbf{Single-reference color alteration and object addition results on FiVE-Bench~\cite{2025five}.}
 Our editing results consistently incorporate corresponding physical effects, such as shadows. The second example demonstrates that our model can identify the editing region with occlusions (e.g., a backpack) and modify it exclusively.
 }
 \label{fig:flowmimic_fig_9}
\end{figure*}

\begin{figure*}[htpb]
 \centering
 \includegraphics[width=1\linewidth, trim=0pt 2pt 0pt 2pt, clip]{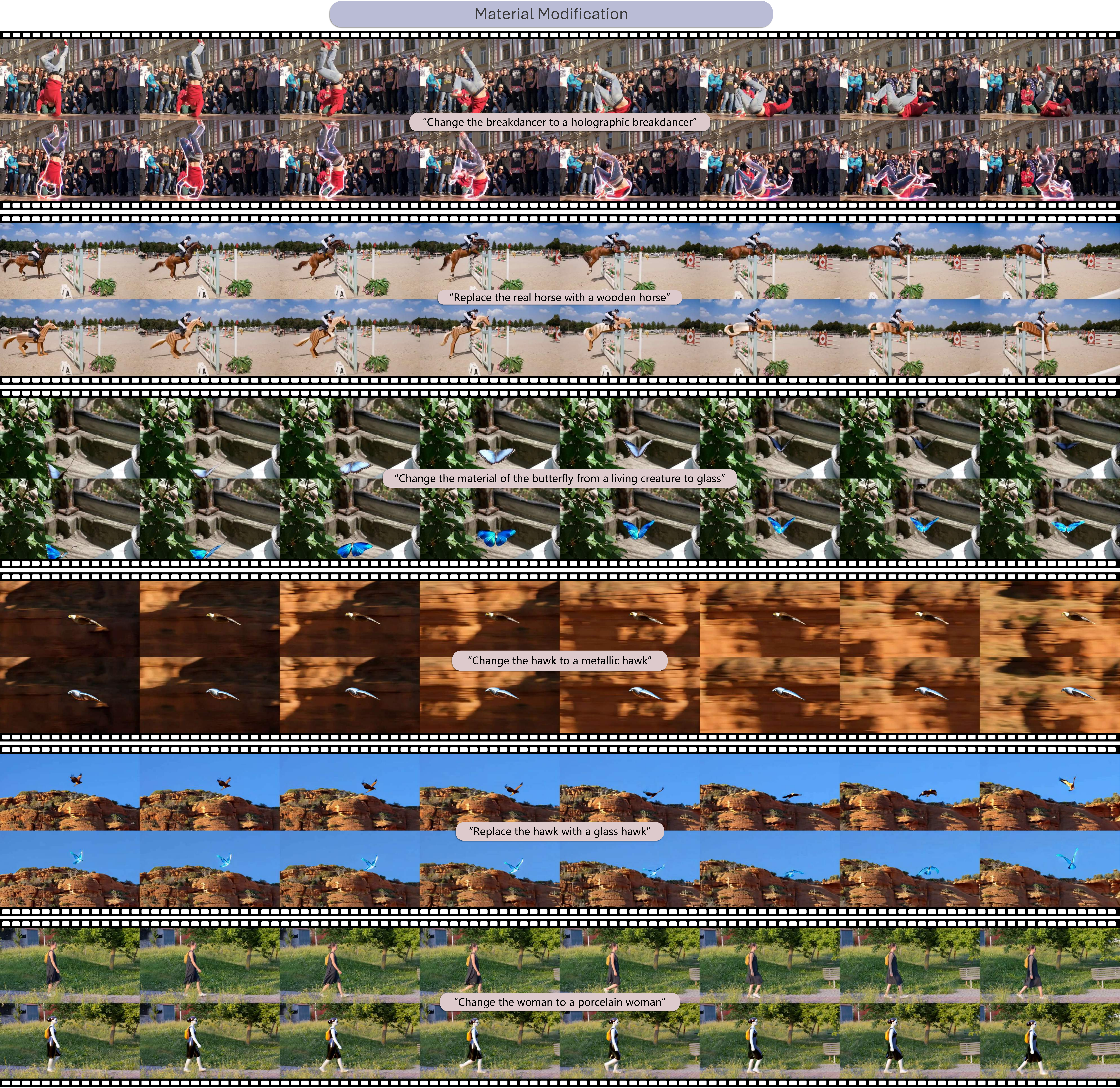} 
 \captionsetup{font=scriptsize}
 \caption{
 \textbf{Single-reference material modification results on FiVE-Bench~\cite{2025five}.}
 Our method maintains pixel-aligned editing results across frames.
 }
 \label{fig:flowmimic_fig_10}
\end{figure*}

\begin{figure*}[htpb]
 \centering
 \includegraphics[width=1\linewidth, trim=0pt 2pt 0pt 2pt, clip]{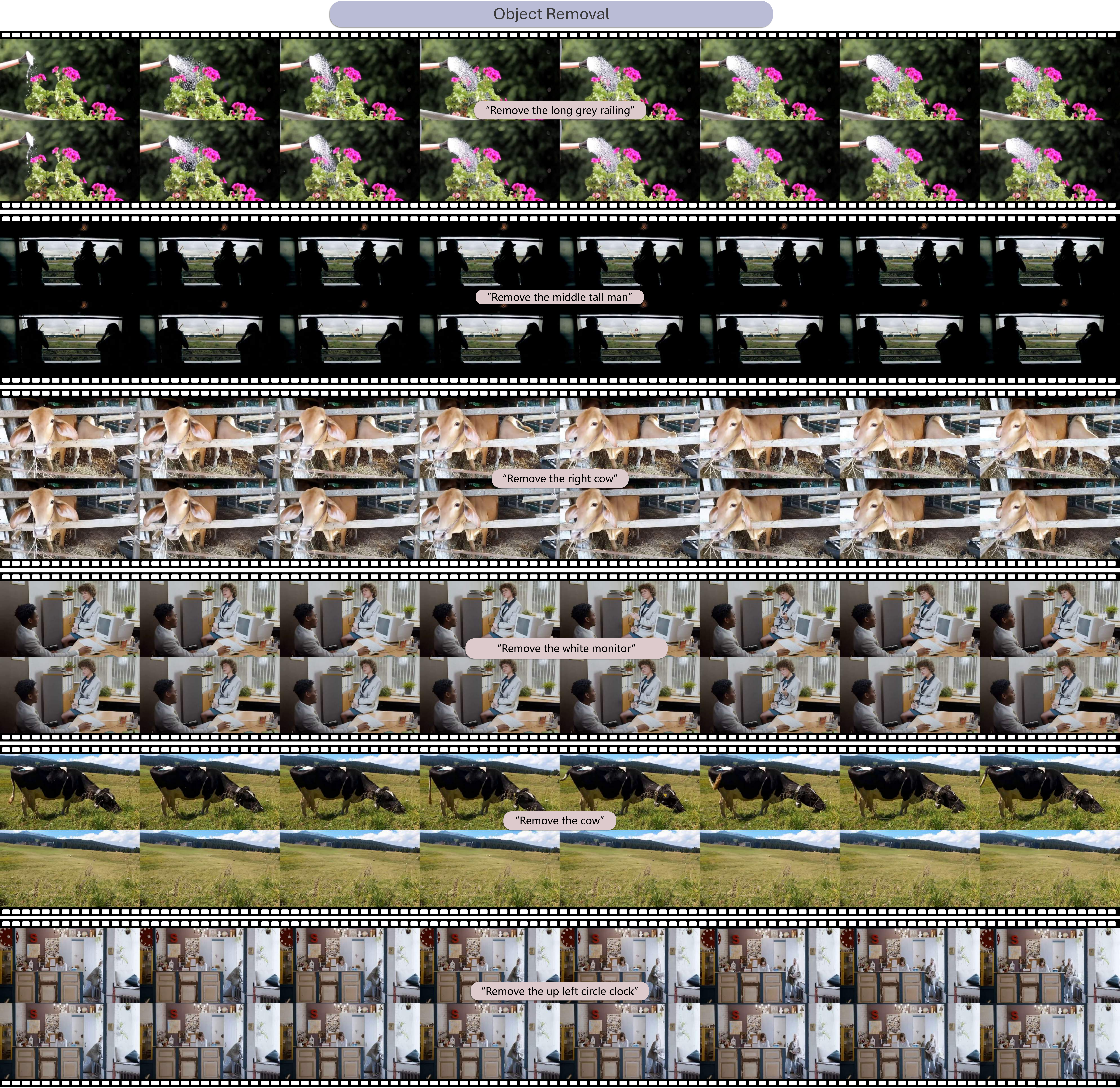} 
 \captionsetup{font=scriptsize}
 \caption{
 \textbf{Object removal results on UNIC-Bench~\cite{ye2025unic}.}
 Unlike previous mainstream approaches~\cite{bian2025videopainter,hu2026} that rely on object mask sequences to help the model locate regions for removal, we employ sense-related tasks and losses, enabling the model to delete specified objects through natural language. The fifth example demonstrates that our model can perceive and remove the associated physical effects, such as shadows, corresponding to the deleted object, whereas our object removal image editing training data lacks this property.
 }
 \label{fig:flowmimic_fig_11}
\end{figure*}

\begin{figure*}[htpb]
 \centering
 \includegraphics[width=1\linewidth, trim=0pt 2pt 0pt 2pt, clip]{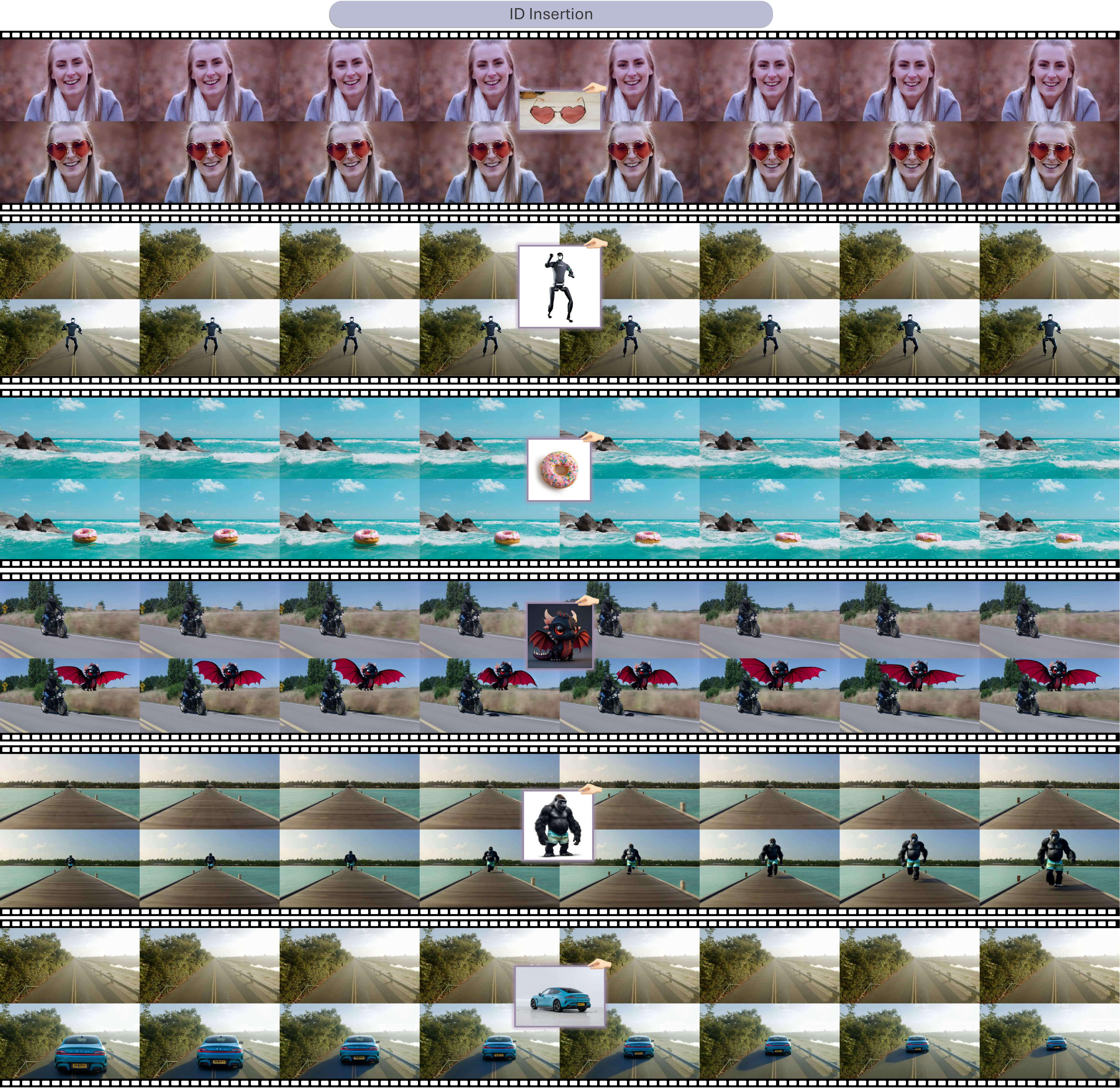} 
 \captionsetup{font=scriptsize}
 \caption{
 \textbf{Multi-reference object insertion results on UNIC-Bench~\cite{ye2025unic}.}
 Given an editing prompt and a reference image, FlowMimic can place the object from the reference image into the specified region of the reference video in a natural, prompt-following manner, while generating corresponding motions. 
 It should be noted that we do not employ any specially constructed video editing data throughout the training process.
 We consider that the model's capability to generate temporally natural motions for the subject based on the caption may arise from FlowMimic implicitly transferring the motion-generation ability—acquired from training on T2V and I2V tasks—to editing tasks such as multi-reference video editing. Moreover, our pixel-wise temporal warped flow field enables the model to learn how to place the subject from a reference image into the source image and maintain it consistently over time. Together, these two abilities allow the model to accurately insert the subject from the reference image into the source video while generating natural and plausible actions for the subject.
 }
 \label{fig:flowmimic_fig_12}
\end{figure*}

\begin{figure*}[htpb]
 \centering
 \includegraphics[width=1\linewidth, trim=0pt 2pt 0pt 2pt, clip]{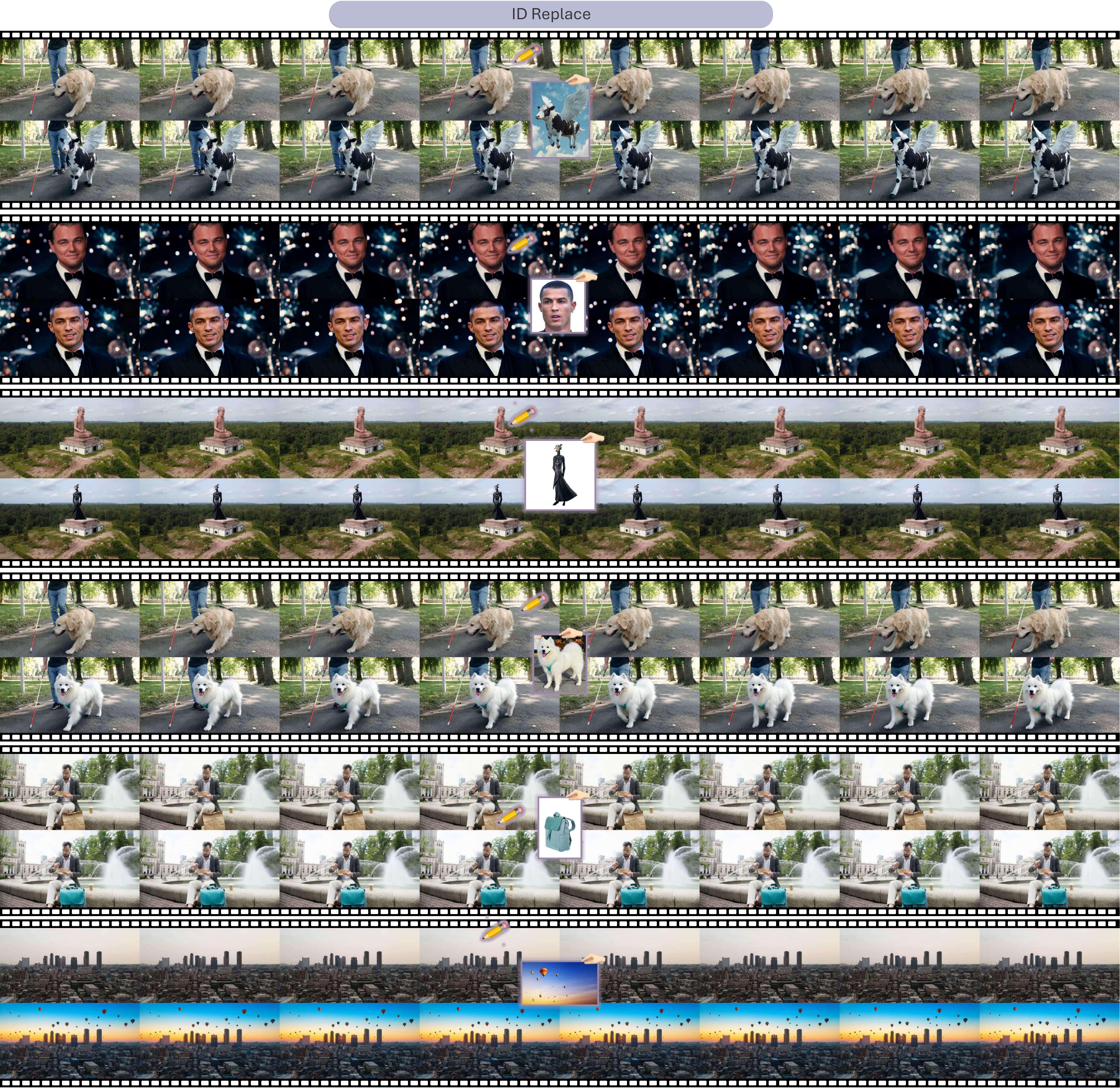} 
 \captionsetup{font=scriptsize}
 \caption{
 \textbf{Multi-reference object replacement results on UNIC-Bench~\cite{ye2025unic}.}
 Given an editing instruction that specifies the subject to be replaced and a reference image, FlowMimic can replace the subject in the reference video with the object from the reference image. When the subject in the reference video is in motion, as in the first and fourth examples, the edited video naturally preserves the corresponding motion. The second example demonstrates that our model retains the original facial expressions when swapping heads. The final example shows that, when altering the background, our method also generates appropriate lighting effects for the remaining regions. The pencil symbol points to the object to be edited, which is specified in the editing caption.
 }
 \label{fig:flowmimic_fig_13}
\end{figure*}

\begin{figure*}[htpb]
 \centering
 \includegraphics[width=1\linewidth, trim=0pt 2pt 0pt 2pt, clip]{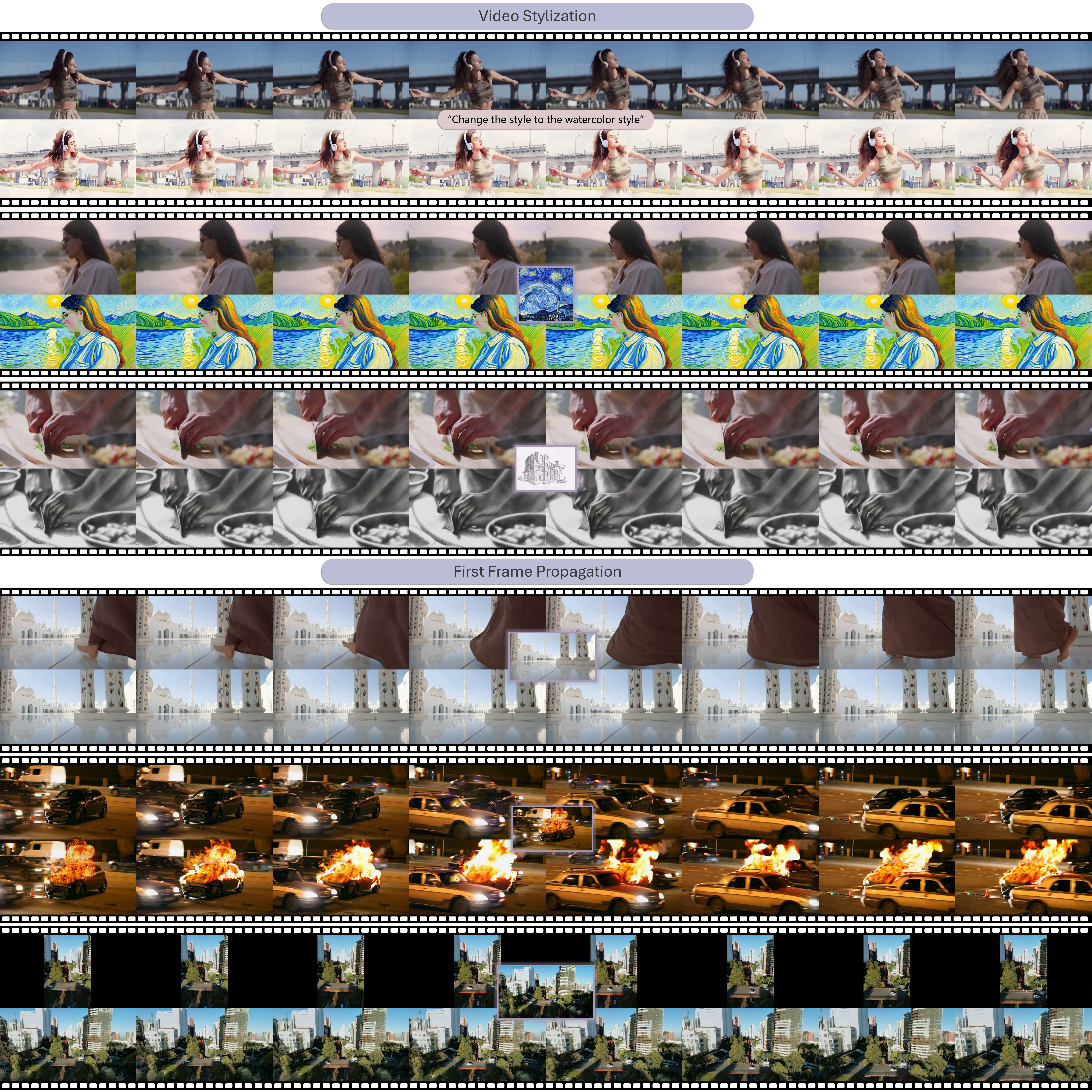} 
 \captionsetup{font=scriptsize}
 \caption{
 \textbf{Stylization and first frame propagation results on UNIC-Bench~\cite{ye2025unic}.} 
 The stylization results demonstrate the model's pixel-wise editing ability, whereas the first frame propagation results show that the editing applied to the initial frame can be propagated to later frames in a natural and physically plausible manner.
 }
 \label{fig:flowmimic_fig_14}
\end{figure*}

\begin{figure*}[htpb]
    \centering
    \includegraphics[width=1\linewidth, trim=0pt 2pt 0pt 2pt, clip]{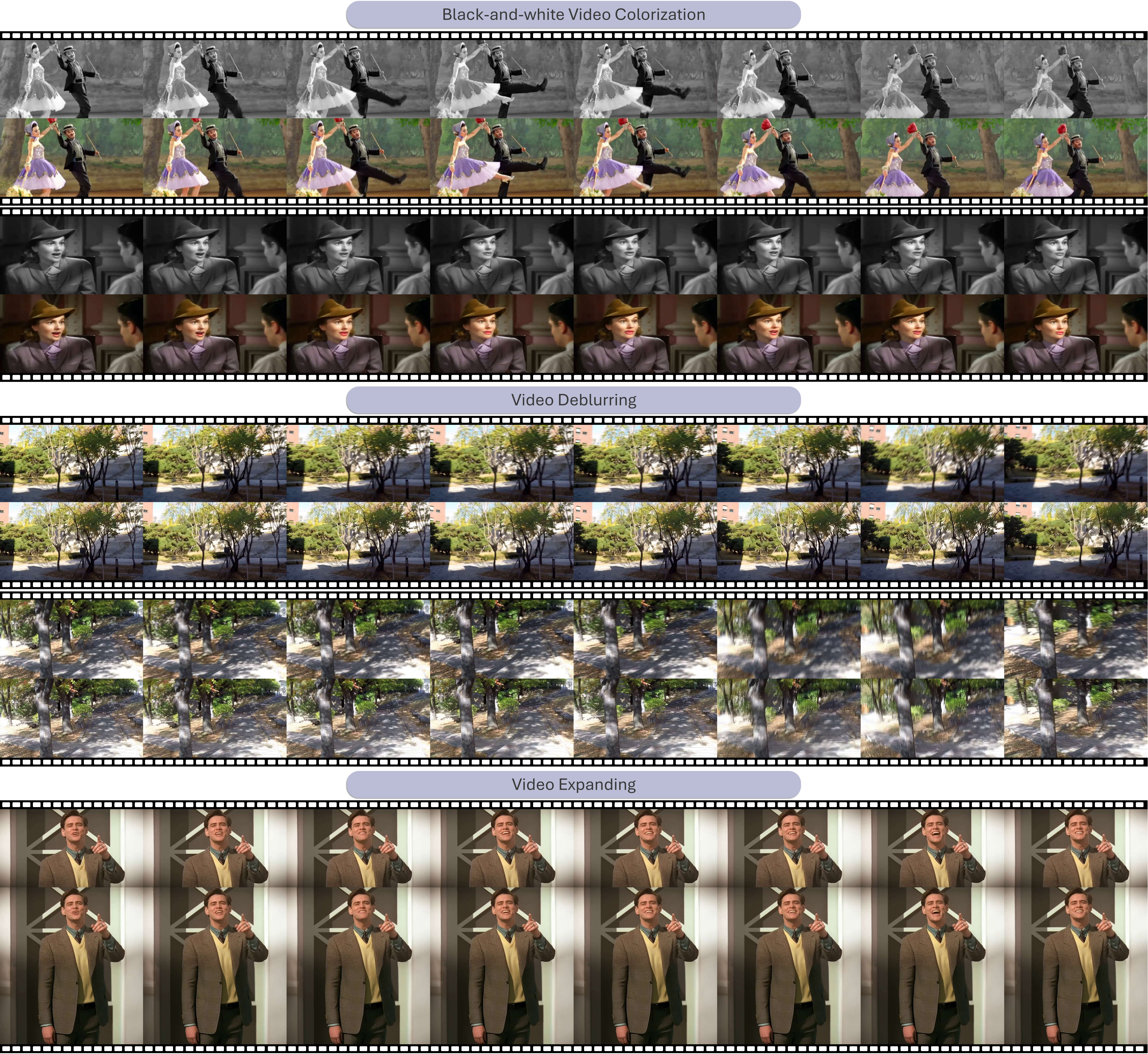} 
    \captionsetup{font=scriptsize}
    \caption{
    \textbf{Single-reference black-and-white video colorization, video deblurring, and video expanding results.}
    The reference videos for deblurring are drawn from the test set of the GOPRO Large~\cite{Nah} dataset.
    }
    \label{fig:flowmimic_fig_15}
\end{figure*}

\begin{figure*}[htpb]
    \centering
    \includegraphics[width=1\linewidth, trim=0pt 2pt 0pt 2pt, clip]{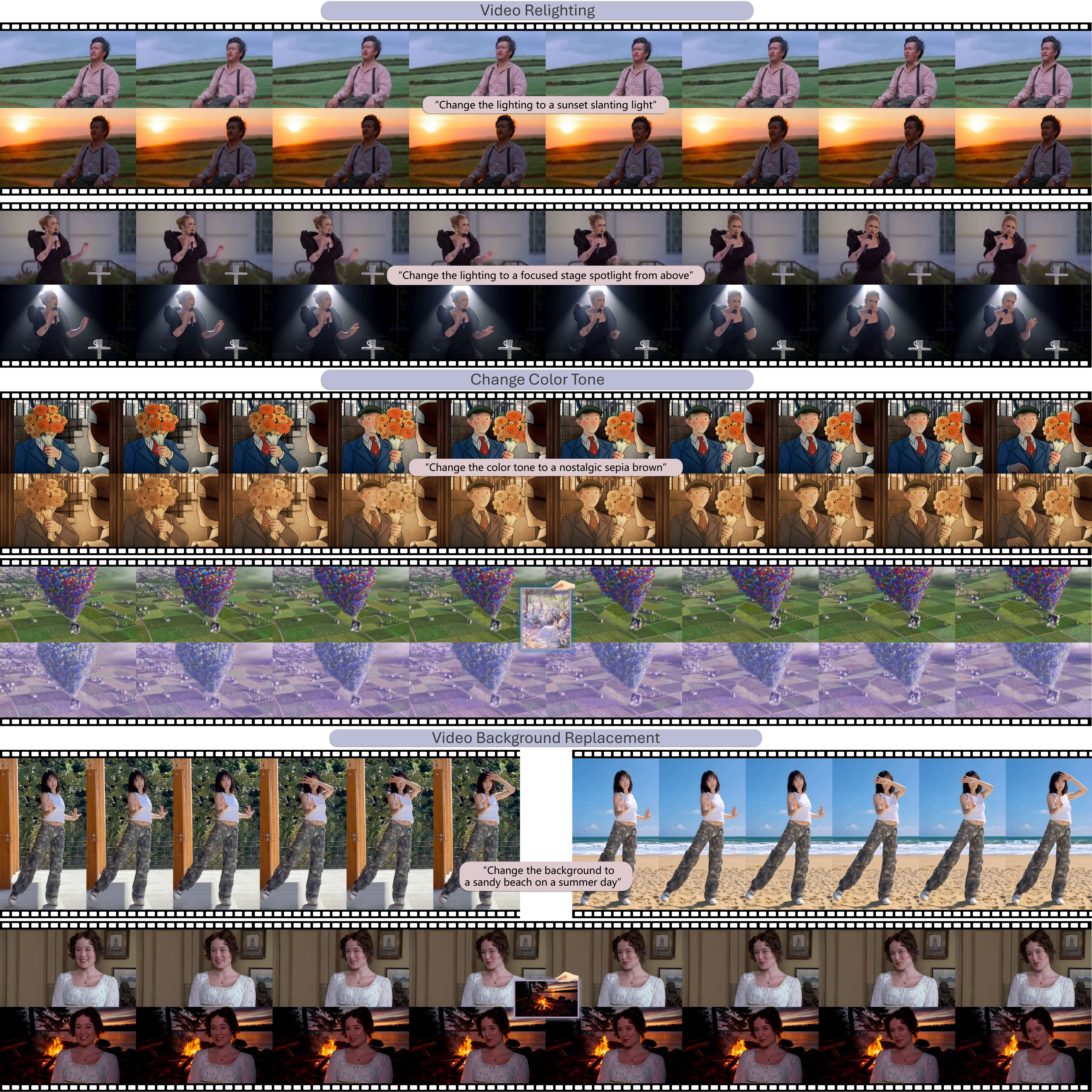} 
    \captionsetup{font=scriptsize}
    \caption{
    \textbf{Video relighting, changing color tone, and video background replacement results.}
    FlowMimic can change the lighting of the original scene and generate plausible physical effects in response to an editing instruction. When altering the color tone, the model faithfully preserves the layout of the original frames. In background replacement, the model also produces natural physical effects based on the editing instruction or an optional reference image, as exemplified by the character's shadow in the fifth example and the flickering of the campfire along with the corresponding changes in character illumination in the sixth. Unlike task-specific methods~\cite{gao2025anyportal}, our model does not rely on foreground mask sequences to distinguish foreground from background; instead, it inherently possesses this semantic capability.
    }
    \label{fig:flowmimic_fig_16}
\end{figure*}

\begin{figure*}[htpb]
    \centering
    \includegraphics[width=1\linewidth, trim=0pt 2pt 0pt 2pt, clip]{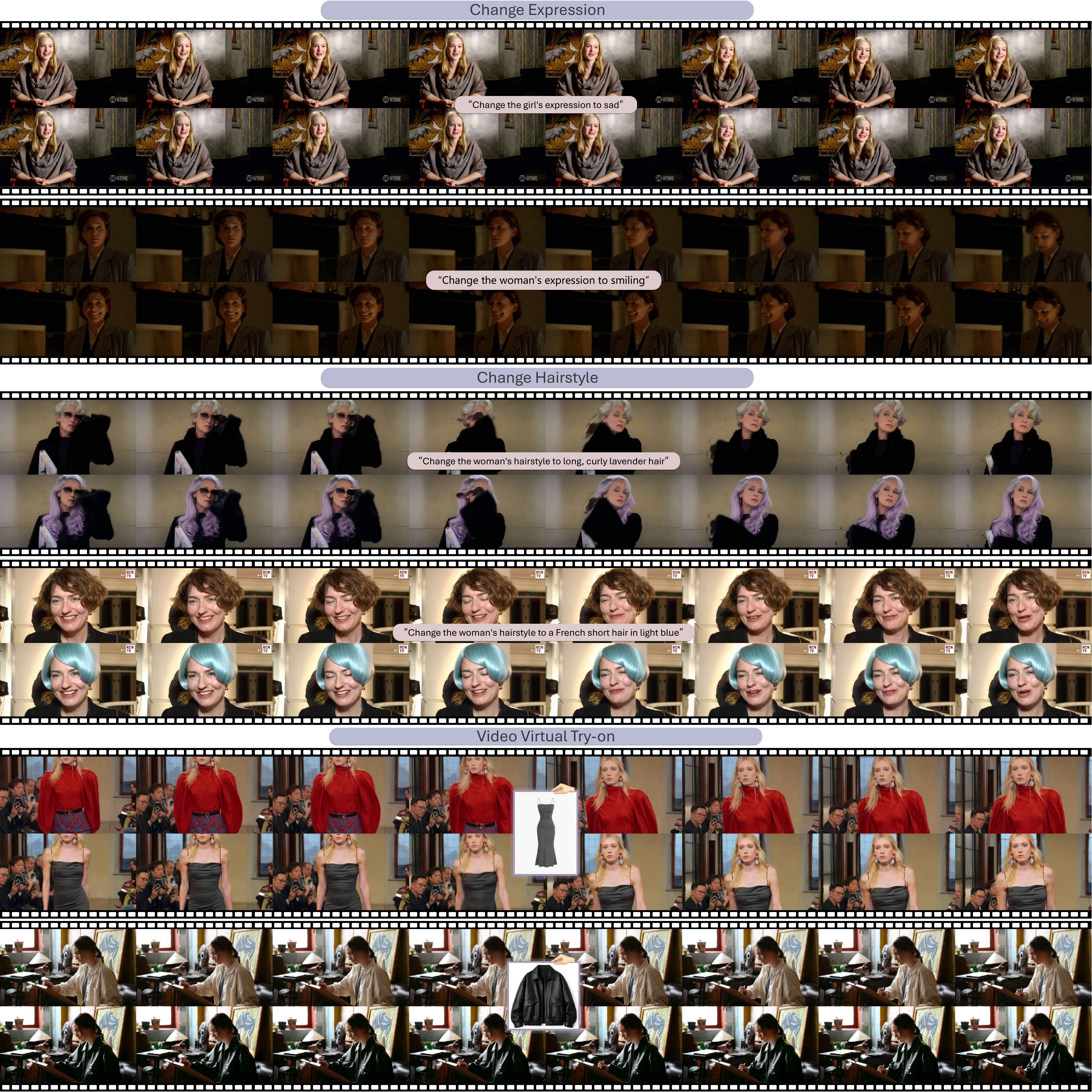} 
    \captionsetup{font=scriptsize}
    \caption{
    \textbf{Changing expression, changing hairstyle, and video virtual try-on results.}
    Example three demonstrates that, even with substantial subject motion, our editing effect remains temporally consistent.
    }
    \label{fig:flowmimic_fig_17}
\end{figure*}

\begin{figure*}[htpb]
    \centering
    \includegraphics[width=1\linewidth, trim=0pt 2pt 0pt 2pt, clip]{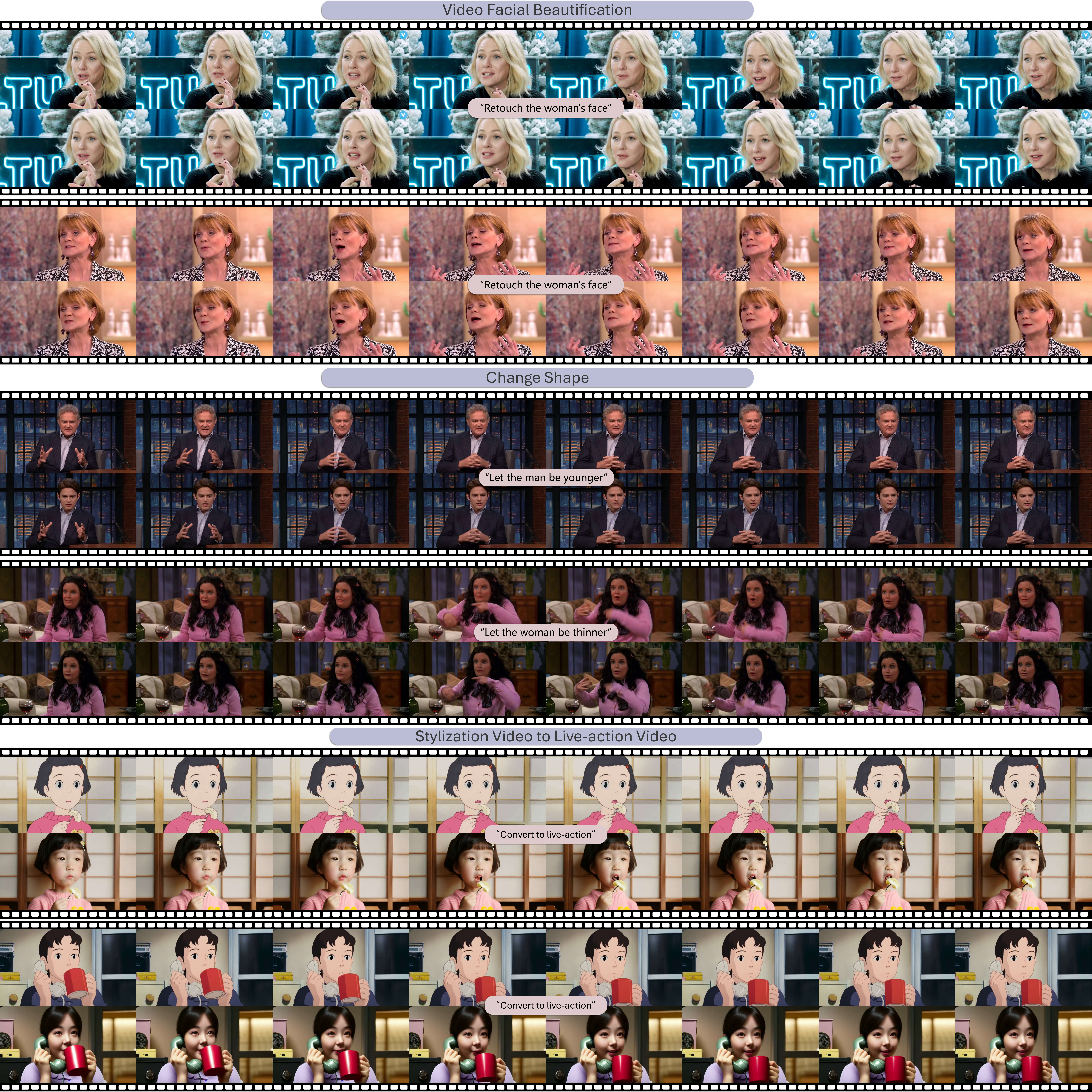} 
    \captionsetup{font=scriptsize}
    \caption{
    \textbf{Video facial beautification, changing shape and stylization video to live-action video results.}
    Here, ``changing shape'' refers to the alteration of the object's objective form, such as its external age state or physical contour. The results of stylization video to live-action video demonstrate both the model's capability to comprehend the semantics of non-photorealistic videos and its ability to preserve pixel-wise editing effects with temporal consistency.
    Some of our image editing samples exhibit notable global color discrepancies between the source and target images, and as a result, the edited outputs inherit these color drifts from the reference inputs due to limitations in the training data. We believe that implementing more rigorous data collection and filtering procedures to exclude samples with color discrepancies will help to mitigate this phenomenon.
    }
    \label{fig:flowmimic_fig_18}
\end{figure*}

\begin{figure*}[htpb]
    \centering
    \includegraphics[width=1\linewidth, trim=0pt 2pt 0pt 2pt, clip]{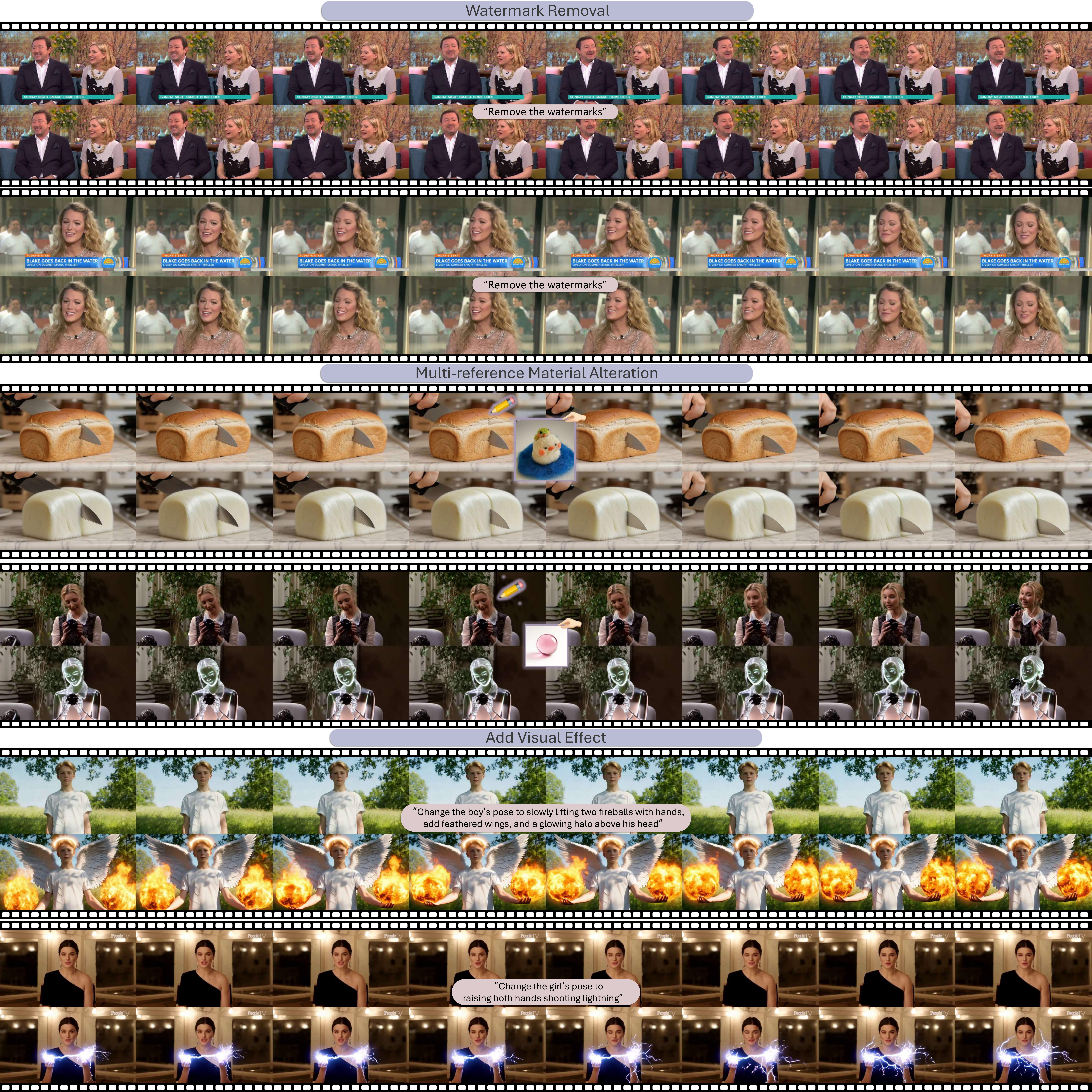} 
    \captionsetup{font=scriptsize}
    \caption{
    \textbf{Watermark removal, multi-reference material alteration and adding visual effect results.}
    In the third and fourth examples, where the editing captions specify the bread and the girl, respectively, as the editing objects, the knife and the cup held by the girl remain unaltered. This demonstrates the model's cross-modality semantic understanding and its region-aware editing ability. Generating visual effects constitutes an interesting application of video editing; despite the absence of dedicated motion-editing data in the training set, the model exhibits a degree of generalization capability.
    }
    \label{fig:flowmimic_fig_19}
\end{figure*}

\begin{figure*}[htpb]
    \centering
    \includegraphics[width=1\linewidth, trim=0pt 2pt 0pt 2pt, clip]{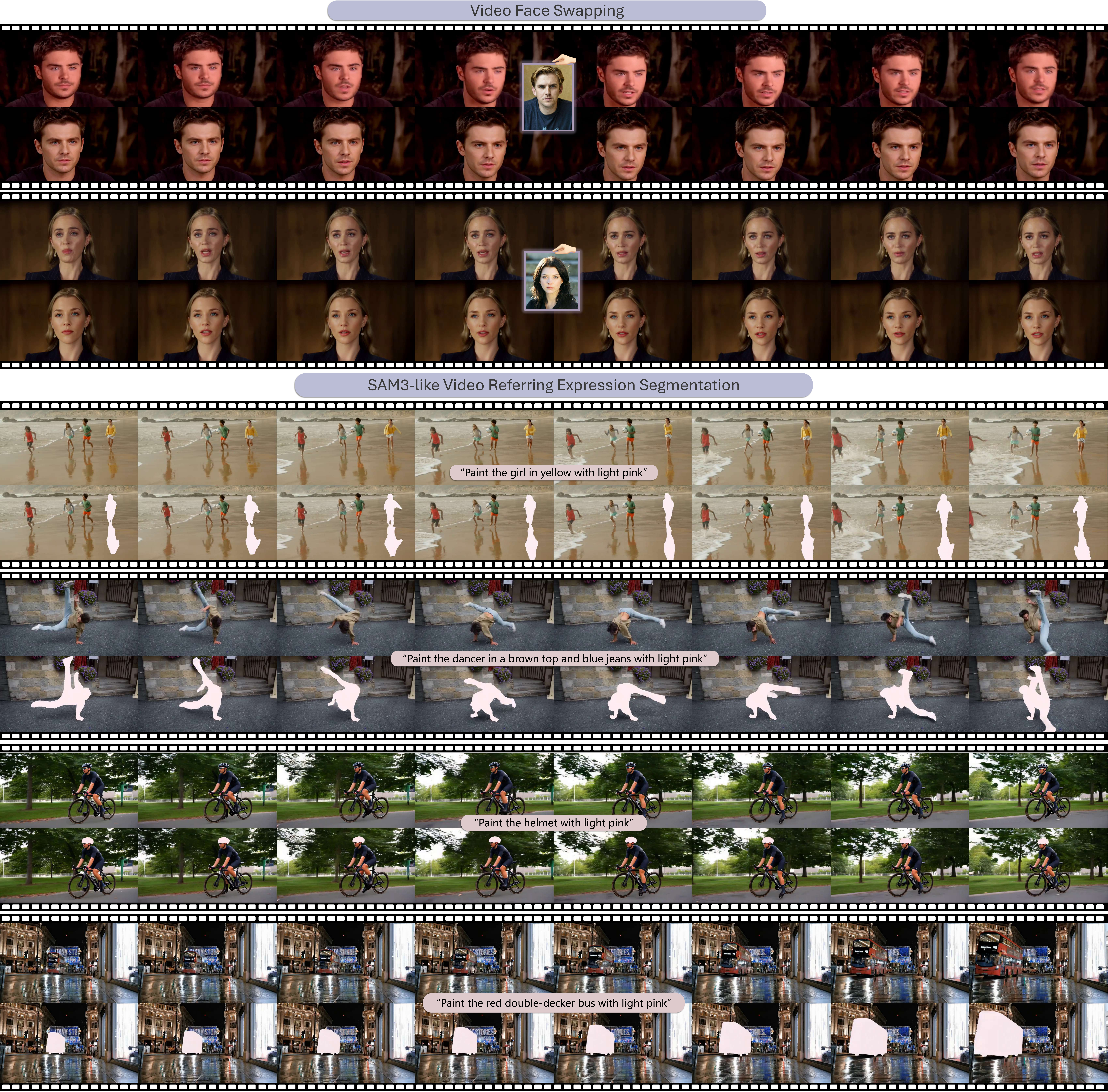} 
    \captionsetup{font=scriptsize}
    \caption{
    \textbf{Video face swapping and SAM3-like video referring expression segmentation results.}
    Our face swapping and head swapping image editing samples each comprise only a few thousand instances and are of relatively low quality. 
    For example, expressions, lip movements, gaze directions, and head poses are often inconsistent before and after the swap; nonetheless, the model demonstrates a degree of generalization capability in preserving lip movements and head poses. 
    Besides, our training data contains no dedicated video editing data; therefore, we have no video referring expression segmentation data or, more broadly, video segmentation data. 
    Solely through our pixel-pair temporal warped flow field, sense-related tasks, and losses, our model achieves visual and language understanding capabilities, as well as temporal tracking abilities. 
    The third example demonstrates that our model can recognize the physical effects of specified objects, such as shadows.
    In fact, our referring expression segmentation image editing samples are derived from the RefCOCO series datasets, resulting in coarse mask contours and the inclusion of noisy data with erroneous mask annotations. 
    We believe that higher-quality training data, such as mask annotations with more refined contours, will yield improved performance.
    }
    \label{fig:flowmimic_fig_20}
\end{figure*}

\begin{figure*}[htpb]
    \centering
    \includegraphics[width=1\linewidth, trim=0pt 2pt 0pt 2pt, clip]{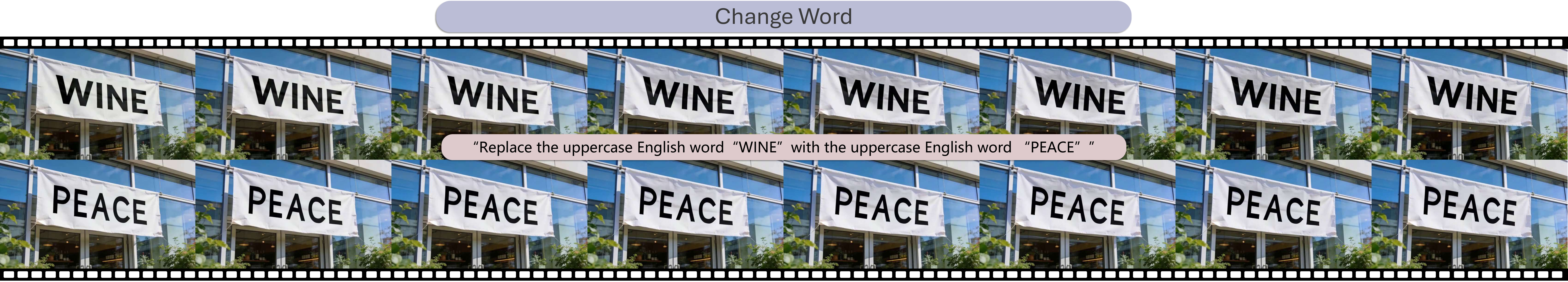} \\ 
    \includegraphics[width=1\linewidth, trim=0pt 2pt 0pt 2pt, clip]{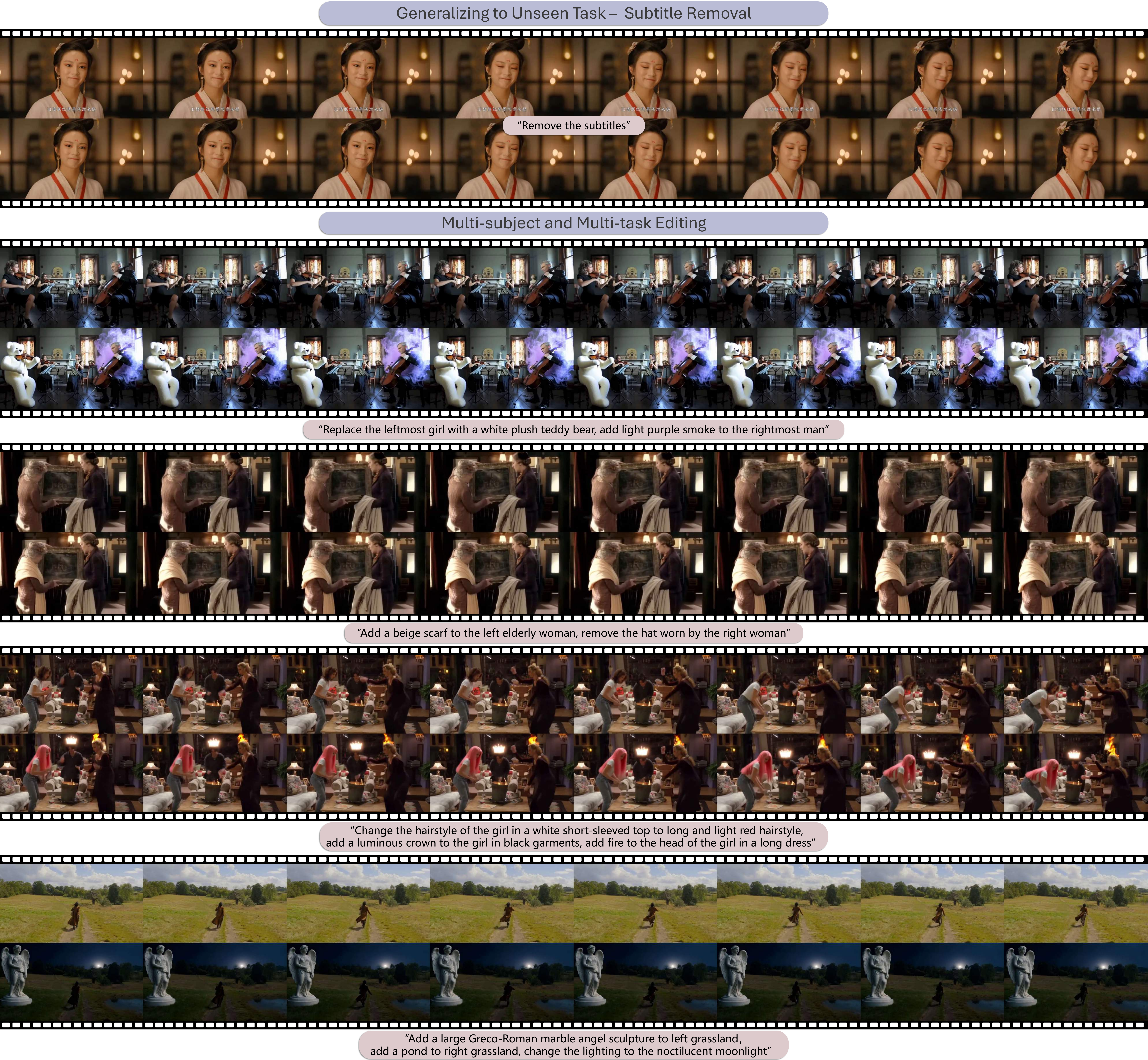}
    \vspace{-1.5em}
    \captionsetup{font=scriptsize}
    \caption{
    \textbf{Multi-subject and multi-task editing results.}
    Although each editing training sample contains only a single-object, single-task editing, when the editing instruction contains multiple objects and various editing tasks at inference time, our model is also capable of identifying each object to be edited and executing the respective editing effects accordingly. 
    This suggests that our model has fundamentally acquired a genuine visual editing capability.
    We also present results for changing words and subtitle removal. With only a tiny weight of the corresponding image and video editing tasks, the model exhibits a degree of capability on this relatively complex, non-rigid video editing task. Besides, although our training samples contain none corresponding to the subtitle removal task, the model demonstrates the ability to remove subtitles.
    }
    \label{fig:flowmimic_fig_21}
    \vspace{-1.3em}
\end{figure*}

\begin{figure*}[htpb]
    \centering
    \includegraphics[width=1\linewidth, trim=0pt 2pt 0pt 2pt, clip]{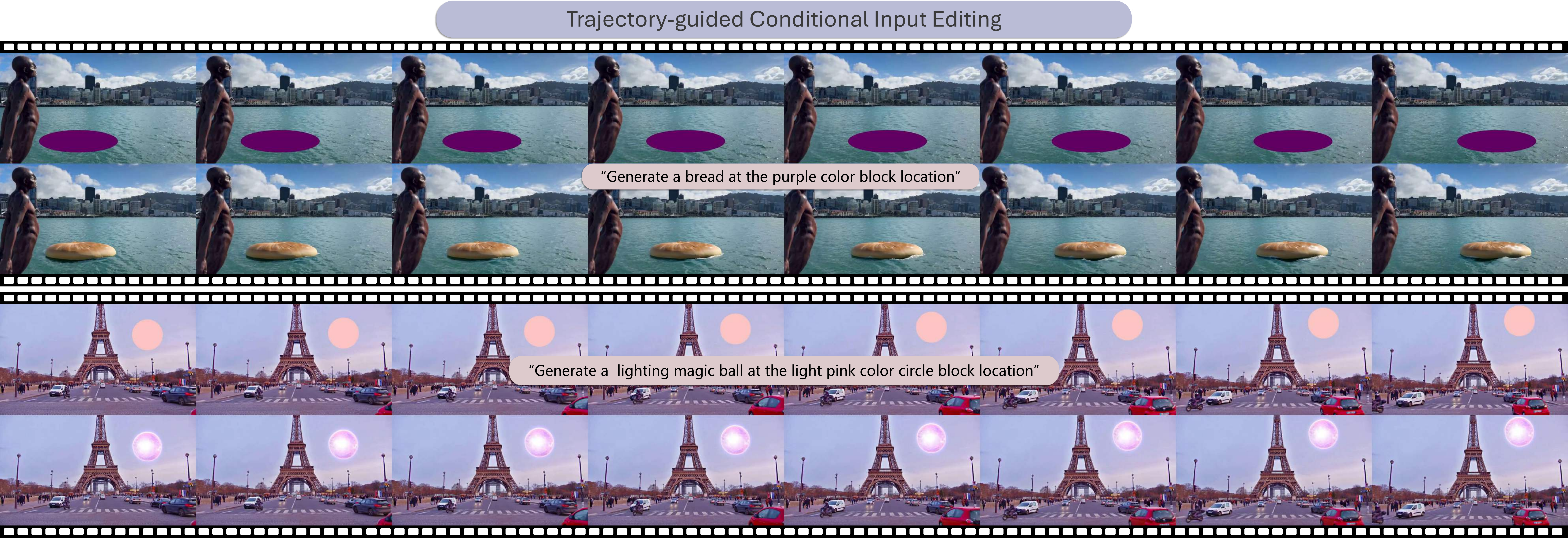} 
    \captionsetup{font=scriptsize}
    \caption{
    \textbf{Trajectory-guided conditional input editing.} Our image editing tasks include both the condition map to image and its inverse operation, among which are approximately 3K image editing pairs—randomly sampled from the RefCOCO dataset—that involve generating a specified object from an image annotated with color-smeared blocks. We recognize that, for video editing tasks, this corresponding capability can give rise to trajectory-conditioned editing. In the first and second examples, for instance, we randomly insert color blocks that traverse in the reference video; this allows the initial position and trajectory of the editing region to be specified precisely and additionally. 
    }
    \label{fig:flowmimic_fig_22}
\end{figure*}

\begin{figure*}[htpb]
    \centering
    \includegraphics[width=1\linewidth, trim=0pt 2pt 0pt 2pt, clip]{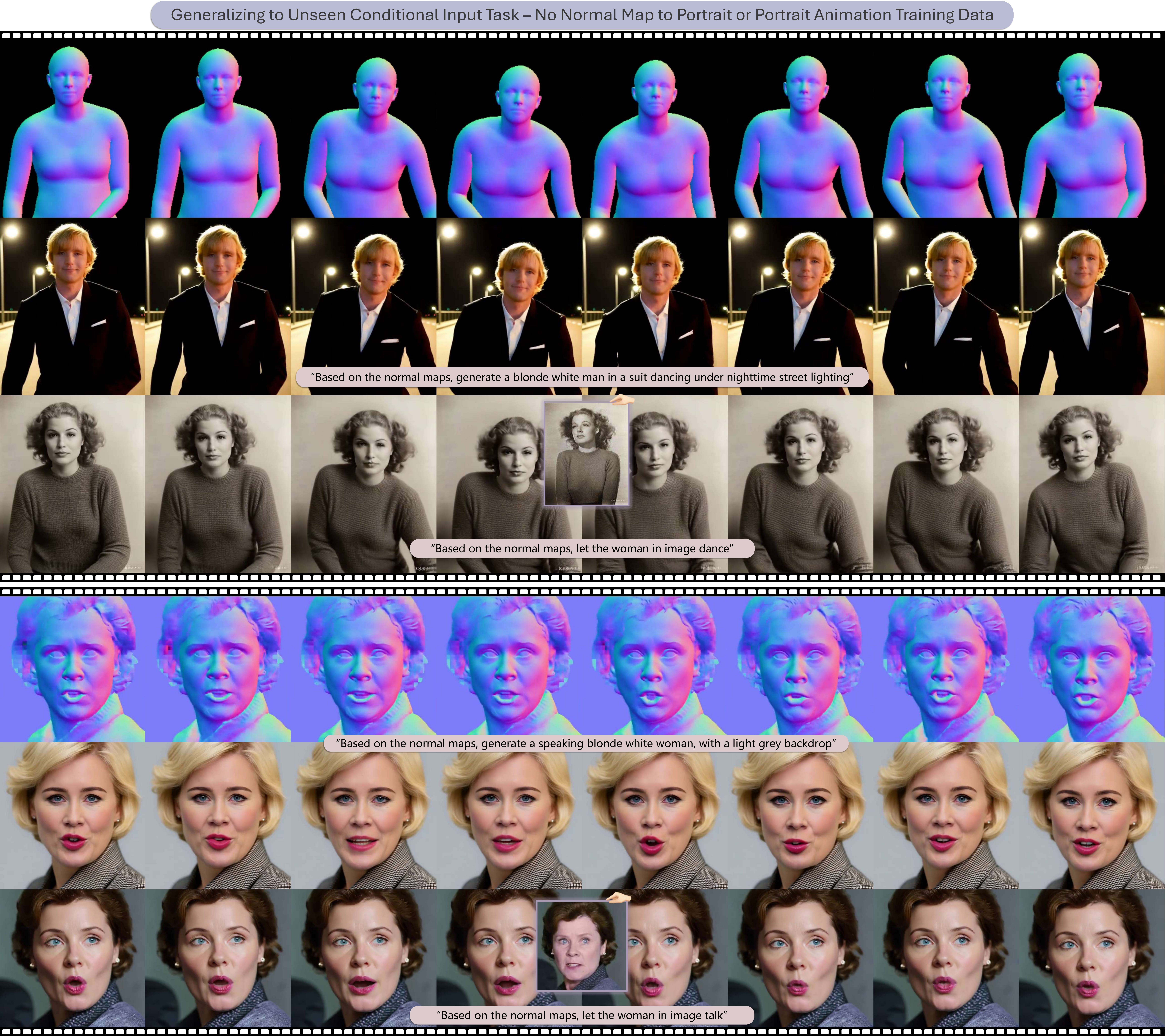} 
    \captionsetup{font=scriptsize}
    \caption{
    \textbf{Generalizing to unseen conditional input task results.} Our training data contain only a small number of samples related to normal maps, all of which are global normal maps of scenes with no data pertaining to bodies or portraits. Nevertheless, for the normal map to image task that involves only the foreground, the model displays some generalization capability. As shown in the second row, provided with a sequence of human normal maps and the corresponding editing caption, we are able to produce a body dance video exhibiting the corresponding motion. Moreover, for the normal map to video editing task that includes a reference image—a task completely unseen during training—the model also demonstrates a degree of generalization.
    }
    \label{fig:flowmimic_fig_23}
\end{figure*}

\begin{figure*}[htpb]
    \centering
    \includegraphics[width=1\linewidth, trim=0pt 2pt 0pt 2pt, clip]{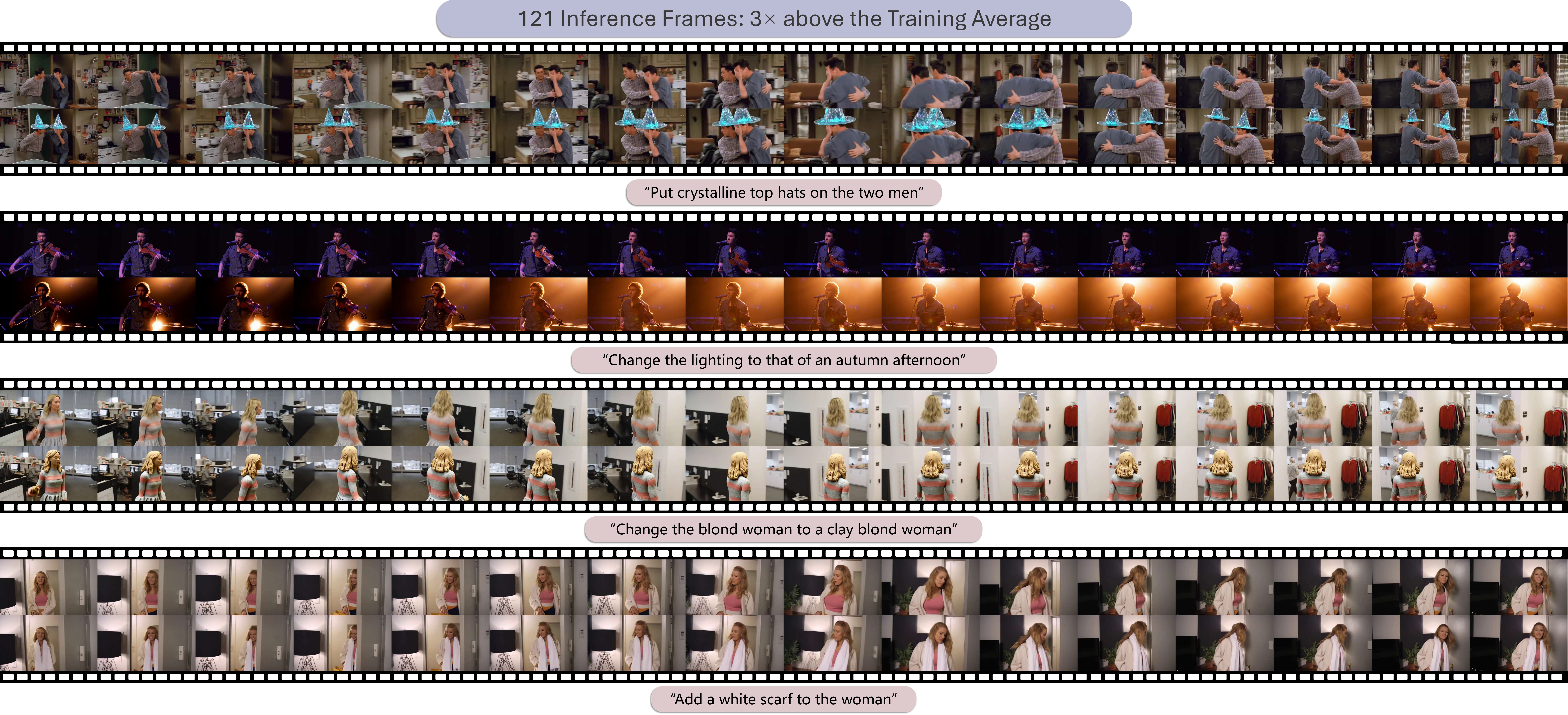} 
    \captionsetup{font=scriptsize}
    \caption{
    \textbf{Temporal generalization to more video frames during inference.}
    During training, to conserve both CPU and GPU memory due to the online generation of video editing data, we generate videos with an average of 37 frames and a maximum of 61 frames. In inference, we investigate whether the model exhibits temporal generalization to longer sequences. Testing on 121 frames—within an acceptable inference time on the mid-range GPU—shows that the model retains a degree of generalization capability for inference lengths beyond the training distribution.
    }
    \label{fig:flowmimic_fig_24}
\end{figure*}

\paragraph{Sense Select Task.}
\label{para:Sense Select Task}
We further incorporate a sense select task to enhance the model's visual comprehension and its text-visual alignment capability.
Specifically, as shown in \cref{fig:flowmimic_fig_5}, we construct each sense select training sample as follows. 
Let $\mathcal{R}$ denote the current cache of the editing reference images. 
We first randomly sample a set of $n$ reference elements from $\mathcal{R}$, denoted as $\{I_{\text{ref}_1}, I_{\text{ref}_2}, \dots, I_{\text{ref}_n}\} \subseteq \mathcal{R}$. 
From this set, one element is uniformly sampled to serve as the target output image, i.e., $I_\text{tar} = I_{\text{ref}_k}$ where \(k \sim \mathcal{U}(\{1, \dots, n\})\). 
The textual captions of all reference elements $\{c_{p,\text{ref}_1}, \dots, c_{p,\text{ref}_n}\}$ are dropped, and the editing instruction is generated via the template “Reconstruct the image that best matches the description: `$\llbracket c_{p,\text{tar}} \rrbracket$'.'', where $c_{p,\text{tar}}$ is the target caption of the selected target image. 
The final sample is therefore represented as the tuple \((\{I_{\text{ref}_1}, I_{\text{ref}_2}, \dots, I_{\text{ref}_n}\}, I_\text{tar})\) with the corresponding output caption, and is used to steer the model to sense and reconstruct the described target from the visual context.

In summary, the overall training objective for the generation step is formulated as:
\begin{equation}
\mathcal{L}_{\text{total, gen}} = \mathcal{L}_{\text{FM,global}} + \alpha_{\text{mimic,gen}} \mathcal{L}_{\text{mimic, gen}},
\label{eq:54}
\end{equation}
where $\alpha_{\text{mimic,gen}}$ denotes the weighting coefficient.
The total objective for the editing step is formulated as:
\begin{equation}
\begin{split}
\mathcal{L}_{\text{total, editing}} &= \mathcal{L}_{\text{FM,global}} + \alpha_{\text{mimic,editing}} \mathcal{L}_{\text{mimic, editing}} \\
&+ \alpha_{\text{FM,SC}} \mathcal{L}_{\text{FM,SC}} + \alpha_{\text{sense,attn}} \mathcal{L}_{\text{sense,attn}},
\end{split}
\end{equation}
where $\alpha_{\text{mimic,editing}}$, $\alpha_{\text{FM,SC}}$ and $\alpha_{\text{sense,attn}}$ are the respective loss weights.
We assign a task ID number to each training sample. 
RES samples are given a distinct number, whereas all other samples receive a default value. This allows us to distinguish RES samples during training and to compute the losses $\mathcal{L}_{\text{FM,SC}}$ and $\mathcal{L}_{\text{sense,attn}}$ solely for them.

\subsection{Model Architecture Adaptation to Editing Task}
\label{subsec:Model Architecture Adaptation to Editing Task}
\paragraph{Reference-inject Self-attention Mask.}
\label{para:Reference-inject Self-attention Mask.}
As shown in \cref{fig:flowmimic_fig_3}, in the editing step, the visual tokens corresponding to the reference and target elements are arranged in an in-context fashion. 
The purpose of the self-attention mechanism is to facilitate information exchange among the visual tokens of a V2V sample. 
A simple way is to preserve the identical intra-sample visual-token interaction as the pretrained T2IV model's 3D full self-attention, whereby the visual tokens corresponding to each element are permitted to attend to both their own visual tokens and the visual tokens of all other elements.
However, the visual tokens of each reference element serve as a visual conditional context that should remain independent and undisturbed by other elements throughout the diffusion process, thereby providing consistent and correct guidance for the target element's learning. 
To this end, as mentioned in \cref{para:Modality Mimic Editing Loss.}, we do not add noise to the reference visual tokens and fix their timesteps at zero.
Moreover, within the self-attention mechanism, we constrain each reference element's visual tokens to attend exclusively to themselves, ensuring that the reference information presented to the model remains intact and uncontaminated throughout the denoising process.

For the visual tokens of the target element, which should learn visual reference information from the reference visual tokens, the self-attention mechanism should ensure that the visual information from every element within the sample flows to the target visual tokens. 
In other words, the target visual tokens are required to attend to the complete set of visual tokens in the sample. 
As illustrated in \cref{fig:flowmimic_fig_3}, we employ a reference-inject self-attention mask to steer the information flow towards reference information injection.

Formally, let the flattened and patchified visual tokens of the complete sample be denoted as $z_p = z_{p, \text{ref}_1} \oplus \dots \oplus z_{p,\text{ref}_n} \oplus z_{p,\text{tar}} \in \mathbb{R}^{l_z \times d_z}$, where $l_z$ is the total sequence length computed as the sum of token lengths across all elements. 
Let the query, key, and value projection matrices be $W_Q, W_K, W_V \in \mathbb{R}^{d_z \times d_z}$. 
The projections for the entire sequence are $Q = z_p W_Q, K = z_p W_K, V = z_p W_V$, with segments $Q_{\text{ref}_i}, K_{\text{ref}_i}, V_{\text{ref}_i}$ and $Q_{\text{tar}}, K_{\text{tar}}$, $V_{\text{tar}}$ corresponding to the $i$-th reference element and the target element, respectively. 
The self-attention with reference-inject attention mask is defined by the following operations:
\begin{equation}
\left\{
\begin{aligned}
O_{\text{ref}_i} &= \operatorname{Softmax}\left( \frac{Q_{\text{ref}_i} K_{\text{ref}_i}^{\top}}{\sqrt{d_h}} \right) V_{\text{ref}_i}, \\[8pt]
O_{\text{tar}} &= \operatorname{Softmax}\left( \frac{Q_{\text{tar}} K^{\top}}{\sqrt{d_h}} \right) V,
\end{aligned}
\right.
\end{equation}
where $O_{\text{ref}_i}$ and $O_{\text{tar}}$ are the self-attention output of the $i$-th reference element and the target element, respectively, and $d_h$ is the head dimension. 
The final self-attention output $O_{\text{self}}$ for the complete sequence is the concatenation of the outputs from all reference elements and the target element:
\begin{equation}
O_{\text{self}} = O_{\text{ref}_1} \oplus O_{\text{ref}_2} \oplus \dots \oplus O_{\text{ref}_n} \oplus O_{\text{tar}} \in \mathbb{R}^{l_z \times d_z}
\end{equation}
where $\oplus$ denotes the concatenation operation along the sequence length dimension. 

\paragraph{Separate 3D RoPE with Frame Correspondence.}
\label{para:Separate 3D RoPE with Frame Correspondence.}
For each training sample, in the self-attention procedure, after the patchified visual tokens are projected into query, key, and value features, and before the scaled dot-product attention scores are computed, each token should be equipped with a relative position encoding that informs the model of its spatio-temporal location within the sample.
The pretrained T2IV model utilizes 3D rotary position encoding (3D RoPE) for this purpose. 
Specifically, 3D RoPE extends the standard RoPE~\cite{su2024roformer} to encode positions in the temporal, height, and width dimensions. 
The feature channels of a visual token $x$ are partitioned into three groups. A dedicated set of rotary angles is computed for each dimension, including the frame index $t_x$, height $h_x$, and width $w_x$. These angle sets are then applied to their respective channel groups via complex plane rotations. Formally, for a visual token $x$, the encoded output $\dot{x}$ is obtained by:
\begin{equation}
\begin{aligned}
\dot{x} &= \operatorname{RoPE}(x, t_x, h_x, w_x) \\ 
 &= \left \langle x^{(t)} \odot e^{i\Theta_t}, \; x^{(h)} \odot e^{i\Theta_h}, \; x^{(w)} \odot e^{i\Theta_w} \right \rangle,
\end{aligned}
\end{equation}
where $x^{(t)}, x^{(h)}, x^{(w)}$ denote the channel groups corresponding to the temporal, height, and width dimensions, $\Theta_t, \Theta_h, \Theta_w$ are the rotary angles derived from the token's coordinates $(t_x, h_x, w_x)$, $\odot$ denotes element-wise multiplication, and $\left \langle \cdot \right \rangle$ denotes concatenation along the channel dimension. This grouped application of rotations ensures that positional information is encoded distinctly per dimension.

For a V2V sample, in the editing step, it is formed by concatenating its constituent elements along the temporal dimension after flattening; a straightforward approach would therefore apply 3D RoPE to the entire concatenated sequence as a single video. 
However, we take the view that the source video and the edited video share a frame-to-frame spatial correspondence. 
Consequently, as shown in \cref{fig:flowmimic_fig_3}, we instead encode the source video tokens and the target video tokens with 3D RoPE in origin-aligned 3D coordinate spaces separately. 
For other reference images, we adopt the perspective suggested in prior work~\cite{chen2025} that video models implicitly treat the initial frame as a conceptual memory buffer that stores visual entities for subsequent reuse during the generation process; consequently, we also encode these reference images as auxiliary first frames of the source video within origin-aligned 3D coordinate spaces separately. 
For other sample types, we consider a T2V sample as a V2V sample with no reference elements, and regard T2I and I2I samples as the single-frame cases of T2V and V2V samples, respectively. 
Moreover, the rotary position encoding for T2V and T2I samples is supposed to be consistent with that of the pretrained T2IV model; hence, the separate 3D RoPE is also applied to these samples.

\paragraph{Element Identity Encoding.}
\label{para:Element Identity Encoding.}
3D RoPE helps the DiT architecture perceive the relative positions of visual tokens within the 3D space of each element. 
We also need to enable the model to distinguish between visual tokens of different elements, especially in multi-reference editing tasks.
During the exploratory phase, we investigated both 1D RoPE applied along the element dimension appended after the 3D RoPE and the conventional absolute positional encoding for representing element identity; the latter is briefly described below.
As shown in \cref{fig:flowmimic_fig_3}, we employ a learnable embedding table with negligible parameters to the flattened visual tokens before applying 3D RoPE, denoted as $E_{\text{elem}} \in \mathbb{R}^{C_{\text{max}} \times d_z}$, where $C_{\text{max}}$ is the maximum number of elements a sample can contain. 
The table is initialized to zero weight to adapt the pretrained T2IV model.
For a given sample $s$ containing $n_s$ elements, we assign a contiguous index to each element according to a fixed convention: the target element is consistently assigned index 0, while the reference elements are assigned indices $1, 2, \dots, n_{s-1}$ . 
The corresponding order index embedding for the entire sample is then retrieved via a lookup operation on the embedding table.
Each retrieved embedding vector \( E_s^{(k)} \in \mathbb{R}^{d_z} \) is subsequently added to visual tokens belonging to the $k$-th element of the sample. 

\paragraph{Reference Position Select Task.}
\label{para:Reference Position Select Task.}
In multi-reference editing tasks, the editing instruction often contains positional descriptors such as “image 1'' or “image 2'' to specify the order of reference elements, e.g., “Add the glasses from image 2 to the man in image 1''.
It is therefore necessary to assist the model in establishing the correspondence between the ordinal terms in the text and the visual tokens of the elements. 
To this end, we introduce a reference position select task. 
Specifically, as shown in \cref{fig:flowmimic_fig_5}, let $\mathcal{R}$ denote the current cache of editing reference images. 
We first randomly sample a set of $n$ reference elements from $\mathcal{R}$, denoted as $\{I_{\text{ref}_1}, I_{\text{ref}_2}, \dots, I_{\text{ref}_n}\} \subseteq \mathcal{R}$. 
From this set, one element is uniformly sampled to serve as the target image, i.e., $I_\text{tar} = I_{\text{ref}_k}$ where \(k \sim \mathcal{U}(\{1, \dots, n\})\). 
The editing instruction is generated via the template, such as “Reconstruct image $\llbracket Ind \rrbracket$.'', where $Ind$ is the order index (e.g., “1'', “2'') of the selected target image within the set. 
The final sample is therefore represented as the tuple \((\{I_{\text{ref}_1}, \dots, I_{\text{ref}_n}\}, I_\text{tar})\) with the corresponding captions, and is used to steer the model to sense and reconstruct the described target from the visual context based on the positional cue.

\subsection{Miscellaneous editing task specifics}
\label{subsec:Miscellaneous editing-task specifics}
\paragraph{First Frame Propagation.}
\label{para:First Frame Propagation.}
First frame propagation (FFP)~\cite{ku2024anyv2v} refers to a video editing paradigm wherein modifications applied solely to the first frame of a video sequence are automatically and coherently extended to all subsequent frames. 
The core objective is to maintain strict temporal consistency—ensuring the visual alterations propagate realistically over time—while preserving the non-editing regions and the original motion dynamics of the source video. 
The approach effectively reframes the complex video editing task into a simpler, two-stage process: (1) performing the desired editing on a single image, i.e., the first frame, with an image editing tool and (2) employing the model to generate a complete, edited video conditioned on this modified first frame and the contextual information from the original video.

We also generate first frame propagation training samples from image editing data pairs via the pixel-pair temporal warped flow field. 
Concretely, let \((I_{\text{ref}_1}, I_{\text{tar}})\) denote the image editing pair of an image editing sample, where $I_{\text{ref}_1}$ is the source image and $I_{\text{tar}}$ is the edited target image. 
In our construction, the source image $I_{\text{ref}_1}$ is retained as the first frame of the source video, i.e., \(V_{\text{ref}_1}^{(0)} = I_{\text{ref}_1}\). 
The target image $I_{\text{tar}}$ is used in two roles: as a conditional reference image (the second element in the sample) and as the first frame of the target video, i.e. \(V_{\text{tar}}^{(0)} = I_{\text{tar}}\). 
The corresponding captions are assigned as follows: the source video is associated with the original source caption $c_{p,\text{ref}_1}$; the second reference image is associated with the original output caption $c_{p,\text{output}}$; and the target video receives an output caption constructed by prepending a template instruction to the original output caption. 
The template instruction takes a form such as: “Referring to the editing result of the first frame of video 1 in image 2, generate a video with image 2 as the first frame and video 1 fully edited''.
The remaining frames of the source and target videos, \(\{V_{\text{ref}_1}^{(t)}\}_{t=1}^{F-1}\) and \(\{V_{\text{tar}}^{(t)}\}_{t=1}^{F-1}\), are synthesized by applying the temporal warped flow field to the source and target first frames.

Furthermore, the propagation test set of UNIC-Bench~\cite{ye2025unic} includes an outpainting task, which involves generating a complete video that starts from a given first frame, conditioned on a reference video with partially masked regions. 
Inspired by this, we incorporate a small portion of such outpainting tasks during training. 
We construct corresponding samples by applying temporally fixed random masks to T2V samples. 
Specifically, within a predefined mask ratio range, e.g., from 0.8 to 0.83, we apply random masks that either retain the interior region of the video by masking the exterior or retain the border region (i.e., the leftmost, bottom, rightmost, or top region) of the video. 
The masked region remains identical for each frame. 
\cref{fig:flowmimic_fig_6} illustrates the types of masks applied to the whole video sequence. 
The masked T2V video is used as the source video, the first frame of the original T2V video serves as the second reference element, and the original video is treated as the target video. 
This process requires no additional data and is performed online.
It should be noted that training samples of the first frame propagation task, the subsequently introduced I2V task, the video inpainting task, and the video expanding task are not included in the sampling process corresponding to the modality mimic editing loss.

\paragraph{I2V Task.}
\label{para:I2V Task.}
In multi-image editing tasks, such as inserting an object from a second reference image into a source video, it is essential to preserve the appearance of the reference object in the edited output and to ensure that it exhibits natural motion. 
To enhance this capability, we introduce an image-to-video (I2V) task. 
Specifically, we construct I2V training samples by taking a T2V training sample $V = \{ V_t \}_{t=0}^{F-1}$ as the target video and using its first frame as the source image. 
Formally, the corresponding I2V sample is defined as the pair \((I, V)\), where $I = V_0$.
The corresponding sample element captions are the first frame caption and the video caption, where the latter comprises the first frame caption together with a motion caption that describes both the subjects' motion and the camera movements within the video.
Similarly, one could construct first-last frame-to-video (FLF2V) samples from T2V training samples. Since the motivation for this task is analogous to that of the I2V task, we retain only the I2V task.

\paragraph{Video Inpainting Task.}
\label{para:Video Inpainting Task.}
For the video object removal task~\cite{quan2024deep}, once the model has sensed the visual region corresponding to the editing instruction, it should possess the latent capability to ``imagine'' and generate the content that was occluded by the removed static or dynamic object. 
To enhance this capability, we introduce a generalized video inpainting task. 
Specifically, as shown in \cref{fig:flowmimic_fig_6}, for a T2V video sample in the original pixel space, we occlude the video with a fixed mask or a mask that moves randomly over time to create a reference video, and use the original T2V sample as the target video. 
Formally, let $V \in \mathbb{R}^{F \times H \times W \times 3}$ denote the original video. 
We define a binary occlusion mask $M \in \{0,1\}^{F \times H \times W}$ that is zero in the regions to be inpainted. 
The reference video is then constructed as $\widetilde{V} = V \odot M$, where $\odot$ denotes element-wise multiplication. 
The model is trained to generate the occluded region from the random masked video input $\widetilde{V}$, thereby learning to ``imagine'' the missing spatio-temporal regions naturally and sensibly.

We take two masking strategies as examples: the fixed random rectangle mask and the moving random rectangle mask.
For the fixed random rectangle mask, which aims to simulate the static object, a rectangular region of random width $w$ and height $h$ is selected at a random position \( (x, y) \) within the frame. 
The same rectangle is set to zero (black) in every frame of the video. 
The mask is therefore static in time:
$$
 M_t(i,j) = \begin{cases}
 0 & \text{if } x \leq j < x+w \text{ and } y \leq i < y+h, \\
 1 & \text{otherwise},
 \end{cases}
$$
where $t \in \{0,\dots,F-1\}$.
For the moving random rectangle mask, which aims to simulate the motion of a moving object, such as a walking lion, a rectangle of random width and height \( (w, h) \) is initialized at a random starting position \( (x_0, y_0) \). 
It is then assigned a constant velocity \( (v_x, v_y) \) . 
The rectangle's position at frame $t$ is \( (x_t, y_t) = (x_0 + t v_x, \; y_0 + t v_y) \). 
The mask at each frame is zero inside the moving rectangle:
$$
 M_t(i,j) = \begin{cases}
 0 & \text{if } x_t \leq j < x_t+w \text{ and } y_t \leq i < y_t+h, \\
 1 & \text{otherwise}.
 \end{cases}
$$
In practice, masks can be positioned arbitrarily, such as a border mask that statically occludes the frame periphery, or a moving mask that is only partially visible in the first frame.
The training pair is then \( (\widetilde{V}, V) \) with the template instruction such as “Generate the invisible black block region of video 1''. 
In contrast to the conventional definition of the video inpainting task, our masks are randomly generated. 
Actually, this design aims to enhance the model's ability to plausibly generate content in occluded regions. 
Consequently, it has no need for additional cost on annotating masks that target specific objects to be removed.

\paragraph{Video Expanding Task.}
\label{para:Video Outpainting Task.}
Video outpainting~\cite{yu2025unboxed} extends a video sequence beyond its original spatial boundaries by synthesizing plausible content in the missing border regions. 
For instance, it can transform a vertically framed (e.g., 9:16) social-media clip captured on smartphones into a widescreen cinematic (e.g., 16:9) aspect ratio. 
We consider that this task also strengthens the model's ability to preserve existing visual tokens while naturally generating contextually coherent extensions—a form of imaginative synthesis—and therefore we incorporate it as a training task.
The prevailing approaches typically mask the border regions of video samples as source video during training, as described in the first frame propagation's outpainting task. 
We explore a direct method for constructing video outpainting training samples via a random cropping operation applied to the VAE latents of a T2V sample. 
We term this video outpainting task, which is distinct from the conventional definition, as \textit{video expanding}.

Specifically, as shown in \cref{fig:flowmimic_fig_6}, for a given T2V sample, we first extract its VAE latent representation. 
Let the original latent be $z_l \in \mathbb{R}^{f_l \times h_l \times w_l \times C}$ , where $f_l$ denotes the number of latent frames, $h_l$ and $w_l$ are the height and width of the latent spatial dimensions, and $C$ is the number of channels.
We define a random cropping function \( \text{Crop}(\cdot) \) that extracts a sub-region from $z_l$ to produce a cropped latent $z_{l,\text{crop}} \in \mathbb{R}^{f_l \times h'_l \times w'_l \times C}$ , with $h'_l \leq h_l$ and $w'_l \leq w_l$. 

The cropping follows one of three modes.
For random cropping mode, the cropping height $h'_l$ and width $w'_l$ are sampled uniformly as: 
\begin{equation}
h'_l = \big\lfloor \eta_h h_l \big\rfloor, \quad w'_l = \big\lfloor \eta_w w_l \big\rfloor,
\end{equation} 
where \( \eta_h, \eta_w \sim \mathcal{U}(\eta_{\min}, \eta_{\max}) \) with $\eta_{\min}, \eta_{\max}$ as default values. 
The top-left corner \((i, j)\) is chosen uniformly such that $i \in \{0,1,\dots,h_l-h'_l\}$ and $j \in \{0,1,\dots,w_l-w'_l\}$. 
The cropped latent is then $z_{l,\text{crop}} = z_l[:, i:i+h'_l, j:j+w'_l, :].$ 
For fixed-height cropping mode, the cropping width $w'_l$ is sampled as: 
\begin{equation}
w'_l = \big\lfloor \xi w_l \big\rfloor, \quad \xi \sim \mathcal{U}(\xi_{\min}, \xi_{\max}),
\end{equation} 
and the cropping is centred horizontally, i.e. 
\begin{equation}
j = \big\lfloor \frac{w_l - w'_l}{2} \big\rfloor.
\end{equation}
The vertical extent covers the full height, so $z_{l,\text{crop}} = z_l[:, :, j:j+w'_l, :].$ 
For fixed-width cropping mode, the cropping height $h'$ is sampled as: 
\begin{equation}
h' = \big\lfloor \zeta H \big\rfloor, \quad \zeta \sim \mathcal{U}(\zeta_{\min}, \zeta_{\max}),
\end{equation} 
and the cropping is centred vertically, i.e. 
\begin{equation}
i = \big\lfloor \frac{h_l - h'_l}{2} \big\rfloor.
\end{equation} 
The horizontal extent covers the full width, giving $z_{l,\text{crop}} = z_l[:, i:i+h'_l, :, :]$.

The cropped latent $z_{l,\text{crop}}$ serves as the reference element, while the original full latent $z_l$ is retained as the target. 
Besides, the template editing caption takes the form such as ``Expand the image 1, keep the content of image 1 unchanged''.
The model is thereby trained to generate the complete video content from a partially observed region.

\begin{figure*}[htpb]
    \centering
    \includegraphics[width=1\linewidth, trim=0pt 2pt 0pt 2pt, clip]{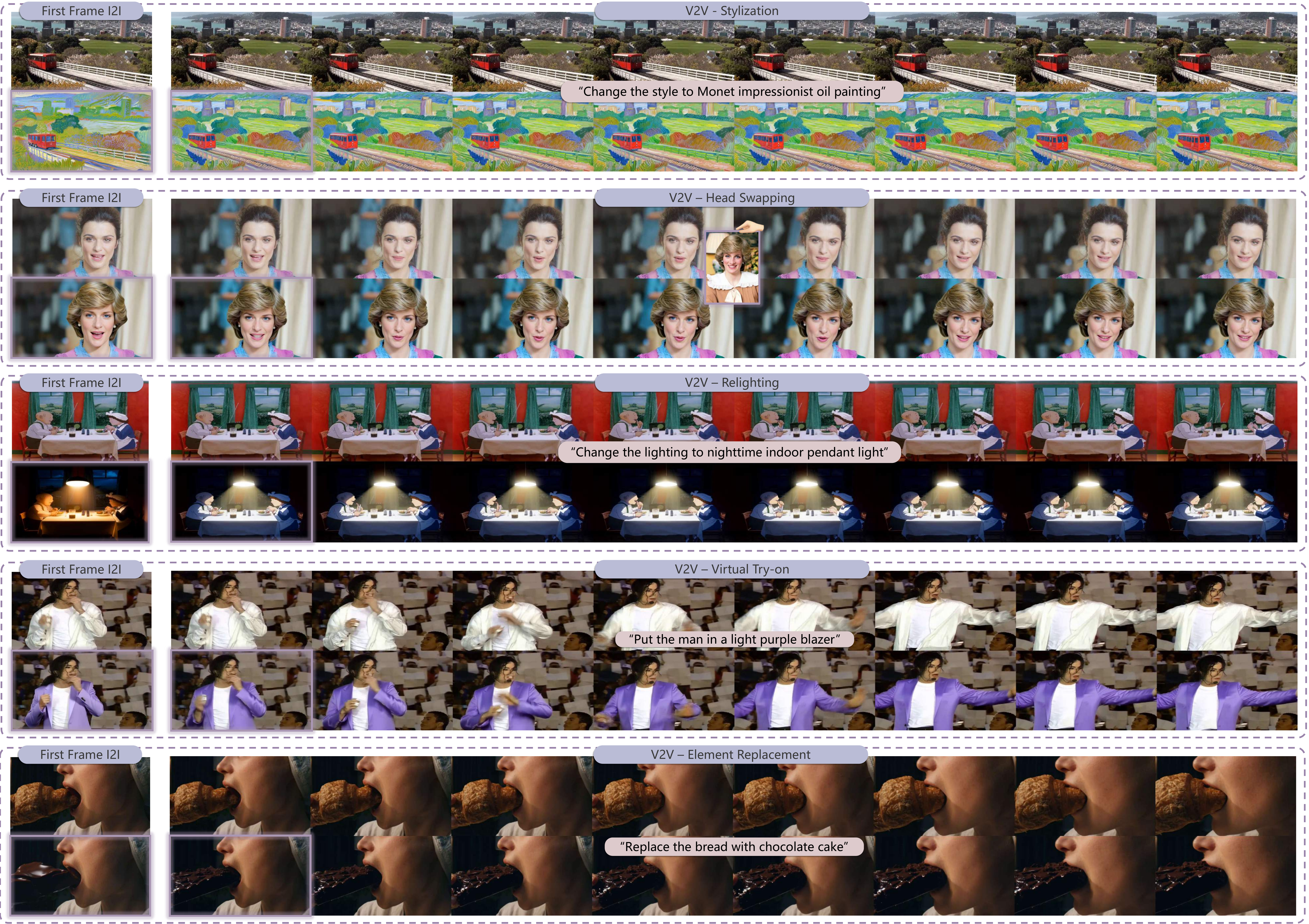} 
    \captionsetup{font=scriptsize}
    \caption{
    \textbf{Comparison of the visual output distributions between video editing and image editing tasks under the same reference first frame.}
    The modality mimic editing loss aims to align the editing capabilities between V2V and I2I. Given a reference video and an editing instruction, we perform video editing while also applying image editing to the first frame of the reference video. For the different tasks, the first and second columns within the purple rectangles show the results of image editing and the first frame of video editing, respectively. It can be observed that the model's editing output distributions for the video and image modalities are relatively aligned.
    }
    \label{fig:flowmimic_fig_25}
\end{figure*}

\begin{figure*}[htpb]
    \centering
    \includegraphics[width=1\linewidth, trim=0pt 2pt 0pt 2pt, clip]{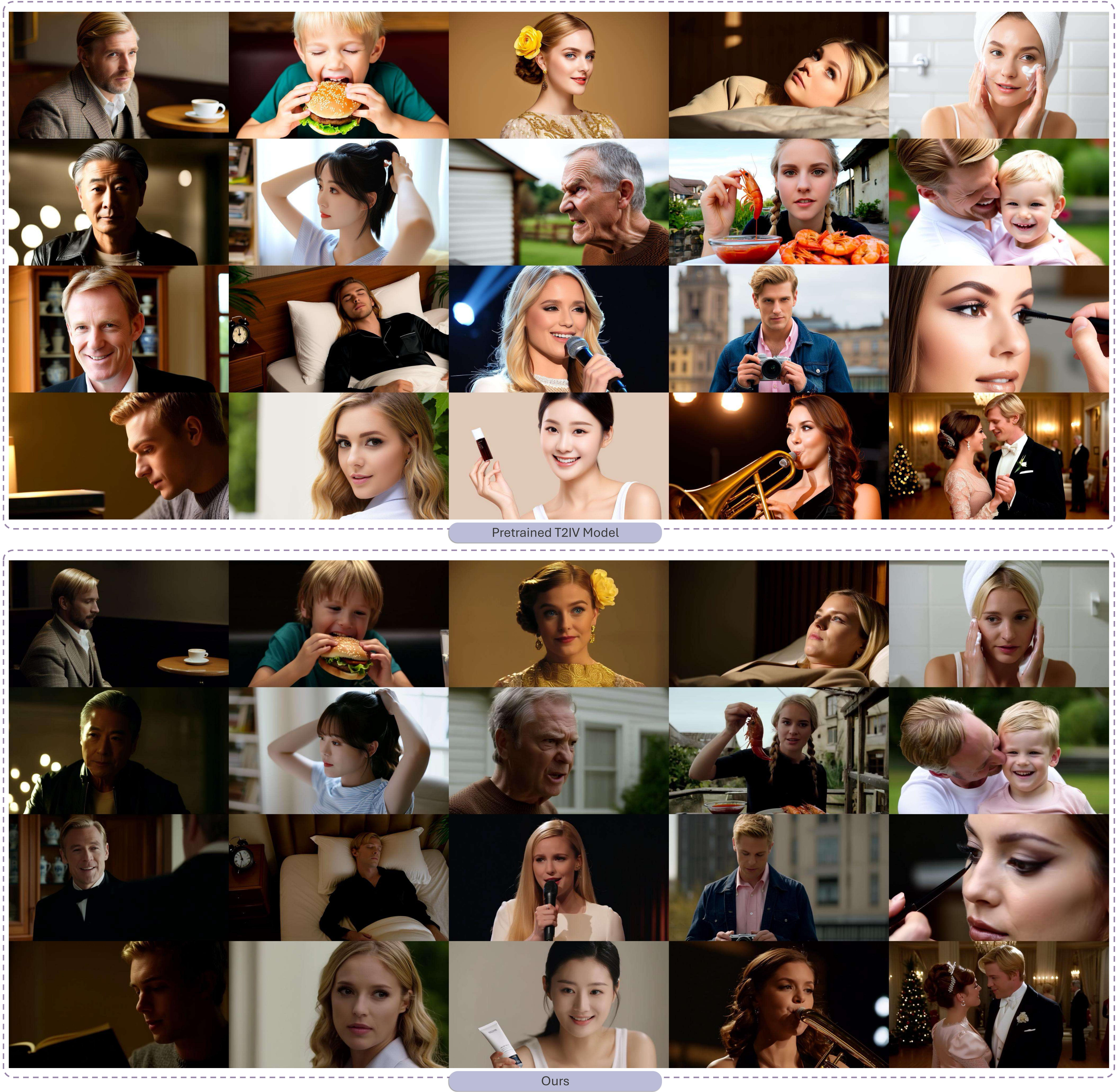} 
    \captionsetup{font=scriptsize}
    \caption{
    \textbf{Comparison of T2I results between the pretrained T2IV model and our model under identical cinematic prompts.}
    The modality mimic generation loss aims to align the generation distributions between T2V and T2I, bringing the latter closer to the former in terms of cinematic realism. The results indicate that, compared to the pretrained T2IV model, our model yields improved cinematic realism in T2I generation.
    }
    \label{fig:flowmimic_fig_26}
\end{figure*}

\begin{figure}[htpb]
    \centering
    \includegraphics[width=1\linewidth, trim=0pt 2pt 0pt 2pt, clip]{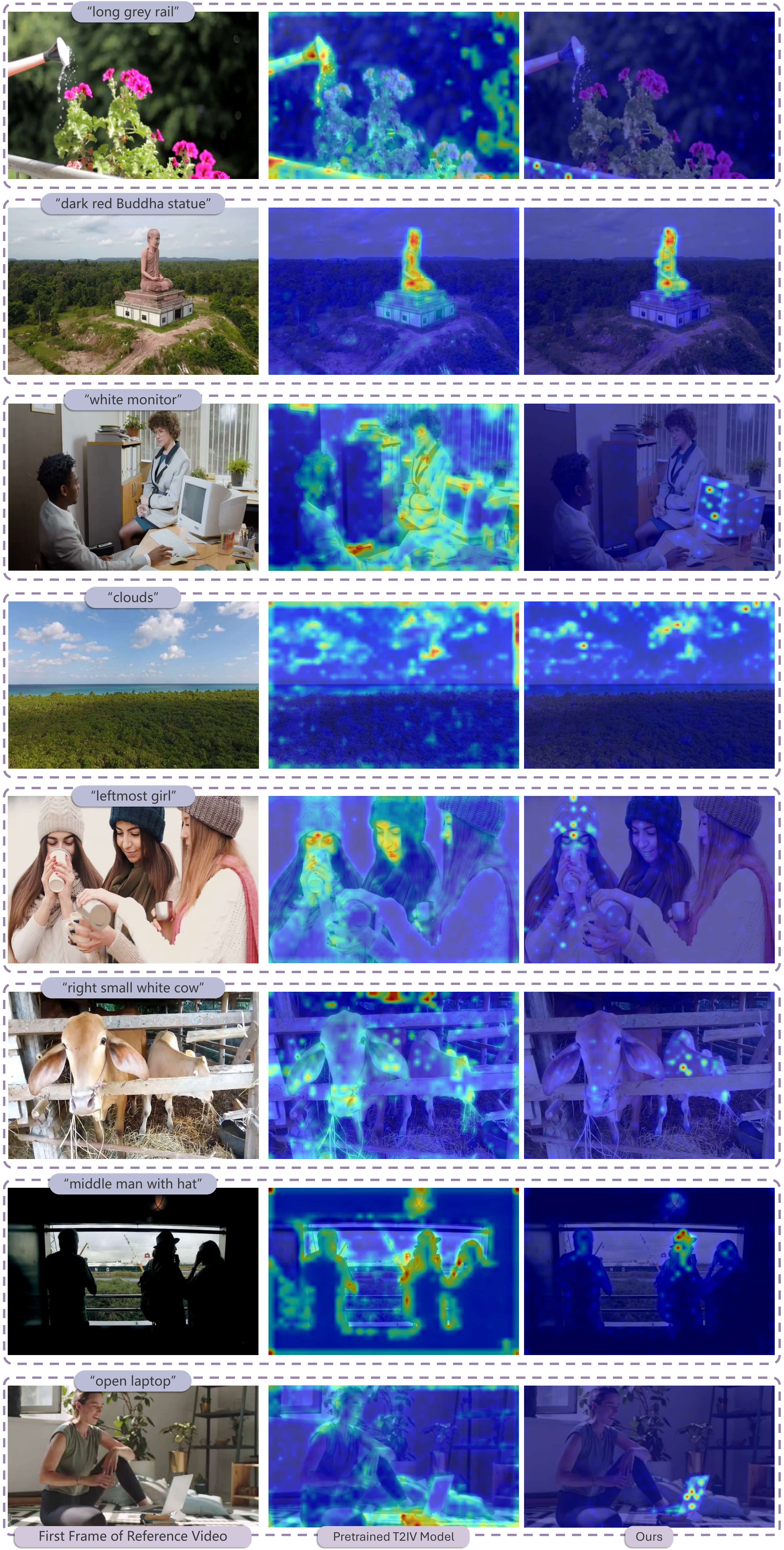} 
    \captionsetup{font=scriptsize}
    \caption{
    \textbf{Visualization comparisons of cross attention maps between referring expressions' corresponding text embeddings and the first frame of the reference video on UNIC-Bench.}
    To enhance the model's comprehension of natural language and visual content, as well as its cross-modality matching capability, we design sense-related tasks and losses. The results demonstrate that, compared with the pretrained T2IV model, our model can more accurately focus on the visual regions corresponding to the referring expressions in the reference video. Please refer to \cref{para:exp:qualitative_sense-relatedrelated-tasks-losses} for details.
    }
    \label{fig:flowmimic_fig_27}
\end{figure}

\begin{figure*}[htpb]
    \centering
    \includegraphics[width=1\linewidth, trim=0pt 2pt 0pt 2pt, clip]{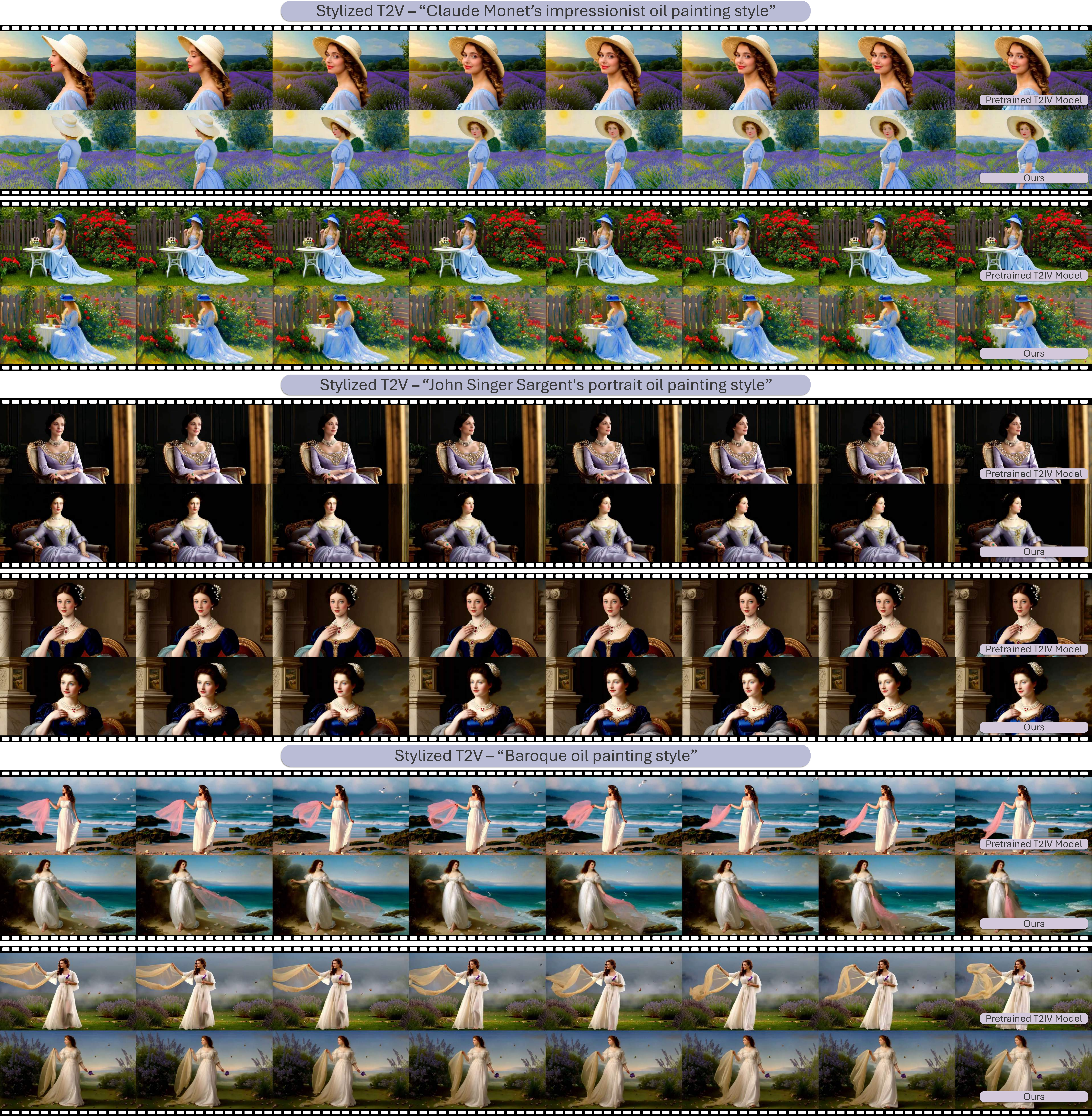} 
    \captionsetup{font=scriptsize}
    \caption{
    \textbf{Comparison of T2V results on stylized prompts.}
    For three common oil painting styles, we employ identical stylization prompts and random seeds for T2V generation. As can be seen from the results, compared to the pretrained T2IV model, our model exhibits a better stylistic response.
    }
    \label{fig:flowmimic_fig_28}
\end{figure*}

\paragraph{Image and Video Colorization Task.}
\label{para:Image and Video Colorization Task.}
The colorization task~\cite{li2026color, zhao2025colorsurge} for monochrome imagery refers to the conversion of a grayscale image or video sequence into a plausible color version. 
A pertinent and compelling application of this technology is the transformation of early black-and-white movie clips from the silent-film era and beyond, such as Charlie Chaplin's Limelight, into vividly colored films.

For the image colorization task, we construct a training sample as follows. 
Let $I_{\text{col}} \in \mathbb{R}^{H \times W \times 3}$ be a color image randomly sampled from the training set of other editing tasks; this serves as the target image. 
The corresponding source grayscale image $I_{\text{gray}} \in \mathbb{R}^{H \times W \times 3}$ is obtained by applying the standard luminance conversion formula to each pixel. 
Denoting the red, green, and blue channels of $I_{\text{col}}$ as $R, G, B \in \mathbb{R}^{H \times W}$, the luminance (grayscale) value $Y$ for each spatial location is computed as:
$
Y = 0.299 \cdot R + 0.587 \cdot G + 0.114 \cdot B.
$
The three-channel grayscale image is then formed by replicating $Y$ across the channel dimension: $I_{\text{gray}} = [Y, Y, Y]^{\top}$. 
The editing instruction is provided via a template sentence, e.g., “Add appropriate colors to image 1''.
For the video colorization task, we also generate video editing samples from the static image colorization data via the pixel-pair temporal warped flow field. 

\subsection{From stylized T2I to stylized T2V}
\label{subsec:From stylized T2I to stylized T2V}
We examine the stylized video generation capability of the pretrained T2IV model—for instance, in Impressionist and Baroque painting styles—and note that the generated videos are typically in a live-action style. 
Similarly, the majority of our T2V training videos are also drawn from live-action films and television series. 
Since our T2I data include a portion of training examples in these artistic styles, a natural extension is to enable T2V to acquire a stylized generation ability comparable to that of T2I. 
Following a common practice, we temporarily duplicate the T2I training image as multi-frame videos for T2V training; these static videos contain subsets generated from stylized images. 
We append the sentence ``The camera remains still.'' to the original T2I prompt to serve as the T2V prompt for such static videos.
Specifically, during the modality mimic generation step, a frame count, $F_{\text{static}}$, is randomly sampled from the set $\{13, 25, 37, 49, 61\}$. 
The first T2I sample of the current training batch is then repeated to $F_{\text{static}}$ frames to form a static video sequence, which replaces the second T2V sample in the training batch.

%% file: 04_exp.tex
\section{Experiments}
\label{sec:experiments}
In \cref{subsec:implementation_details}, we introduce the implementation details of FlowMimic.
The benchmarks employed in our experiments are described in \cref{subsec:benchmarks}. 
Then, in \cref{subsec:qualitative_video_editing_results}, we qualitatively demonstrate the FlowMimic's editing results trained solely on image editing data and video editing data generated online from them via pixel-wise temporal warped flow field—of video editing tasks across different levels.
Besides, in \cref{subsec:qualitative_discussion}, we qualitatively discuss the methods including the modality mimic losses and the sense-related tasks and corresponding losses.
Finally, we demonstrate the stylized T2V results in \cref{para:exp:stylized-t2v-results}.

\subsection{Implementation Details}
\label{subsec:implementation_details} 
We use the publicly available Wan2.1-T2V-1.3B model~\cite{wan2025wan} as the pretrained T2IV model. The experiments are conducted on 32 mid-range 80GB GPUs. 
We employ the programmatic function ``torch.nn.functional.kl\_div'' to implement $\mathcal{L}_{\text{mimic, gen}}$ in \cref{eq:39}, with the ``target'', ``input'', and ``reduction'' arguments set to $p^1_{\text{t2v-f,0}}$, $\log(p^1_{\text{t2i,0}})$, and ``batchmean'', respectively.
The learning rate is linearly increased from $5 \times 10^{-6}$ to $1 \times 10^{-5}$ over 5000 warm-up steps and then held constant.  
The ratio of generation training steps to editing training steps is 4:1.
The following loss coefficients are applied:  
$
n_\text{mimic,gen}=5,\; n_\text{mimic,editing}=3,\; 
\alpha_\text{mimic,gen}=0.1,\; \alpha_\text{mimic,editing}=0.1,\; 
\alpha_\text{FM,SC}=0.5,\; \alpha_\text{sense,attn}=1.0.
$
For the hyperparameters introduced in \cref{subsec:Miscellaneous editing-task specifics}, we set:
$
\eta_\text{min}=0.5,\; \eta_\text{max}=0.95,\; 
\xi_\text{min}=0.2,\; \xi_\text{max}=0.33,\; 
\zeta_\text{min}=0.2,\; \zeta_\text{max}=0.33.
$
In contrast to the current large-parameter models, our model is designed to verify the feasibility of the flow field paradigm and algorithms with a relatively light number of parameters and a limited training sample size. 
Specifically, to prevent memory overflow, the number of frames in the video editing samples generated online is uniformly sampled from 13, 25, 37, 49, and 61. 
Each editing step's total training batch comprises 28 image editing samples and 12 video editing samples.

Besides, given that in our multi-reference image editing samples, some of the reference elements share identical resolutions and exhibit a degree of content coupling, we also process the reference images excluding the source image during online multi-reference V2V sample generation. 
Specifically, for each such reference image pertaining to the object insertion task using internal data, we either flip it, or replace it with the last frame of a 25-frame video online generated by applying a pan-based, zoom-based or rotation-based motion type with a certain probability. 
For other cases, we replace it with the last frame of a 25-frame video online generated by applying a zoom-based motion type with a certain probability. 
Finally, we apply a mild centre-focused resolution cropping to the boundary of these reference images with a random ratio sampled between 0.05 and 0.1.

Furthermore, in addition to the referring expression segmentation task, we also introduce a referring expression object removal task.
Specifically, we generate image editing data of the referring expression object removal task using RefCOCO~\cite{yu2016modeling}, RefCOCOg~\cite{yu2016modeling}, RefCOCO+~\cite{yu2016modeling}, and gRefCOCO~\cite{liu2023gres}. 
We input the original image and the ground-truth instance mask into the mask-guided image object removal model ObjectClear~\cite{zhao2025objectclear}, and use the resulting edited image as the target image for referring expression object removal samples, with the original image serving as the source image. 
Similarly, we generate referring expression object removal video editing samples using the pixel-wise temporal warped flow field. 
The editing caption is a template sentence such as ``Remove $\llbracket R \rrbracket$.'', where $R$ is the referring expression.

For the multi-reference wearable object insertion task, we construct about 5K image editing samples from an existing dataset AnyInsertion~\cite{son}.
Each original data sample provides: a reference image of the wearable object, the object type text label (e.g., ``necklace''), a target image including a person wearing the reference object, and a corresponding mask for the target image that indicates the location of the reference object within it. 
We utilize the subset where the reference object type text label belongs to ``glasses'', ``scarf'', ``necklace'', ``shoes'', ``bracelet'', ``ring'', and ``watch''. 
In a similar manner, we apply ObjectClear to the target image and its mask to produce an inpainted image; this becomes the source image for the wearable insertion task. Subsequently, the original reference image is adopted as an additional reference, and the original target image is kept as the ground-truth output.
We also employ our flow field paradigm to produce corresponding video editing samples. The editing instruction follows a template sentence, such as ``Add the $\llbracket tl \rrbracket$ in image 2 to the person in image 1.'', where $tl$ denotes the text label.

In addition to the publicly available image editing datasets mentioned above, the remaining training data for T2V, T2I, and I2I tasks are internal. 
Some of our I2I training samples exhibit a noticeable global color discrepancy between the source and target images. 
Owing to this, the inference results may therefore display a similar global color discrepancy from the reference inputs in some cases.
The T2V training videos available to us have a resolution around $832\times 480$ and exhibit certain limitations: they may contain subtitles, watermarks, and black borders. Moreover, as these videos were obtained by sampling one frame per 12 frames from the original footage, inter-frame smoothness is sometimes suboptimal. Please refer to \cref{sec:conclusion} for more details.
Using these resources allows for some efficiency gains in data processing.
In practice, training data for T2V, T2I, and I2I tasks are readily available in large quantities from existing public datasets.

\subsection{Benchmarks}
\label{subsec:benchmarks}
We present qualitative results for FlowMimic on the public video editing benchmarks FiVE-Bench~\cite{2025five} and UNIC-Bench~\cite{ye2025unic}, along with results on additional tasks using publicly accessible online videos (e.g., from YouTube), the latter encompassing tasks not covered by the standard benchmarks.
FiVE-Bench~\cite{2025five} includes 420 source-target prompt pairs spanning six video editing tasks: rigid object replacement, non-rigid object replacement, color alteration, material modification, object addition, and object removal.
Moreover, UNIC-Bench~\cite{ye2025unic} contains single-reference object removal task and multi-reference tasks including object insertion, object replacement, video stylization and first frame propagation.
Specifically, for video deblurring task, we employ the test set of GOPRO Large~\cite{Nah} dataset.
For all reference videos except for those employed in the temporal generalization experiment, we retain their original frame rates and use the first 25 frames as the final reference videos.
The reference videos for the temporal generalization experiment are also kept at their original frame rates and comprise 121 frames.
We perform inference on the mid-range GPU.

\subsection{Qualitative Video Editing Results}
\label{subsec:qualitative_video_editing_results}
Theoretically, provided that the image editing task meets a certain correspondence in layout between the source and target images—which need not be strict, for instance, skeletons to human bodies or vice versa—such editing capability can also be extended to videos via our pixel-pair temporal warped flow field. 
This type of editing task allows substantial freedom of definition and can be designed as required; we hereby select some classic or interesting tasks for demonstration.
During training, the weight of samples corresponding to each editing task is distributed in a prescribed ratio, which can be adjusted in practice to prioritize specific tasks as required.

\paragraph{Classical Video Editing Tasks.}
\label{sec:exp:qualitative_classical}
\cref{fig:flowmimic_fig_7}, \cref{fig:flowmimic_fig_8}, \cref{fig:flowmimic_fig_9}, \cref{fig:flowmimic_fig_10}, \cref{fig:flowmimic_fig_11}, \cref{fig:flowmimic_fig_12}, \cref{fig:flowmimic_fig_13}, and \cref{fig:flowmimic_fig_14} demonstrate the editing results on classical tasks on two common benchmarks: FiVE-Bench and UNIC-Bench.
Empirically, multi-task training does not pose a significant learning challenge, but it does slow the convergence speed. We regard this as a trade-off for enhancing the model's comprehensive editing capability. For training, we use only online-constructed video editing data based on pixel-pair temporal warped flow field. Results show that this is sufficient for the model to master multi-task editing, while the sense-related tasks and losses free the model from reliance on editing-region mask sequences.

\paragraph{More Practical and Interesting Video Editing Tasks.}
\label{sec:exp:qualitative_more-practical}
\cref{fig:flowmimic_fig_15}, \cref{fig:flowmimic_fig_16}, \cref{fig:flowmimic_fig_17}, \cref{fig:flowmimic_fig_18}, \cref{fig:flowmimic_fig_19}, \cref{fig:flowmimic_fig_20}, and \cref{fig:flowmimic_fig_21} present the model's performance on a broader range of tasks. 
Our training data also includes tasks not illustrated here, such as image to normal map, depth map to image, skeleton map to image, and sketch to image editing—both rigid and non-rigid—each assigned a modest weight. Indeed, our video editing results demonstrate that the definition of an editing task is highly flexible: provided that an image pair exhibits a loose correspondence in layout, the model can learn the corresponding video editing task through our data paradigm and algorithms.

\paragraph{Further Exploration of Editing Capabilities.}
\label{sec:exp:qualitative_further-capabilities}
\cref{fig:flowmimic_fig_21}, \cref{fig:flowmimic_fig_22}, \cref{fig:flowmimic_fig_23} and \cref{fig:flowmimic_fig_24} discuss the editing abilities of the model from different aspects.
The ability to respond to editing instructions involving multiple objects and multiple tasks is a required capability for an editing model.
Although our training data only contain single-object and single-task editing samples, our model has fundamentally acquired a genuine visual editing capability and can apply it to multi-object and multi-task scenarios. 
Relevant examples are presented in \cref{fig:flowmimic_fig_21}.

Condition-guided editing constitutes a popular application scenario, where recent works inject trajectory, explicit 2D, or explicit 3D-based signals via a control network into video models to accomplish tasks such as trajectory-customized video editing, body animation~\cite{zhu2024champ}, or portrait animation~\cite{zhang2025rigyourportrait}.
Our image editing data include a modest number of condition map to image pairs and their inverses—such as normal, skeleton, and depth maps—which endow the model with a degree of conditioned video generation capability, the ability to estimate 3D maps (e.g., normal and depth) from images, and the ability to estimate 2D maps (e.g., referring expression segmentation and semantic segmentation maps) from videos. Beyond these commonly used conditioning signals, we explore a video editing task associated with trajectory customization. Specifically, our image editing pairs include approximately 3K samples reversed from the RefCOCO dataset, in which a specified object is generated within a given color block region. In the video editing scenario, this ability corresponds to generating the specified object within a prescribed color block trajectory. As shown in \cref{fig:flowmimic_fig_22}, the model indeed exhibits this capability and can produce corresponding physical effects, such as the splashing of water. 

The normal map to image data comprises fewer than 2K samples, all of which are global normal maps of scenes—that is, they contain no data pertaining to bodies or portraits. The second row of \cref{fig:flowmimic_fig_23} demonstrates the model's generalization capability, for which the conditioning sequence of body animation typically consists of foreground body normal maps. Furthermore, \cref{fig:flowmimic_fig_23} illustrates that, for the unseen task of animating a reference image with the condition maps, the model also exhibits a certain degree of generalization ability.

Whether the number of inference frames can exceed the average or maximum number of training frames is a pertinent concern. The results in \cref{fig:flowmimic_fig_24} demonstrate that the editing capability of our model is not confined to the training frame distribution; rather, it exhibits a degree of temporal generalization.

\subsection{Qualitative discussion on the methods}
\label{subsec:qualitative_discussion}
\paragraph{Modality mimic losses.} 
Modality mimic losses are designed to align the generation distributions of the model across the video and image modalities. Specifically, we introduce an editing loss and a generation loss respectively. As shown in \cref{fig:flowmimic_fig_25}, the editing results exhibit a close distributional similarity regardless of whether a full reference video or only its first frame is provided. Modality mimic generation loss aims to bring the cinematic realism of the T2I generation distribution closer to that of the model's T2V capability. In \cref{fig:flowmimic_fig_26}, it can be observed that, compared to the pretrained T2IV model, our model yields better cinematic realism for cinematic prompts. The resolution for T2I here is $832\times 480$.
\paragraph{Sense-related tasks and losses.} 
\label{para:exp:qualitative_sense-relatedrelated-tasks-losses}
Sense-related tasks and losses, in particular the sense text cross attention map loss, are designed to strengthen the model's comprehension of the natural language and visual modalities, as well as its ability to perform cross-modal matching between them. To examine the model's capability in this respect, we visualize the cross attention maps corresponding to specified phrases. 
Specifically, as shown in \cref{fig:flowmimic_fig_27}, we reconstruct the first 25 frames of reference videos from the UNIC-Bench, i.e., the output caption matches the reference caption. 
For the first seven examples, the referring expressions are the objects to be removed in the object removal task; for the final stylization example, we use ``open laptop'' as the referring expression. 
During inference, for the final sampling step, we first compute the cross attention maps between the visual tokens of the reference video's first frame and the text embeddings corresponding to the referring expression, then average them across DiT attention head numbers and DiT block numbers, and finally upsample and visualize the result as a heatmap.
From the results, it can be observed that, compared to the pretrained T2IV model, our model is capable of attending to a more accurate visual region for a given specific expression.

\subsection{Stylized T2V results}
\label{para:exp:stylized-t2v-results}
The strategy we employ—generating static video sequences from T2I training images to serve as T2V training samples—is designed to enable the video modality to acquire a stylized generation capability close to that of the image modality. 
As shown in \cref{fig:flowmimic_fig_28}, the results demonstrate that, compared to the pretrained T2IV model, our model achieves better stylistic responsiveness.

%% file: 10_conclusion.tex
\section{Conclusion}
\label{sec:conclusion}
In this paper, we aim to model the video and image modalities jointly for both generation and language-guided editing with optional reference images. 
Specifically, we propose a pixel-wise temporal warped flow field, an online, real-time data generation paradigm that enables scalable acquisition of video editing data—comprehensible and learnable by the model—from image editing data alone. 
This paradigm can be applied to freely defined editing tasks, provided a loose layout correspondence exists between the source and target images. 

To align the generative capabilities of the video and image modalities and to allow each modality to learn from the strengths of the other, we design a modality mimic generation loss and a modality mimic editing loss. 
These losses encourage the T2I outputs to approach the cinematic realism of the T2V outputs, and steer the video editing output distribution towards the image editing distribution, which converges faster and benefits from a larger number of training samples per step.

The foundation of editing lies in the model's correct comprehension of the editing instruction and the reference visual content, its ability to accurately locate the visual region corresponding to the subject specified in the text within the reference video or image, and its capacity to modify only that specific region. 
Existing methods often simplify this process by either connecting a Multimodal Large Language Model (MLLM) via a connector followed by multi-stage fine-tuning to identify the target region, or by directly inputting a mask sequence to explicitly indicate the editing area. 
We aim for the model itself to internalize this capability, while simultaneously reducing parameter count and the number of training stages. 
Therefore, we design sense-related tasks and corresponding region-aware latent-level and attention-level losses. 
The visualization of cross-attention maps demonstrates that our model can, based on a specific phrase in the text, attend relatively accurately to the corresponding region in the reference visual input. Moreover, the editing results across multiple tasks indicate that the model is able to correctly locate and modify only the intended editing region. This collectively signifies that the model has, to a considerable extent, internalized this capability.

\paragraph{Future works.}
The pixel-pair temporal warped flow field enables the pixel-level preservation and real-time synthesis of corresponding video pairs from both condition-map-to-image and image-to-estimated-map data. 
Even with limited quality and quantity of image training samples, the results demonstrate that this capability can be generalized to high-level video understanding tasks (e.g., semantic segmentation, referring expression segmentation), low-level video processing tasks (e.g., video deblurring), and controllable video generation, representing a promising research direction. 
Therefore, enhancing the quality and quantity of these image training samples provides a plausible path for further improvement. 
For example, data for sense-related tasks is currently composed solely of the RefCOCO series datasets.
These data contain a relatively limited variety of referring expressions, and the instance masks are often delineated with relatively coarse contours; this could be improved by supplementing them with data from a broader range of categories, as well as by providing instance masks with more finely detailed contours.

Furthermore, the majority of the T2V training data available to us comprises dimly lit film and television footage of moderate quality and low frame rate—a result of further frame subsampling from the original material, which leads to pronounced inter-frame motion discrepancies and a degradation of temporal smoothness. 
These clips are also frequently marred by subtitles, watermarks, or black bars at the top and bottom.  We may therefore attempt to acquire higher-quality, clean data to improve training performance.
Besides, for many of our tasks, the number of corresponding image editing samples is quite limited. 
The available data also includes contaminated or low-quality samples. 
For instance, training data for face swapping and head swapping tasks is not only scarce but also of low precision. 
For example, expression, lip movement, gaze direction and head pose before and after face swapping or head swapping lack consistency.
Although ObjectClear is an impressive mask-guided image object removal model at present, its inpainting results on the RefCOCO series of datasets are often blurry, failing to fill the masked regions with plausible and clear content. 
This issue is particularly pronounced on the gRefCOCO dataset, where instance masks encompass multiple objects. 
Consequently, when we use such editing results as target images for referring expression object removal samples, the model's effectiveness is affected to some degree, manifested as reduced output clarity and diminished reasonable inpainting capability. 
We believe that with the rapid development in the field of image inpainting, more suitable image object removal models will soon emerge to address the limitations introduced by such third-party tools.
Furthermore, a noticeable global color discrepancy exists between the source and target images in some of our I2I training samples. This can lead to a corresponding color shift between the inference results and the reference inputs. We believe this phenomenon can be mitigated by employing higher-quality training data and implementing a filter to exclude samples with significant global color discrepancy.

Additionally, we have not performed dedicated hyperparameter tuning for the weighting of each individual task. 
In practice, for tasks of greater interest, increasing their corresponding data weights could expedite convergence in those areas, and the model's performance would consequently become more proficient on these tasks. 
Some of the editing tasks presented in this paper are assigned minimal weights; nevertheless, experimental results from the exploratory phase under task-focused training indicate that the performance ceiling for each editing task, when trained in a focused manner, is relatively high.

\paragraph{Ethics considerations.}
AI visual models must not be maliciously employed in any context, such as to attack individuals or to generate or edit defamatory video content. We firmly oppose such practices, and we hold that relevant legal frameworks and ethical guidelines should be progressively refined alongside the advancement of visual models.

\paragraph{Acknowledgments and Timeline.}
This project was initially launched to verify whether hybrid training across image and video modalities could enable the two modalities to benefit each other in both generation and editing tasks. The exploration gave rise to the data paradigms and algorithms included in this paper—for instance, the data paradigm transfers the image editing capabilities to the video editing counterparts.
The methods and the motivations described in this paper, as well as the exploratory attempts, were documented in detail concurrently in the timestamped weekly reports, together with the associated timestamped weblinks to results visualizations (e.g., results of different video editing tasks, image editing results, results of the reference-to-video task using the method introduced in \cref{subsec:Reference-to-Video task}, results of the broader notion of I2V with editing instructions, and T2I and I2I results under various explored modality mimic losses and strategies). The implementation and experiment logs of each method, as well as the writing of this study, were recorded using Git.

Wei Liu led this project. 
Dingyun Zhang proposed and implemented the core algorithms for FlowMimic. 
This includes the methods detailed in \cref{sec:method} (excluding the element identity encoding with absolute positional encoding), the color block abstraction method provided in \cref{sec:FFP Details and Color Block Abstraction}, the modality mimicry paradigms in \cref{sec:The Modality Mimicry Paradigms}, and other attempts during the exploration phase in \cref{sec:Other Attempts During the Exploration}.
Among these, the modality mimic generation loss was proposed and implemented in April 2025, with the associated exploratory experiments completed during the same period; the strategy in \cref{subsec:From stylized T2I to stylized T2V} was explored in June 2025; the adjustments made to the model architecture for the editing tasks, as described in \cref{subsec:Model Architecture Adaptation to Editing Task}—including the element identity encoding with 1D RoPE for element dimension to each sample element appended after the 3D RoPE—were proposed and implemented in August 2025; the pixel-pair temporal warped flow field and the modality mimic editing loss were proposed and implemented in mid-September 2025; the editing-task details in \cref{subsec:Miscellaneous editing-task specifics} were explored in September and October 2025; and the sense-related tasks, together with their corresponding losses, were proposed and implemented in November and December 2025.
Between September and early December 2025, she validated that, when training on both generation and editing tasks across video and image modalities simultaneously, the pixel-pair temporal warped flow field enabled FlowMimic to master the video editing capabilities of all the editing tasks presented in this paper—and beyond—without requiring specially constructed or collected video editing training samples. This was observed under both focused training on a single editing task and mixed training on all editing tasks.
She wrote this paper and completed it in mid-May 2026 (with the unrevised manuscript finished in April 2026; the paper became internally accessible on the originally scheduled date of 22 May 2026, and was uploaded to the publication review platform pending approval; the PDF of this paper received a credible timestamp certification\footnote{\url{https://ipr.tsa.cn/}} dated 20 May 2026), trained the models presented in the paper, integrated the Wan2.1 training framework, and handled the processing of offline T2V sample VAE latents. 
Lixue Gong proposed and implemented the element identity encoding to the reference elements with absolute positional encoding as the replacement of the 1D RoPE. She performed frame resampling and caption labeling for the original T2V data, and integrated the T2I and the in-context I2I dataloaders within the Wan2.1 framework. 
Dingyun Zhang applied the element identity encoding to the target element with index zero, while reference elements are assigned indices beginning from 1 in September 2025.
We also thank Zixuan Zhang for acquiring the T2V cinematic data and constructing the internally compiled evaluation set. 
We would like to thank the people involved in constructing the internal T2I and I2I data used in this study.

%% file: 12_appendix.tex
\section{FFP Details and Color Block Abstraction}
\label{sec:FFP Details and Color Block Abstraction}

\begin{figure}[htpb]
    \centering
    \includegraphics[width=1.0\linewidth, trim=0pt 0pt 0pt 0pt, clip]{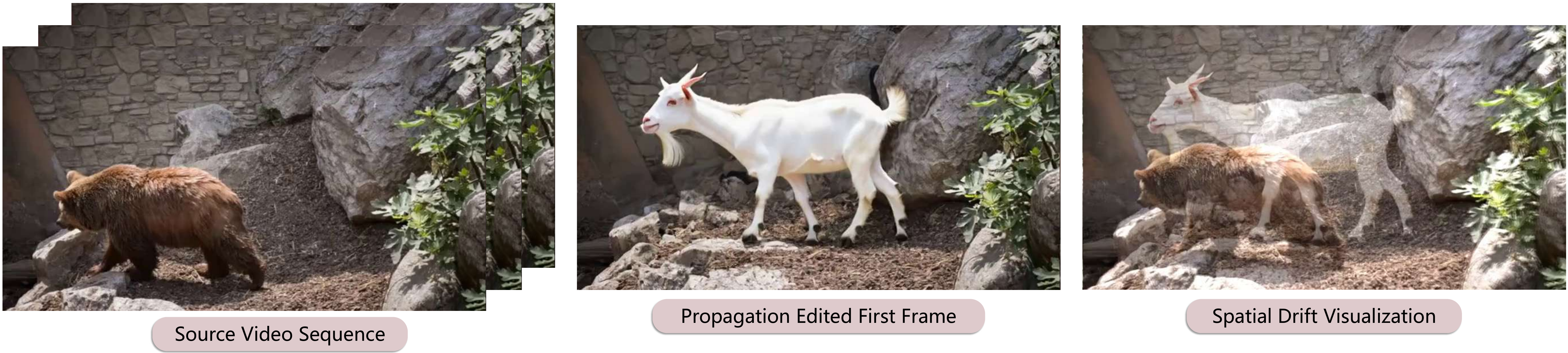} 
    \captionsetup{font=scriptsize}
    \caption{
        In the first frame propagation test samples of UNIC-Bench~\cite{ye2025unic}, instances occur where the object replacement editing result of a specified object in the first source video frame by a third-party tool exhibits significant spatial drift from the original object. For example, in the input video depicting a walking brown bear, the editing applied to the first frame yields a walking goat. The goat is expected to appear at the location of the brown bear; however, the visualization result demonstrates a noticeable spatial drift between the goat and the bear. This discrepancy becomes even more pronounced at the pixel level, where details such as the precise positioning of the four legs during the stepping motion fail to align.
    }
    \label{fig:flowmimic_fig_29}
\end{figure}

In this paper, we balance the diversity of training tasks with convergence speed to validate the effectiveness of our data paradigm and algorithms. 
Accordingly, for first frame propagation, we primarily train on subtasks such as object addition, object removal, and outpainting, while also including a very limited proportion of subtasks like stylization, object replacement, and background replacement. 
The training data for all these tasks is generated online from single-reference image editing samples. 
In practice, in our focused exploration of the first frame propagation task, we additionally generate training data for these tasks online from corresponding image pairs in multi-reference image editing samples.
The specific methodology for online generation is detailed in \cref{para:First Frame Propagation.}.

In the first frame propagation object replacement test samples of UNIC-Bench~\cite{ye2025unic}, we observed that the replacement of specified objects in the first frame of the source video by third-party image editing tools occasionally resulted in a significant spatial drift between the replaced object and the original, as illustrated in \cref{fig:flowmimic_fig_29}. 
Consequently, in the focused experiments, we devised a targeted approach in October 2025 to address the significant spatial drift in the replacement results caused by such third-party tool errors.
Naturally, we aim to expose the model during training to similar cases where the replacement result exhibits substantial spatial drift from the original object. 
Rather than collecting additional image editing data that meets the requirements, we employ an abstraction-based method and verify its effectiveness.
Specifically, we employ ``color block abstraction'' to simulate this scenario, positing that from the model's perspective, the two are analogous and thus learnable.
In particular, in line with the motivation behind the pixel-pair temporal warped flow field, we assume that, from the model's perspective, two spatially drifted objects can be represented by two color blocks of distinct colors and shapes that are also spatially drifted. 
This approach assists the model in learning the object replacement editing between two objects when their pixels are only coarsely corresponding.

\begin{figure}[htpb]
    \centering
    \includegraphics[width=1.0\linewidth, trim=0pt 0pt 0pt 0pt, clip]{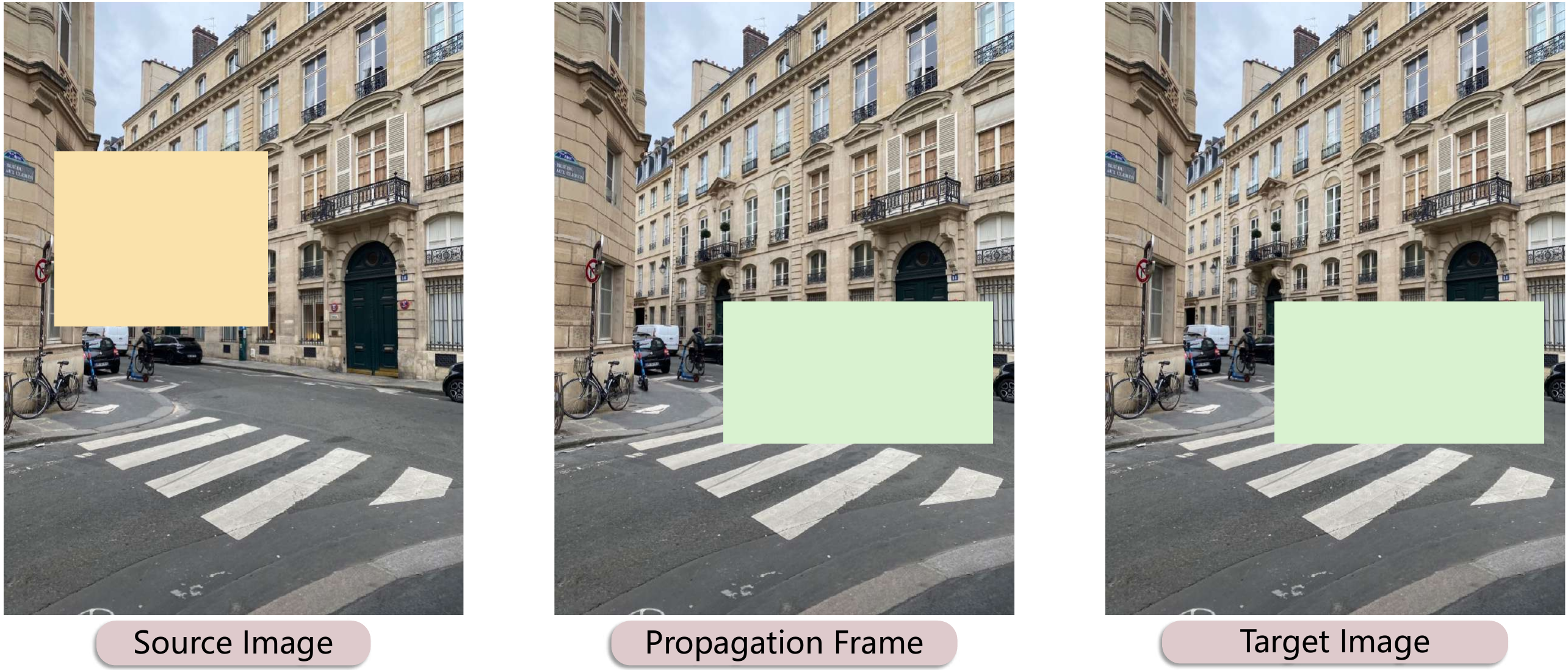} 
    \captionsetup{font=scriptsize}
    \caption{
        \textbf{Illustration of the color block abstraction method.}
        We employ color blocks to abstractly represent objects, where the original color block and the replacement color block exhibit significant differences in color, shape, and position. This approach simulates scenarios where the editing result in the propagation frame demonstrates notable localization drift relative to the original object in the first source frame.
    }
    \label{fig:flowmimic_fig_30}
\end{figure}

As shown in \cref{fig:flowmimic_fig_30}, for an image containing a subject, such as a person, or a background-only image without a subject, we replicate it into three copies. 
In the first image, we add a rectangle of random color at a random position, representing the original subject. 
In the second image, we add a rectangle of a different random color, with its size and position varying within a certain threshold relative to the first rectangle.
The positional relationship is further categorized into three cases: complete containment within the first rectangle, partial overlap with the first rectangle, or no overlap at all. 
The rectangle in the second image represents the editing result of object replacement, exhibiting significant positional drift from the original object. 
The third image is identical to the second image containing the rectangle, serving as the target image. 

\begin{figure}[htpb]
    \centering
    \includegraphics[width=1.0\linewidth, trim=0pt 0pt 0pt 0pt, clip]{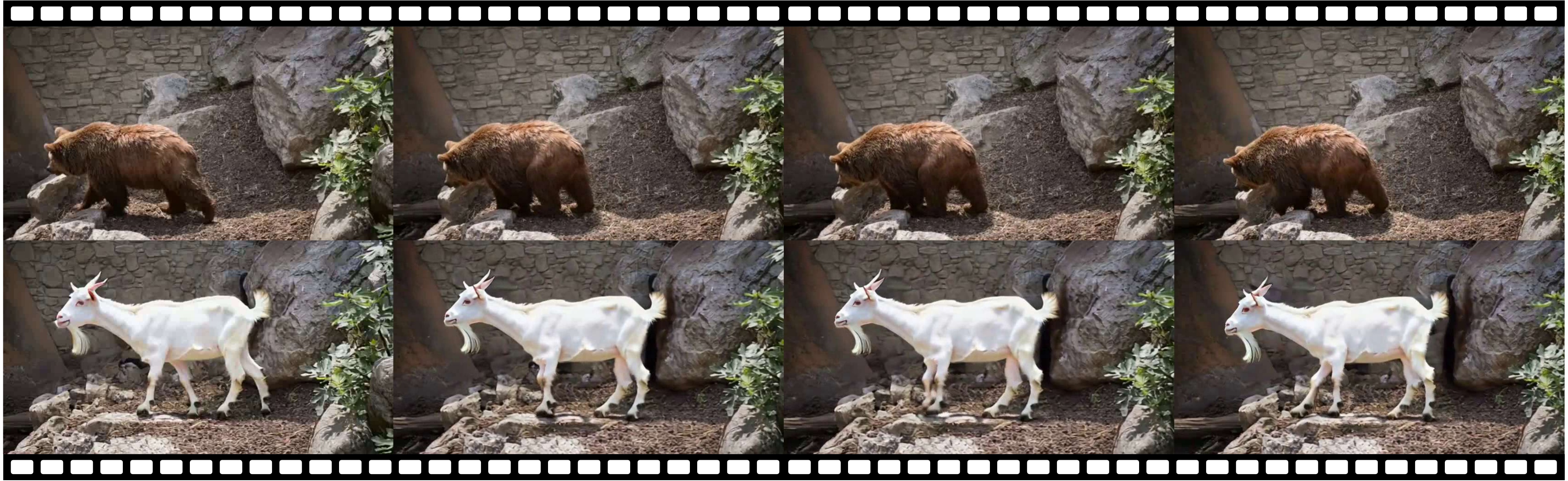} 
    \captionsetup{font=scriptsize}
    \caption{
        \textbf{First frame propagation results with the color block abstraction methodology.}
        It can be observed that the model is capable of generating plausible motion for the replaced object, despite distinct spatial discrepancies and action misalignment between it in the propagation frame and the original object in the first source frame. 
        This holds true even when background inconsistencies are present due to editing artifacts introduced by the third-party tool.
    }
    \label{fig:flowmimic_fig_31}
\end{figure}

Based on this image sample, we generate a video editing sample online, where the first and third images form the image pair for obtaining the video pair, and the second image acts as the reference propagation first frame.
Through this approach, the replaced object in the first frame propagation results exhibits natural movement consistent with that of the original object, as demonstrated in \cref{fig:flowmimic_fig_31}.

The color block abstraction method also encompasses several variants. 
For instance, one may initially generate a video pair online from an image pair, then incorporate two rectangles meeting the aforementioned conditions into the first frames of the source and target videos, respectively, and allow them to undergo smooth, unidirectional motion over time. 
In the context of object removal tasks, we have also explored the addition of temporally static or moving color blocks to reference videos containing subjects, with the target video remaining the original reference video, thereby simulating video inpainting data under occlusion scenarios.
The shape of the color block is not limited to rectangles; it may, for example, deform over time or be replaced by a color instance mask. 
Color block abstraction is a method we have validated as effective during our exploration; however, the results presented in this paper do not include this strategy, as we intend for the model to learn to preserve pixel-wise editing effects during object replacement, and therefore do not consider the significant editing errors introduced by such third-party tools.

\section{The Modality Mimicry Paradigms}
\label{sec:The Modality Mimicry Paradigms}
We consider that the motivation and methodology of the modality mimicry section in \cref{subsec:First Frame Modality Mimic Loss} can be generalized into different paradigms from different perspectives and may potentially be applied to corresponding suitable scenarios. 
The concrete implementations presented in this paper, together with the approaches explored during the exploration phases later introduced in \cref{sec:Other Attempts During the Exploration}, may potentially be adapted to these specific scenarios. 
Two such examples are listed below.

\subsection{On-policy Self-distillation of Flow Matching}
\label{subsec: On-policy Self-distillation of Flow Matching}
The modality mimicry paradigm can be generalized as follows: given two samples that share identical configurations except that one enjoys a more favourable condition, the model performs better on the sample with the favourable condition. 
By applying a suitable loss or strategy, we encourage the model's performance on the naive sample to imitate its performance on the favourable-condition counterpart. 
Through such capability imitation, the model exhibits comparable performance at inference time regardless of the presence of the favourable condition.
That is, the capability originally associated with the favourable condition is elevated to a permanent capability that persists even in its absence.

In this paper, taking the modality mimic generation method as an example, other configurations correspond to the original latents, noises, timesteps, noisy latents, and caption text embeddings, while the favourable condition is undergoing temporal attention alongside subsequent frames. 
Multi-frame T2V, by virtue of its temporal dimension, is inherently more suited to the entangled spatio-temporal full self-attention architecture. 
Consequently, the T2I generation ability of the model is encouraged to mimic its photorealistic T2V counterpart. 
Similarly, for models trained with specially constructed subject-to-video (S2V)~\cite{chen2025phantom} and subject-to-image (S2I)~\cite{xiao2025omnigen} data—such as an appealing model, Phantom~\cite{liu2025phantom}—a gap in cinematic realism and identity preservation was also observed between S2I and S2V.
Therefore, one can also employ the modality mimic editing loss to encourage the model's S2I ability to mimic its more photorealistic and lifelike S2V counterpart. 
Besides, our attempts based on the pretrained T2IV model on the reference-to-video task are briefly described in \cref{subsec:Reference-to-Video task}.

\paragraph{FlowMimic-OPSD.}
As this paper was nearing completion, the term \textit{on-policy self-distillation} (OPSD)~\cite{zhao2026self,song2026survey,1996reinforcement} has gained considerable traction. 
We consider that, viewed from the perspective of OPSD, the mimicry procedure underlying the modality mimicry paradigm, which we explored last April, essentially constitutes a form of OPSD for flow matching models. 
The student generates its own ``mistakes'' under arbitrary states and is corrected in real time by the teacher, which also originates from the same model.
The modality mimicry paradigm was conceived during the early exploration phase, based on the observations exemplified in \cref{fig:flowmimic_fig_2}; consequently, the modality mimicry section of this paper does not employ the term OPSD. 
When viewed from the OPSD perspective, however, the editing and generation results of modality mimicry in \cref{fig:flowmimic_fig_25} and \cref{fig:flowmimic_fig_26} demonstrate that \textbf{OPSD of flow matching models is indeed feasible} and that OPSD for both generation and editing can be performed simultaneously.
The mimicry methods described in \cref{subsec:First Frame Modality Mimic Loss} and \cref{sec:Other Attempts During the Exploration}—for instance, the modality mimic generation feature score loss or feature loss in \cref{subsec:Mimicry on Visual Perceptual Feature Scores}—can also be applied to the OPSD of image or video flow matching models, such as Z-image~\cite{cai2025z} and Stable Diffusion 3.5 Medium~\cite{esser2024scaling}.

Furthermore, we consider that this modality mimicry paradigm may also be applied to other favourable conditions (e.g., more detailed captions, additional reference images, reference images with better image quality, or views with a larger visible portion), thereby enhancing the model's performance or capability associated with those conditions.
For instance, consider a pretrained model endowed with both T2I and I2I capabilities. 
The favourable condition might involve providing a more detailed output caption relative to the vanilla one, with the corresponding capability being finer-grained generation or editing quality. 
Ideally, through the mimicry methods described in \cref{subsec:First Frame Modality Mimic Loss} and \cref{sec:Other Attempts During the Exploration}, the model can achieve finer-grained generation or editing under simple captions than would be attainable via vanilla supervised fine-tuning (SFT).
In particular, some self-distillation methods~\cite{chen2026enhancing,jin2026unisd} employ an exponential moving average (EMA) of the student model weights at the current training step as the updated teacher model weights to stabilize the training procedure, whereas our modality mimicry methods converge stably without resorting to this strategy.

\subsection{Modality Self-alignment of Multi-task Models}
\label{subsec: Modality Self-alignment of Multi-task Models}
In \cref{subsec: On-policy Self-distillation of Flow Matching}, we generalized the motivation and approach of modality mimicry into a paradigm which can be viewed from a distillation perspective.
From the perspective of distribution alignment, meanwhile, the motivation and methodology of modality mimicry can also be generalized into another paradigm.
Specifically, for tasks that are interrelated, it is a common view that training them simultaneously holds promise for mutually enhancing each other's performance.
In particular, consider two tasks that are isomorphic or inverse, and whose input modalities can be represented in a shared input feature space while their output modalities can be represented in a shared output feature space. 
We consider that it is natural to desire the distributions of the two output modalities generated from the unified model to be aligned in the shared output feature representation space when the inputs are also close in the shared input feature representation space at inference time—that is, modality self-alignment.
However, taking the simultaneous training of T2I and T2V tasks as an example, the results exemplified in \cref{fig:flowmimic_fig_2} indicate that simply training multiple tasks together is not sufficient in some scenarios. 
Therefore, we employ the mimicry methods described in \cref{subsec:First Frame Modality Mimic Loss} and \cref{sec:Other Attempts During the Exploration} to explicitly constrain this self-alignment during training, where the loss function can be a distribution alignment loss, a feature score loss or a feature distance loss, thereby bringing the cinematic realism of the T2I generation distribution closer to that of the model's T2V capability. 
This self-supervised paradigm, which incorporates the perspective and methodology of modality self-alignment, may also be applied to other multi-task training scenarios if the modality self-alignment is considered likely to explicitly or implicitly improve model performance across tasks.

For example, consider a pretrained multimodal model capable of both generating captions for images with the image captioning ability and synthesizing images from text prompts with the image generation ability. 
At inference time, if one inputs an image and a text prompt whose features lie close to each other in the shared CLIP bimodal feature space for Image-to-Text (I2T)~\cite{tang2026cccaption} and T2I tasks respectively—say, a matched text-image pair—then, ideally, the CLIP features of the generated text caption and image should also remain close in that feature space.
Therefore, the corresponding post-training can be conducted.
Specifically, during the post-training process, the I2T and T2I tasks are still trained simultaneously, and the same match text-image pair can serve simultaneously as both an I2T and a T2I training sample in the same training step.
The I2T branch can be trained under the cross-entropy loss with autoregressive next-token prediction, while the T2I branch can be trained under the flow matching loss. 
Similar to the approach described later in \cref{subsec:Mimicry on Visual Perceptual Feature Scores}, for such an I2T and T2I sample pair, one could compute a CLIP feature loss between the predicted text caption and the decoded image sampled at non-high-noise timesteps during training.
This may implicitly improve the image-text alignment and the prompt faithfulness for the I2T and T2I tasks, respectively.

In this case, the T2I and T2V tasks in this paper are replaced by the I2T and T2I tasks, and the image and video modalities are replaced by the image and textual modalities. 
Correspondingly, the distance measure between the output modalities changes from being computed between estimated original latents or the corresponding extracted perceptual feature scores to being computed in the shared bimodal feature space of a CLIP-like model.
Similarly, the two training tasks could be T2V under different settings, conditioned on low-quality RGB renderings of a 3D scene representation~\cite{kerbl20233d} along with camera parameters, with the visual modality involved being 2D RGB derived from the 3D representation.

\section{Other Attempts During the Exploration}
\label{sec:Other Attempts During the Exploration}
We briefly describe here a small subset of the other methods and tasks explored during the exploratory phase, in particular those related to the modality mimicry methods.
If readers are interested, we may supplement with more methods and tasks explored during the exploratory phase.

\subsection{Measures for Modality Mimicry}
\label{subsec:Measures for Modality Mimicry}
In April 2025, we also experimented with using Mean Squared Error (MSE) as the measure of modality mimic generation loss, applied between the flattened estimated original latents \(\mathcal{F}(\hat{z}^1_{\text{t2v-f},0})\) and \(\mathcal{F}(\hat{z}^1_{\text{t2i},0})\) during the modality mimic generation step:
\begin{equation}
\mathcal{L}_{\text{mimic, gen, mse}} = \bigl\| \mathcal{F}(\hat{z}^1_{\text{t2v-f},0}) - \mathcal{F}(\hat{z}^1_{\text{t2i},0}) \bigr\|^2.
\end{equation}
Intuitively, the MSE measure afforded less flexibility in the learning procedure of modality mimicry, positioning it closer to a latent space reconstruction loss rather than the distribution alignment loss—one that encouraged the model to further learn the distributions of abstract visual perceptual characteristics. 
This experiment was therefore conducted to empirically verify that the latter was indeed more suitable.
Based on our comparative observations at the time, $\mathcal{L}_{\text{mimic, gen, mse}}$ exhibits noticeably slower convergence speed and inferior performance in modality mimicry—as observed by the visual perceptual differences between the T2I outputs and the first frame inference results of long-clip T2V sequences—compared with the $\mathcal{L}_{\text{mimic, gen}}$ adopted in this paper.
Similar findings were also corroborated for the modality mimic editing loss in September 2025, where the MSE measure occasionally led to abnormal global color flickering artifacts in the V2V inference outputs.

\subsection{Mimicry on Visual Perceptual Feature Scores}
\label{subsec:Mimicry on Visual Perceptual Feature Scores}
While $\mathcal{L}_{\text{mimic, gen}}$ operates on the distribution of the estimated original latents, in April 2025 we also attempted to investigate whether, by estimating the visual perceptual feature score of the decoded latents in the image space and computing an absolute score loss (e.g., an $L_1$ or $L_2$ loss) between the estimated cinematic aesthetic feature scores, the T2I generation distribution could likewise mimic the T2V generation distribution at the cinematic aesthetic level. 
Specifically, following the same sample construction procedure and configurations as the modality mimic generation loss introduced in \cref{subsec:First Frame Modality Mimic Loss}, we employed an internal cinematic aesthetic feature score model to obtain cinematic aesthetic feature scores from the estimated original images $\widetilde{\mathcal{D}}(\hat{z}^1_{\text{t2v-f},0})$ and $\widetilde{\mathcal{D}}(\hat{z}^1_{\text{t2i},0})$ in the image space, and computed the score loss. 
A higher estimated score value indicates better cinematic aesthetics.
To avoid potential estimation errors caused by the score model (also referred to as a ``reward model'' in some contexts), the modality mimic generation feature score loss is set to zero when the score of the former is not higher than that of the latter. 
To ensure sufficient clarity of the estimated original images in the image space, we restricted the timestep sampling to the non-high-noise regime, specifically within the timestep interval $[15, 860)$.

Let $\text{sc}^1_{\text{t2v-f},0}=\mathcal{S}\bigl(\widetilde{\mathcal{D}}(\hat{z}^1_{\text{t2v-f},0})\bigr)$ and $\text{sc}^1_{\text{t2i},0}=\mathcal{S}\bigl(\widetilde{\mathcal{D}}(\hat{z}^1_{\text{t2i},0})\bigr)$ denote the estimated cinematic aesthetic feature scores respectively, where $\mathcal{S}$ denotes the feature score model. 
We employed the $L_1$ loss to compute the feature score loss.
The modality mimic generation feature score loss was formulated as:
$$
\mathcal{L}_{\text{mimic, gen, score}} = \begin{cases}
\bigl\| \text{sc}^1_{\text{t2v-f},0} - \text{sc}^1_{\text{t2i},0} \bigr\|_1, & \text{if } \text{sc}^1_{\text{t2v-f},0} > \text{sc}^1_{\text{t2i}}, \\
0, & \text{otherwise}.
\end{cases}
$$
Empirically, the cinematic aesthetics of the T2I outputs showed improvement; however, both the convergence speed and the overall performance were markedly weaker than those achieved by $\mathcal{L}_{\text{mimic, gen}}$, which provides a more direct learning signal for the visual model and is not bottlenecked by the capacity of the third-party feature extractor.

We also experimented with using the HPSv2 model~\cite{wu2023human} to estimate visual perceptual feature scores by simultaneously feeding the first frame caption corresponding to $\hat{z}^1_{\text{t2v-f},0}$ into the HPSv2 model, which yielded inferior performance compared to the cinematic aesthetic feature estimation model.
Here, we restricted the timestep sampling to the low-noise regime, specifically within the timestep interval $[80, 390)$.
The feature score model employed here can also be replaced by any feature score model corresponding to the feature level of interest, such as the PickScore model~\cite{kirstain2023pick} and the CLIPScore model.
If multiple modality mimicry sample pairs are constructed within the current training batch—noting that, in April 2025, for the modality mimic generation loss $\mathcal{L}_{\text{mimic, gen}}$ in \cref{eq:39} with KL divergence or MSE as the loss measures, we observed this practice (with the loss for the constructed modality mimicry sample pairs computed in a batched manner) to accelerate convergence to some extent, while the number of sample pairs does not alter the core nature of the method—then it may be possible to treat the score vectors for the video and image modalities as two distinct score features, compute a distributional loss between their softmax-transformed distributions, and thereby obtain an auxiliary score distribution loss.
Furthermore, if score models that can estimate the score distribution for an image are available, distribution measures like KL divergence may also be used to compute the modality mimic generation feature score loss, and the temperature parameter can be employed to scale the logit inputs of the softmax function.

Similarly, one could use a perceptual feature estimation model to extract the visual perceptual features from the decoded estimated original latents $\widetilde{\mathcal{D}}(\hat{z}^1_{\text{t2v-f},0})$ and $\widetilde{\mathcal{D}}(\hat{z}^1_{\text{t2i},0})$ in the image space (e.g., the features extracted by the feature score model itself) and compute the corresponding modality mimic generation feature loss $\mathcal{L}_{\text{mimic, gen, feature}}$ between the estimated visual perceptual features—for instance, using a cosine similarity loss—which shares the same motivation as computing the feature score loss.

\subsection{Mimicry on Attention Features}
\label{subsec:Mimicry on Attention Features}
Previous study~\cite{fahim2025stam} argues that the key and value features of the self-attention layers primarily determine the visual perceptual characteristics of the diffusion model's generation distribution, such as style. 
Therefore, in July 2025, we also attempted to employ the key and value features of the self-attention as the medium for modality mimicry.
Taking the modality mimicry generation as an example, following the same sample construction procedure and configurations as the modality mimic generation loss introduced in \cref{subsec:First Frame Modality Mimic Loss}, we extract the self-attention key and value features from the last DiT block corresponding to the estimated original latents $\hat{z}^1_{\text{t2v-f},0}$ and $\hat{z}^1_{\text{t2i},0}$, respectively.
Denote the self-attention features corresponding to the first frame of the selected T2V sample as $K^1_{\text{t2v-f}}$, $V^1_{\text{t2v-f}}$, and those corresponding to the newly added T2I sample as $K^1_{\text{t2i}}$, $V^1_{\text{t2i}}$. 
Then the modality mimic generation loss with attention features can be formulated as:
\begin{equation}
\mathcal{L}_{\text{mimic, gen, attn}} = \bigl\| K^1_{\text{t2v-f}} - K^1_{\text{t2i}} \bigr\|^2 + \bigl\| V^1_{\text{t2v-f}} - V^1_{\text{t2i}} \bigr\|^2.
\end{equation}
Based on our comparative observations at the time, $\mathcal{L}_{\text{mimic, gen, attn}}$ alone exhibited a relatively noticeable gap in both convergence and performance compared to the original loss $\mathcal{L}_{\text{mimic, gen}}$ adopted in this paper. 
Nevertheless, it may serve as an auxiliary loss. 
In this paper, for the simplicity of the loss functions, we do not employ this loss. 

\subsection{Loss-free Modality Mimicry}
\label{subsec:Loss-free Modality Mimicry}
In July 2025, we also investigated a self-driving form of modality mimicry that does not rely on a dedicated loss function.
Taking the modality mimicry generation as an example, following the same sample construction procedure and configurations as the modality mimic generation loss introduced in \cref{subsec:First Frame Modality Mimic Loss}, we swap the estimated original latent of the first frame of the selected T2V sample with that corresponding to the newly added T2I sample.
Formally, the swapping procedure can be expressed as:
$
\hat{z}^1_{\text{t2v-f},0} \leftrightarrow \hat{z}^1_{\text{t2i},0}.
$
We also exchange the corresponding estimated noise latents:
$
\hat{z}^1_{\text{t2v-f},T} \leftrightarrow \hat{z}^1_{\text{t2i},T}.
$
These two samples are then used only to compute their respective flow matching losses, in the same manner as other samples, without imposing a dedicated modality mimic generation loss.
It is worth noting that, in our implementation, $\hat{v}_t$ in \cref{eq:u} is computed as $\hat{z}_T - \hat{z}_0$, which follows directly from \cref{eq:x0} and \cref{eq:xT}; therefore, the flow matching loss is inherently tied to $\hat{z}_0$ and $\hat{z}_T$.
The overall training objective for the generation step in \cref{eq:54} is then reduced to:
$
\mathcal{L}_{\text{total, gen}} = \mathcal{L}_{\text{FM, global}}.
$
The motivation behind this approach is to encourage the model to learn, in a self-driving manner, to achieve similar performance on samples from the two modalities under the same configuration, thereby encouraging the model's T2I generation ability to mimic its photorealistic T2V counterpart.
Based on our comparative observations at the time, the effectiveness of this method was comparable to that of $\mathcal{L}_{\text{mimic, gen, attn}}$, and thus it can serve as an auxiliary training strategy.

As mentioned previously, the methods introduced in \cref{subsec:Measures for Modality Mimicry}, \cref{subsec:Mimicry on Visual Perceptual Feature Scores}, \cref{subsec:Mimicry on Attention Features}, and \cref{subsec:Loss-free Modality Mimicry} can also be applied to the OPSD of flow matching models for improving the model's performance.

\subsection{Modality Mimicry with On-policy Distillation}
\label{subsec: Modality Mimicry with On-policy Distillation}
In April 2025, while exploring the MSE measure for modality mimicry, we also attempted to investigate whether the original T2V generation capability of the pretrained T2IV model is more suitable as a reference for the T2I generation ability than the updated capability obtained during the post-training procedure. 
To this end, in addition to post-training the pretrained T2IV model, we introduced an additional fixed-weight pretrained T2IV model as a teacher model.
Specifically, in April 2025, following the same sample construction procedure and configurations as the modality mimic generation loss introduced in \cref{subsec:First Frame Modality Mimic Loss}, we additionally duplicated the DiT blocks of the pretrained T2IV model, freezing their parameters and stopping their gradients during training, thereby serving as a teacher-like model $\mathcal{D}_{\text{teacher}}$. 
The terms $\left\{ \mathcal{P} \circ \mathcal{F} (\tilde{z}^1_{\text{t2v, t}}), c^1_{\text{t2v-f}}, t \right\}$ in \cref{eq:30} were also fed into $\mathcal{D}_{\text{teacher}}$, and the modality mimic generation loss was subsequently computed using the MSE measure:
\begin{equation}
\mathcal{L}_{\text{mimic, gen, frozen}} = \bigl\| \mathcal{F}(\hat{z}^1_{\text{t2v-f},0,\text{teacher}}) - \mathcal{F}(\hat{z}^1_{\text{t2i},0}) \bigr\|^2,
\end{equation}
where $\hat{z}^1_{\text{t2v-f},0,\text{teacher}}$ denoted the flattened estimated original first frame latent obtained from $\mathcal{D}_{\text{teacher}}$.
Based on our comparative observations with the pure self-distillation ablation experiment that also employed the MSE measure at the time, this approach also exhibited a certain gap in both convergence and performance compared to the original self-distillation paradigm adopted in this paper. 
Motivated similarly, in a validation experiment based on the original self-distillation paradigm—where we modified \cref{eq:39} by stopping the gradient of $p^1_{\text{t2v-f,0}}$ during training—our comparative observations at the time also indicated a significant degradation in both convergence speed and performance of modality mimicry compared to the original.

Viewed from the perspective of the recently prominent on-policy distillation literature, on the one hand, the first experiment demonstrates the feasibility of on-policy distillation for flow matching models; on the other hand, these two experiments also indicates that the paradigm adopted in this paper—where the generation distributions of two modalities mimic each other during training—is a more appropriate choice for our scenario.
Similarly, the example in \cref{subsec: Modality Self-alignment of Multi-task Models} and the attempted methods in \cref{subsec:Measures for Modality Mimicry}, \cref{subsec:Mimicry on Visual Perceptual Feature Scores}, \cref{subsec:Mimicry on Attention Features}, \cref{subsec:Loss-free Modality Mimicry} can be adapted into an on-policy distillation post-training scenario. 
For instance, one could employ a frozen pretrained multimodal model as an I2T teacher to enhance the prompt faithfulness of the T2I task in its trainable counterpart.

\begin{figure*}[htpb]
    \centering
    \includegraphics[width=1.0\linewidth, trim=0pt 0pt 0pt 0pt, clip]{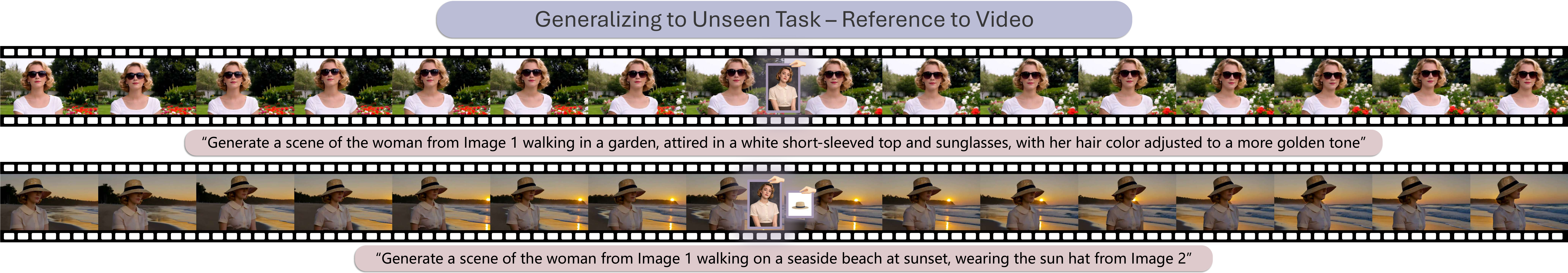} 
    \captionsetup{font=scriptsize}
    \caption{
        \textbf{Reference-to-Video (R2V) results with 61 frames and a resolution of $832\times 480$ generated at inference.}
        Given an editing prompt and reference images, our model demonstrates a degree of ability to generate the corresponding video in a natural, prompt-following manner, conditioned on the editing prompt and the visual information provided by the reference images.
        It should be noted that the training data of the demonstrated model in this paper does not contain any R2V samples or global R2I samples.
        We consider that the model's capability to generate videos conditioned on reference visual information and a given caption may arise from FlowMimic implicitly transferring the ability to generate prompt-following motion—acquired from training on T2V and I2V tasks—to the unseen R2V task. 
        Moreover, our pixel-wise temporal warped flow field enables the model to learn how to maintain the reference visual information consistently over time from the online generated multi-reference video editing samples. 
        Together, these two abilities allow the model to demonstrate a certain ability to preserve the reference visual information across generated video frames while generating natural and plausible actions in a prompt-following manner.
        Besides, the first example demonstrates that the model has implicitly learned to employ its visual editing ability to the unseen R2V task, for instance by changing the clothes of the subject and adding sunglasses to the subject appearing in the reference visual information.
    }
    \label{fig:flowmimic_fig_33}
\end{figure*}

\subsection{Perceptual Guidance from Clean Images}
\label{subsec:Perceptual Guidance from Clean Images}
In July 2025, we attempted to further enhance the model's stylized generation capability for both video and image modalities, building upon the method described in \cref{subsec:From stylized T2I to stylized T2V}. 
To this end, similar to the motivation behind \cref{subsec:Mimicry on Visual Perceptual Feature Scores}, we employed a style perceptual feature loss computed in the image space.

Specifically, during the post-training procedure, we temporarily duplicate each T2I training image into a multi-frame video for T2V training. 
For each such image-video sample pair, let $I$ and $V$ denote the T2I and T2V samples in the image space, and let $\widetilde{\mathcal{D}}(\hat{z}_{\text{t2i},0})$ and $\widetilde{\mathcal{D}}(\hat{z}_{\text{t2v},0})$ denote the corresponding decoded estimated original latents in the image space, respectively. 
We randomly sample a single frame from the first four frames of $\widetilde{\mathcal{D}}(\hat{z}_{\text{t2v},0})$—say, the $i$-th frame $\widetilde{\mathcal{D}}(\hat{z}_{\text{t2v},0})_i$—as the observation frame. 
Using the style perceptual feature estimation model CSD~\cite{somepalli2024measuring}, we estimate the style perceptual features of $I$, $\widetilde{\mathcal{D}}(\hat{z}_{\text{t2i},0})$, and $\widetilde{\mathcal{D}}(\hat{z}_{\text{t2v},0})_i$. 
The style perceptual feature loss is then defined as the cosine similarity loss between the features of the clean image and each of the two decoded latents:
\begin{equation}
\begin{split}
\mathcal{L}_{\text{guide, style}} &= 1 - \frac{1}{2} \Bigl[ \text{cos}\bigl( \phi(I),\, \phi(\widetilde{\mathcal{D}}(\hat{z}_{\text{t2i},0})) \bigr) \\
&+ \text{cos}\bigl( \phi(I),\, \phi(\widetilde{\mathcal{D}}(\hat{z}_{\text{t2v},0})_i) \bigr) \Bigr],
\end{split}
\end{equation}
where $\phi$ denotes the CSD model. 
To ensure sufficient clarity of the estimated original images in the image space, we restricted the timestep sampling to the non-high-noise regime, specifically within the timestep interval $[0, 900)$.

Based on our comparative observations at the time, this loss accelerated the convergence of stylistic responsiveness for stylized T2V. 
In terms of T2I performance, it enhanced the style accuracy (i.e., how closely the visual style of the generated image matches the style keywords specified in the prompt) and aesthetics of stylistic responsiveness, while also, somewhat unexpectedly, improving the aesthetics and prompt faithfulness of generations under non-stylized prompts.
We consider that this method can also be generalized to other feature levels of interest.
Besides, for methodological simplicity, we did not adopt this incremental method in this paper.

\subsection{Additional Reference Injection}
\label{subsec:Additional Reference Injection}
In August 2025, beyond the reference-inject self-attention mask described in \cref{subsec:Model Architecture Adaptation to Editing Task}, we also experimented with another method to strengthen reference information injection via cross-attention. We briefly introduce below.
For each reference element, we extract a per-reference visual representation by taking its first frame and encoding it with two pretrained encoders: Dinov2-giant~\cite{oquab2023dinov2} and OpenCLIP-xlm-roberta-large-vit-huge~\cite{radford2021learning}, whose outputs are concatenated to form a single reference visual feature. 
For the target element, the corresponding feature is set to a zero-valued feature of the same shape as the reference visual feature.
Each element's visual feature is then passed through a zero-initialized MLP with negligible parameters to map it into the cross-attention embedding dimension of the pretrained T2IV model. 
Subsequently, additional projection layers corresponding to the visual features project them into query and key features respectively.
An element-wise cross-attention mask restricts attention so that each element can attend only to its own visual feature; the resulting cross-attention output is then added to the original text-conditioned cross-attention output to produce the element's final cross-attention stream.
During training, we drop out each reference visual feature with a small probability. At inference, we include the visual classifier-free guidance (CFG)~\cite{ho2022classifier} to the original text CFG.
Although this cross-attention injection improved reference preservation to some extent, we considered that it departed from our design principle of keeping the model architecturally clean. Accordingly, the model in this paper does not include this design.

\subsection{Reference-to-Video task}
\label{subsec:Reference-to-Video task}
In September and October 2025, we attempted the Reference-to-Video (R2V) task and enabled the model to exhibit the corresponding ability.
Specifically, for a global Reference-to-Image (R2I) image sample, we employed the pixel-pair temporal warped flow field to generate a motion video with static or dynamic camera movements from the target image. 
This video served as the target element for the R2V video sample, and motion captions were incorporated into the original target caption. 
Here, the global R2I task referred to generating a target image according to the reference images and a global editing instruction—for example, ``Place the woman from Image 1 and the child from Image 2 together on a beach'', ``Change the camera angle of Image 1 to a top-down view'' or ``Generate an exploded-view diagram of the toy model shown in Image 1''. 
Following the same sample construction procedure and configurations as the modality mimic editing loss introduced in \cref{subsec:First Frame Modality Mimic Loss}, we also applied the modality mimic editing loss to the R2V samples and the corresponding newly added R2I samples.

Simultaneously, for a T2V sample, we used a YOLO~\cite{redmon2016you}-based detection model during training to detect the subjects in the first frame online, cropped the corresponding bounding box region, and then applied rotation or flipping to it. 
The processed results were used as the reference elements for the Subject-to-Video (S2V)~\cite{liu2025phantom,deng2025cinema} sample—a special form of R2V sample—while the latent frames from the subsequent five latent frames were used as the target element. 

It is worth noting that, during the exploratory stage, we observed that the model exhibited a certain degree of R2V capability regardless of whether the R2V task and global R2I task were excluded or included without using S2V samples constructed from T2V samples.
For example, the results in \cref{fig:flowmimic_fig_33} demonstrate that when R2V and global R2I training samples were not included, the model still exhibited a certain degree of R2V capability.
This suggests that our pixel-pair temporal warped flow field enables the model not only to acquire video editing capabilities from image editing samples, but also to generalize—through joint training with the T2V task—to tasks such as R2V, which demand both temporal editing and temporal generation abilities. 
Furthermore, under the condition where the R2V task and global R2I task were included without using S2V samples constructed from T2V samples, we conducted an ablation experiment on the video editing samples, R2V samples and the modality mimic editing loss. 
It was observed that they improved the editing quality—for example, the aesthetics of stylization, relighting and color tone adjustment—as well as the identity preservation of the subjects in the reference images, for the model's image editing performance at inference time.

In this paper, to verify the effectiveness of the pixel-pair temporal warped flow field in enabling the model to learn video editing capabilities from image editing samples alone, the R2V task was not included during training. This was to prevent the R2V samples, which were partially constructed from T2V samples, from potentially conferring video editing abilities to the model and thus introducing confounding factors.
Meanwhile, we observed that there was occasional confusion between the editing instructions and the target images in the global R2I samples—for instance, an editing instruction specifying a local editing while the target image corresponded to a global editing effect. 
Therefore, to maintain data cleanliness, the model demonstrated in this paper did not employ the global R2I training samples.

\subsection{Addressing Excessive Mimicry Parameters}
\label{subsec:Addressing Excessive Mimicry Parameters}
In April 2025, we intensified the influence of the modality mimic generation loss in various ways to observe phenomena under extreme hyperparameter settings—for example, increasing the modality mimic generation loss weight by orders of magnitude, setting $n_{\text{mimic,gen}}$ to 1, incorporating the additional distribution measures—namely, the Jensen-Shannon (JS) divergence and the Hellinger distance—as loss functions, and substantially increasing the number of constructed modality mimicry sample pairs at each modality mimic generation step, with the modality mimic generation loss subsequently computed in a batched manner between the two groups of sample pairs corresponding to the video and image modalities, respectively.
Specifically, these multiple modality mimicry sample pairs could be constructed either from different T2V samples, each paired with a newly added T2I sample derived from its first frame, or from different latent frames of the same T2V sample (obtained via different sampling strategies), each paired with a newly added T2I sample derived from that sample's first frame. In the latter case, the number of sample pairs was capped at three-quarters of the minimum between the number of T2I samples and the number of T2V latent frames. 

For brevity of notation, in the following of this subsection, we still formulate under the setting where a single modality mimicry sample pair is constructed at each modality mimicry step.
When all of the exemplified extreme hyperparameter settings were in place and the modality mimic generation loss weight exceeded an extreme threshold, although the convergence speed and performance of modality mimicry improved, the logit inputs of the softmax function $\mathcal{F}(\hat{z}^1_{\text{t2v-f},0})$ and $\mathcal{F}(\hat{z}^1_{\text{t2i},0})$ in \cref{eq:37} occasionally exhibited a value at a same particular logit entry position (e.g., both at the fifth logit entry position) that was much larger than the values at all other logit entry positions, and these two values were also numerically close to each other.

From the model's perspective, this reduces $\mathcal{L}_{\text{mimic, gen}}$ in \cref{eq:39}, because it causes both $p^1_{\text{t2v-f,0}}$ and $p^1_{\text{t2i,0}}$ in \cref{eq:37} to become peaked distributions at the same particular logit entry position, making the two distributions similar to each other and thus yielding a relatively small value of $\mathcal{L}_{\text{mimic, gen}}$; however, this is not the desired natural learning procedure. 
We first introduce several common approaches, such as applying temperature scaling to the logit inputs, normalizing the logit inputs, using a regularization loss to smooth the logit inputs, and imposing an entropy maximization loss on the probability distribution $p^1_{\text{t2v-f,0}}$.
\begin{itemize}
    \item Temperature scaling. The logit inputs are divided by a temperature parameter $\tau$, which is either held constant or annealed from an initial value $\tau_s$ to a final value $\tau_e$ (with $\tau_s > \tau_e$) according to a linear or cosine schedule: $\widetilde{\mathcal{F}}(\hat{z}^1_{\text{t2v-f},0})=\frac{\mathcal{F}(\hat{z}^1_{\text{t2v-f},0})}{\tau}, \widetilde{\mathcal{F}}(\hat{z}^1_{\text{t2i},0}) = \frac{\mathcal{F}(\hat{z}^1_{\text{t2i},0})}{\tau}$. $\mathcal{L}_{\text{mimic, gen}}$ is then further multiplied by $\tau^2$ to preserve scale.
    \item Normalizing the logit inputs. Let $\mu$ and $\sigma$ denote the mean and standard deviation of $\mathcal{F}(\hat{z}^1_{\text{t2v-f},0})$. Then $\widetilde{\mathcal{F}}(\hat{z}^1_{\text{t2v-f},0})=\frac{\mathcal{F}(\hat{z}^1_{\text{t2v-f},0}) - \mu}{\sigma + \varepsilon_{\text{pert}}}$, and analogously for $\widetilde{\mathcal{F}}(\hat{z}^1_{\text{t2i},0})$. $\varepsilon_{\text{pert}}=1 \times 10^{-8}$ is a small perturbation.
    \item Regularization loss to smooth the logit inputs. This regularization loss is defined as $\mathcal{L}_{\text{mimic, gen, reg, t2v-f}}=\bigl(\max\limits_{j} | \mathcal{F}(\hat{z}^1_{\text{t2v-f},0})_j | \bigr)^2$, where $j$ indexes over the flattened feature dimensions. And analogously for $\mathcal{L}_{\text{mimic, gen, reg, t2i}}$.
    \item Entropy maximization loss. The entropy of the distribution $p^1_{\text{t2v-f,0}}$ is maximized via $\mathcal{L}_{\text{mimic, gen, entropy}} = -\alpha_{\text{adaptive}} \cdot H(p^1_{\text{t2v-f,0}})$, where $H(p^1_{\text{t2v-f,0}}) = -p^1_{\text{t2v-f,0}} \log p^1_{\text{t2v-f,0}}$, $\alpha_{\text{adaptive}}$ is an adaptive weight inversely proportional to $H(p^1_{\text{t2v-f,0}}) \cdot \tau$, and $\tau$ is the temperature described in temperature scaling. In particular, for the normalization, regularization loss, and entropy maximization loss, we also experimented with keeping only $\widetilde{\mathcal{F}}(\hat{z}^1_{\text{t2i},0})$ and $\mathcal{L}_{\text{mimic, gen, reg, t2i}}$, or replacing $H(p^1_{\text{t2v-f,0}})$ with the corresponding $H(p^1_{\text{t2i,0}})$.
    \item Clamp and renormalize the probability distribution. For the computation of the KL divergence measure, we first apply clamping and renormalization to \(p^1_{\text{t2v-f},0}\):
    $$
    \widetilde{p}^1_{\text{t2v-f},0} = \operatorname{normalize}\bigl(\operatorname{clamp}(p^1_{\text{t2v-f},0};\,1\times10^{-10})\bigr),
    $$
    where \(\operatorname{clamp}(\cdot;1\times10^{-10})\) sets each entry below \(1\times10^{-10}\) to \(1\times10^{-10}\), and \(\operatorname{normalize}(\cdot)\) rescales the vector to sum to one. For the computation of the JS divergence measure detailed later in \cref{subsec:Details of the Distribution Measures}, we similarly first apply clamping and renormalization to \(m\).
    \item Exclusion of the flow matching loss for the modality mimicry pairs. During the modality mimicry generation step, we exclude the flow matching loss corresponding to $z^1_{\text{t2v-f}}$ and $z^1_{\text{t2i}}$.
    \item Continuous distributional measures. As a simplifying approximation, we treat each estimated original latent as an isotropic Gaussian rather than a full discrete distribution. Let $\mu_{\text{t2v-f}}$ and $\sigma_{\text{t2v-f}}^2$ denote the mean and variance of $\mathcal{F}(\hat{z}^1_{\text{t2v-f},0})$, respectively, and let $\mu_{\text{t2i}}$ and $\sigma_{\text{t2i}}^2$ denote those of $\mathcal{F}(\hat{z}^1_{\text{t2i},0})$. The continuous forms of the distribution measures described in this study are computed as:
    $$
    \left\{
    \begin{aligned}
    & D_{\text{KL,c}}\bigl(p^1_{\text{t2v-f},0} \,\big\|\, p^1_{\text{t2i},0}\bigr) \\
    = &\; 
    \frac{1}{2}\Bigl(
        \ln\frac{\sigma_{\text{t2i}}^2}{\sigma_{\text{t2v-f}}^2}
        + \frac{\sigma_{\text{t2v-f}}^2 + (\mu_{\text{t2v-f}} - \mu_{\text{t2i}})^2}{\sigma_{\text{t2i}}^2}
        - 1
    \Bigr), \\[12pt]
    & D_{\text{JS-r,c}}\bigl(p^1_{\text{t2v-f},0} \,\big\|\, p^1_{\text{t2i},0}\bigr) \\
    = &\; \frac{1}{2} \, D_{\text{KL,c}}\!\bigl(m \,\big\|\, p^1_{\text{t2v-f},0}\bigr)
    + \frac{1}{2} \, D_{\text{KL,c}}\!\bigl(m \,\big\|\, p^1_{\text{t2i},0}\bigr), \\[12pt]
    & H_{c}\bigl(p^1_{\text{t2v-f},0},\, p^1_{\text{t2i},0}\bigr) \\
    = &\; 
    \Biggl[
        1 - \Biggl(
            \frac{2\sqrt{ \sigma_{\text{t2v-f}}^2 \sigma_{\text{t2i}}^2 }}
                { \sigma_{\text{t2v-f}}^2 + \sigma_{\text{t2i}}^2 }
        \Biggr)^{\frac{1}{2}}
        \exp\!\Biggl(
            -\frac{(\mu_{\text{t2v-f}} - \mu_{\text{t2i}})^2}
                {4(\sigma_{\text{t2v-f}}^2 + \sigma_{\text{t2i}}^2)}
        \Biggr) \\
        &\qquad + \kappa_H
    \Biggr]^{\frac{1}{2}},
    \end{aligned}
    \right.
    $$
    where \( m = \frac{1}{2}(p^1_{\text{t2v-f},0} + p^1_{\text{t2i},0}) \) is the mixture Gaussian with mean \( \mu_m = \frac{1}{2}(\mu_{\text{t2v-f}} + \mu_{\text{t2i}}) \) and variance \( \sigma_m^2 = \frac{1}{4}\bigl[\sigma_{\text{t2v-f}}^2 + \sigma_{\text{t2i}}^2 + (\mu_{\text{t2v-f}} - \mu_{\text{t2i}})^2\bigr] \), and \( \kappa_H = 1 \times 10^{-8} \).
\end{itemize}
The above strategies were primarily intended to prevent the probability distribution of the logit inputs from tending towards a sharp, single-point distributions, yet none produced a discernible effect in our experiments.
For instance, a large constant temperature can avoid the aforementioned phenomenon, but it leads to blurred generation results during inference; conversely, a smaller temperature fails to produce a noticeable mitigating effect.
After conducting experiments and analyses on the distribution measures—namely, the KL divergence, the JS divergence, and the Hellinger distance—as well as on the operations related to the softmax function, we devised the additive-dimension strategy, which can eliminate this phenomenon whilst preserving the original modality mimicry effect and introducing no adverse side effects during inference.

Specifically, we appended an extra dimension to both $\mathcal{F}(\hat{z}^1_{\text{t2v-f},0})$ and $\mathcal{F}(\hat{z}^1_{\text{t2i},0})$, setting its value to $k$ times the observed maximum value, denoted as $c_{\text{add}}$, and kept this value fixed throughout training. 
The value of $k$ depends on the training scenario: $k = 1.3$ if we continue training from a model that has already exhibited the described phenomenon, and $k = 1.1$ if we commence post-training directly from the pretrained T2IV model.
Formally, we define:
\begin{equation}
\begin{cases}
\widetilde{\mathcal{F}}(\hat{z}^1_{\text{t2v-f},0}) = \bigl[\mathcal{F}(\hat{z}^1_{\text{t2v-f},0}), c_{\text{add}}\bigr], \\[6pt]
\widetilde{\mathcal{F}}(\hat{z}^1_{\text{t2i},0}) = \bigl[\mathcal{F}(\hat{z}^1_{\text{t2i},0}), c_{\text{add}}\bigr].
\end{cases}
\end{equation}

Subsequently, $\widetilde{\mathcal{F}}(\hat{z}^1_{\text{t2v-f},0})$ and $\widetilde{\mathcal{F}}(\hat{z}^1_{\text{t2i},0})$ are used to compute $\mathcal{L}_{\text{mimic, gen}}$ following \cref{eq:37}, \cref{eq:38} and \cref{eq:39}. 
Accordingly, the softmax logit inputs for the JS divergence and the Hellinger distance measures are also replaced by $\widetilde{\mathcal{F}}(\hat{z}^1_{\text{t2v-f},0})$ and $\widetilde{\mathcal{F}}(\hat{z}^1_{\text{t2i},0})$.
After applying the additive-dimension strategy, the described sharp value pair in $\mathcal{F}(\hat{z}^1_{\text{t2v-f},0})$ and $\mathcal{F}(\hat{z}^1_{\text{t2i},0})$ gradually diminished until they disappeared completely (if continuing training from a model that has already exhibited the phenomenon) or never emerged at all (if post-training from the pretrained T2IV model); that is, the constant $c_{\text{add}}$ enables the softmax function to locate a logit entry position with extremely high probability from the very beginning, ensuring that the loss value remains relatively small, and thus allows other logits tokens to learn gently and without undue pressure even under extremely large loss weights.
Whilst such extreme settings receive little attention in practice, and the described phenomenon may occur only under these extreme conditions, we nevertheless briefly describe the method here, as we consider it possesses some probabilistic interest.

\subsection{Details of the Distribution Measures}
\label{subsec:Details of the Distribution Measures}
We detail the Jensen-Shannon (JS) divergence and the Hellinger distance described in this study, taking the modality mimicry generation as an example.

The JS divergence was employed in its reverse form. 
Given the two probability distributions $p^1_{\text{t2v-f},0}$ and $p^1_{\text{t2i},0}$, this distribution measure was computed as:
$$
\begin{aligned}
&D_{\text{JS-r}}\bigl(p^1_{\text{t2v-f},0} \,\big\|\, p^1_{\text{t2i},0}\bigr) \\
= &\; \frac{1}{2} \, D_{\text{KL}}\!\bigl(m \,\big\|\, p^1_{\text{t2v-f},0}\bigr) + \frac{1}{2} \, D_{\text{KL}}\!\bigl(m \,\big\|\, p^1_{\text{t2i},0}\bigr),
\end{aligned}
$$
where \(m = \frac{1}{2}\bigl(p^1_{\text{t2v-f},0} + p^1_{\text{t2i},0}\bigr)\). 
The logarithms of the probability distributions $p^1_{\text{t2v-f},0}$ and $p^1_{\text{t2i},0}$ were computed after adding a small constant of $1 \times 10^{-8}$, respectively.

The Hellinger distance quantified the similarity between two probability distributions through the Bhattacharyya coefficient:
$$
H\bigl(p^1_{\text{t2v-f},0},\, p^1_{\text{t2i},0}\bigr) 
= \left(
1 - \sum_{j=1}^D \left( p_{\text{t2v-f},0}^1(j) \, p_{\text{t2i},0}^1(j) \right)^{\frac{1}{2}}
+ \kappa_{\text{H}}
\right)^{\frac{1}{2}},
$$
where the summation $\sum\limits_{j=1}^D \left( p_{\text{t2v-f},0}^1(j) \, p_{\text{t2i},0}^1(j) \right)^{\frac{1}{2}}$ constituted the Bhattacharyya coefficient, and $\kappa_{\text{H}} = 1 \times 10^{-8}$ was a small constant.  
Before the inner square root, each entry of the probability distributions was clamped to be no less than $1 \times 10^{-10}$; before the outer square root, the squared Hellinger distance was clamped to be non-negative.
When these distribution measures, the JS divergence and Hellinger distance, were additionally incorporated as loss functions for modality mimicry, $D_{\text{JS-r}}\bigl(p^1_{\text{t2v-f},0} \,\big\|\, p^1_{\text{t2i},0}\bigr)$ and $H\bigl(p^1_{\text{t2v-f},0},\, p^1_{\text{t2i},0}\bigr)$ were added in a weighted form to $\mathcal{L}_{\text{mimic, gen}}$ in \cref{eq:39}.
Besides, as previously described, when multiple modality mimicry sample pairs were constructed within the current training batch, the modality mimic generation loss was computed in a batched manner. 
Specifically, for the Hellinger distance, the Bhattacharyya coefficient was further averaged over the modality mimicry sample pair dimension during its computation.

\subsection{Miscellaneous Insights}
\label{subsec:Miscellaneous Insights}
We briefly describe a subset of the attempted methods that could not be individually ablated due to limited computational resources, in the hope that they may provide potential insights for related work.

\paragraph{Configurations of Modality Mimic Feature Score Loss.}
\label{para:Configurations of Modality Mimic Feature Score Loss.}
In April 2025, for the modality mimic generation feature score loss described in \cref{subsec:Mimicry on Visual Perceptual Feature Scores}, which was an image space loss, we also experimented with the MSE measure:
$$
\begin{aligned}
& \mathcal{L}_{\text{mimic, gen, score, mse}} \\
= &\; \begin{cases}
\bigl\| \text{sc}^1_{\text{t2v-f},0} - \text{sc}^1_{\text{t2i},0} \bigr\|^2, &\text{if } \text{sc}^1_{\text{t2v-f},0} > \text{sc}^1_{\text{t2i},0}, \\
0, &\text{otherwise}.
\end{cases}
\end{aligned}
$$
Besides, we experimented with detaching the gradients of $\text{sc}^1_{\text{t2v-f},0}$.
Additionally, when multiple modality mimicry sample pairs are constructed within the current training batch, we experimented with computing the loss for the constructed pairs in a sample-wise manner rather than a batched manner. 
Specifically, the sample-wise MSE over the batch is defined as: $\sum\limits_{i=1}^{N} \mathcal{L}_{\text{mimic, gen, score, mse}}^{(i)}$, where $N$ is the number of modality mimicry sample pairs in the current training batch, and $\mathcal{L}_{\text{mimic, gen, score, mse}}^{(i)}$ is the modality mimic generation feature score loss corresponding to the $i$-th sample pair; that is, the sample-wise MSE does not involve averaging over the modality mimicry sample pair batch size $N$.

\paragraph{Frame-wise Noise Offset.}
\label{para:Frame-wise Noise Offset.}
In July 2025, to strengthen the decoupling between spatial and temporal modelling in T2V generation, following common practice, for the T2V samples in the generation step, we incorporate a spatial perturbation into the original flow matching loss as a decoupling loss. 
Specifically, we add a Gaussian noise offset perturbation \(\Delta\) of shape \((1 , \lfloor \frac{H}{8} \rfloor , \lfloor \frac{W}{8} \rfloor , 1)\), broadcasted to \((N+1 , \lfloor \frac{H}{8} \rfloor , \lfloor \frac{W}{8} \rfloor , 16)\) along the frame and channel dimensions, to the original noise \(\epsilon\), and take the resulting loss as a decoupling loss:
$$
\epsilon' = \epsilon + \Delta, \quad \Delta \sim \mathcal{N}(0, \mathbf{I}).
$$
The decoupling loss is then computed as:
$$
\begin{aligned}
& \mathcal{L}_{\text{gen, decoupling}} \\
= &\; \bigl\| (\hat{z}_{\text{t2v},T} - \hat{z}_{\text{t2v},0}) - (\epsilon' - z_{\text{t2v}}) \bigr\|^2.
\end{aligned}
$$
We also experimented with a configuration where the decoupling loss was additionally applied to the T2I samples, while the original flow matching loss for the T2IV samples was omitted.

\paragraph{From Single Frame to Sub-clip.}
\label{para:From Single Frame to Sub-clip.}
In July 2025, when constructing modality mimicry sample pairs during the modality mimic generation step, we generalized the newly added T2I sample to a newly added T2V sub-sample. 
The distinction is that the latter consists of the initial latent frames of the original T2V sample rather than just the first frame. 
Let the latent representation of the original T2V sample be \(z^1_{\text{t2v}}\). 
We construct a sub-sample latent \(z^1_{\text{t2v, sub}}\) where the number of frames $k$ is randomly drawn from \(\{1,2,3,4\}\).

We maintain a frozen, gradient-detached copy of the DiT blockes \(\mathcal{D}\), denoted \(\mathcal{D}_{co}\), which is synchronized at each optimization step with the parameters of the trainable blockes \(\mathcal{D}\). 
\(z^1_{\text{t2v, sub}}\) is passed through \(\mathcal{D}_{co}\), yielding the estimated original latents \(\hat{z}^1_{\text{t2v, sub, 0}}\).
The auxiliary modality mimicry generation loss \(\mathcal{L}_{\text{mimic, gen, aux}}\) is then computed as the mean squared error between the estimated original latents corresponding to \(z^1_{\text{t2v}}[:k]\) and \(z^1_{\text{t2v, sub}}\):
$$
\mathcal{L}_{\text{mimic, gen, aux}} = \bigl\| \hat{z}^1_{\text{t2v, 0}}[:k] - \hat{z}^1_{\text{t2v, sub, 0}} \bigr\|^2.
$$

\paragraph{Configurations of Cross-attention Masks.}
\label{para:Configurations of Cross-attention Masks.}
In August 2025, we explored different cross-attention mask designs in multi-reference editing tasks beyond the element-wise one introduced in this paper.
Specifically, we implemented the 1D text RoPE described below as an optional component. 
We describe some of the scenarios in which this option is employed:
\begin{itemize}
    \item Sample-wise cross-attention mask with 1D text RoPE. In contrast to the element-wise cross-attention mask, the sample-wise cross-attention mask allows all visual tokens within a sample to attend to the concatenated text embeddings corresponding to all elements within the sample. Additionally, before the cross-attention output is computed, a 1D RoPE is applied along the element dimension to the text embeddings of each element caption, thereby enabling the model to distinguish between text embeddings originating from different elements within the sample.
    \item Target-inclined cross-attention mask with 1D text RoPE. Similar to the reference-inject self-attention mask, each reference element's visual tokens are constrained to attend exclusively to their corresponding text embeddings, while the visual tokens of the target element attend to the concatenated text embeddings corresponding to all elements within the sample. Likewise, before the cross-attention output corresponding to the target visual tokens is computed, a 1D RoPE is applied along the element dimension to the text embeddings of each element caption.
    \item Disjoint reference-target cross-attention mask with 1D text RoPE. In this configuration, the reference visual tokens are constrained to attend exclusively to the concatenated text embeddings corresponding to the reference elements, while target visual tokens attend exclusively to the output text embeddings. Similarly, before the cross-attention output is computed, a 1D RoPE is applied along the element dimension to the text embeddings of each element caption. When applying this 1D text RoPE, we experimented with treating the output caption either as a single unified caption or as two independent captions—namely, the editing caption and the target caption—with the text embeddings of each encoded separately.
\end{itemize}

\paragraph{Configurations of Self-attention Masks.}
\label{para:Configurations of Self-attention Masks.}
In early September 2025, aside from the reference-inject self-attention mask introduced in \cref{para:Reference-inject Self-attention Mask.}, in which each reference element attends only to itself while the target element attends to the reference and target elements of the same sample, we investigated an additional self-attention configuration for multi-reference samples:
\begin{itemize} 
    \item Inter-reference self-attention mask. All visual tokens belonging to the reference elements within a sample are allowed to attend to each other, while attention to the target visual tokens is masked out. Meanwhile, the target visual tokens continue to attend to the full sample.
\end{itemize}

\paragraph{Configurations of Element Identity Encoding.}
\label{para:Configurations of Element Identity Encoding.}
In early September 2025, in addition to the two element identity encoding methods introduced in \cref{para:Element Identity Encoding.}—both of which are applied at every self-attention layer—we explored an additional variant of element identity encoding designed for multi-reference samples:
\begin{itemize}
    \item Single-use absolute positional encoding. Under the reference-inject self-attention mask described in \cref{para:Reference-inject Self-attention Mask.}, the absolute positional encoding introduced in \cref{para:Element Identity Encoding.}, which represents element identity, is applied only once, before the visual tokens enter the DiT blocks, rather than at every self-attention layer (applied to the query and key features respectively).
\end{itemize}

\paragraph{Configurations of RoPE.}
\label{para:Configurations of RoPE.}
In August 2025, we explored RoPE designs beyond the separate 3D RoPE with frame correspondence introduced in \cref{para:Separate 3D RoPE with Frame Correspondence.}, intended for scenarios such as global editing and reference-based video generation, including tasks such as the single-reference or multi-reference reference-to-video task described in \cref{subsec:Reference-to-Video task}.
We briefly summarize a subset of these designs below.
\begin{itemize}
\item 3D RoPE configuration (i).
We devised an asymmetric two-sided design based on a different RoPE layout. 
Let $i = 1,2,\dots,n$ enumerate the reference elements. A coordinate symbol with subscript $x$ (e.g., $t_x$) denotes the global index of a visual token in the shared frequency table, and the same symbol with subscript $x,o$ (e.g., $t_{x,o}$) denotes its local, origin-based coordinate within the corresponding element. Throughout this section, $t_{\mathrm{tar}}$, $h_{\mathrm{tar}}$, and $w_{\mathrm{tar}}$ denote the temporal, height, and width counts of the target element, while $t_{\text{ref}}^{(j)}$, $h_{\text{ref}}^{(j)}$, and $w_{\text{ref}}^{(j)}$ denote the temporal, height, and width counts of the $j$-th reference element:
$$
\begin{cases}
t_x^{(i)} = t_{x,o}^{(i)},\\[4pt]
h_x^{(i)} = h_{x,o}^{(i)} + h_{\mathrm{tar}} + \displaystyle\sum_{j=1}^{i-1} h_{\text{ref}}^{(j)},\\[8pt]
w_x^{(i)} = w_{x,o}^{(i)} + w_{\mathrm{tar}} + \displaystyle\sum_{j=1}^{i-1} w_{\text{ref}}^{(j)},
\end{cases}
$$
wrapping back to $0$ whenever the accumulated spatial offset exceeds the corresponding table segment. 
The target temporal band begins at the configurable offset $f_o$ ($t_x = f_o + t_{x,o}$), while its spatial indices start at $0$. 
By default, $f_o$ is set to $0$. We also experimented with $f_o$ values of $1$, $600$, and $800$.

\item 3D RoPE configuration (ii).
The target starts at the origin $(0,0,0)$. The references are stacked entirely after the target on all three axes: their temporal band begins at $t_{\mathrm{tar}} + f_o$ with an internal offset that commences at $f_{\text{off}}^{(1)} = 0$ and accumulates as $f_{\text{off}}^{(i)} = \sum\limits_{j=1}^{i-1}\bigl(t_{\text{ref}}^{(j)} + f_o\bigr)$ per reference before any reset, resetting to $0$ when out of bounds; their spatial indices accumulate from $h_{\mathrm{tar}}$ and $w_{\mathrm{tar}}$ without extra gaps, wrapping back to $0$:
$$
\begin{cases}
t_x^{(i)} = t_{\mathrm{tar}} + f_o + f_{\text{off}}^{(i)} + t_{x,o}^{(i)},\\[4pt]
h_x^{(i)} = h_{x,o}^{(i)} + h_{\mathrm{tar}} + \displaystyle\sum_{j=1}^{i-1} h_{\text{ref}}^{(j)} ,\\[8pt]
w_x^{(i)} = w_{x,o}^{(i)} + w_{\mathrm{tar}} + \displaystyle\sum_{j=1}^{i-1} w_{\text{ref}}^{(j)} .
\end{cases}
$$
We experimented with $f_o$ values of $1$, $5$, and $15$.

\item 3D RoPE configuration (iii).
The target starts at the origin $(0,0,0)$. 
All references share a single fixed temporal starting point $t_{\mathrm{tar}} + f_o$ (no accumulation across elements), while their spatial indices accumulate from the far end of the target's spatial extent, with boundary wrapping back to $0$:
$$
\begin{cases}
t_x^{(i)} = t_{\text{tar}} + f_o + t_{x,o}^{(i)},\\[4pt]
h_x^{(i)} = h_{x,o}^{(i)} + h_{\text{tar}} + \displaystyle\sum_{j=1}^{i-1} h_{\text{ref}}^{(j)} ,\\[8pt]
w_x^{(i)} = w_{x,o}^{(i)} + w_{\text{tar}} + \displaystyle\sum_{j=1}^{i-1} w_{\text{ref}}^{(j)} .
\end{cases}
$$
We experimented with $f_o$ values of $0$ and $15$.

\item 3D RoPE configuration (iv).
The target's temporal, height, and width indices all start at the same configurable offset $f_o$, i.e., $t_x = f_o + t_{x,o}$, $h_x = f_o + h_{x,o}$, $w_x = f_o + w_{x,o}$. Each reference's temporal index restarts from $0$ ($t_x^{(i)} = t_{x,o}^{(i)}$). Spatially, all reference elements are placed at the same location—the bottom-right corner of the target area shifted by $f_o$—with no cumulative offset between references, so they completely overlap:
$$
\begin{cases}
t_x^{(i)} = t_{x,o}^{(i)},\\[4pt]
h_x^{(i)} = h_{x,o}^{(i)} + h_{\text{tar}} + f_o,\\[8pt]
w_x^{(i)} = w_{x,o}^{(i)} + w_{\text{tar}} + f_o.
\end{cases}
$$

\item 3D RoPE configuration (v).
The target starts at the origin $(0,0,0)$. The reference temporal band follows the same stacked-after-target rule as in configuration (ii) (beginning at $t_{\mathrm{tar}} + f_o$, advancing by $t_{\text{ref}}^{(i)} + f_o$, and resetting to $0$ when out of bounds). The reference spatial indices, however, restart from $0$ for every reference, independent of the target, with no accumulation between references:
$$
\begin{cases}
t_x^{(i)} = t_{\text{tar}} + f_o + f_{\text{off}}^{(i)} + t_{x,o}^{(i)},\\[6pt]
h_x^{(i)} = h_{x,o}^{(i)},\\[8pt]
w_x^{(i)} = w_{x,o}^{(i)}.
\end{cases}
$$
The cumulative sum above wraps back to $0$ whenever it would exceed the table. We experimented with $f_o$ values of $5$ and $15$.

\item 3D RoPE configuration (vi).
The target commences from $(f_o, f_o, f_o)$ on all three axes. The reference height and width begin at $h_{\mathrm{tar}} + f_o$ and $w_{\mathrm{tar}} + f_o$, respectively, accumulate by $h_{\text{ref}}^{(j)} + f_o$ and $w_{\text{ref}}^{(j)} + f_o$ per reference, and wrap back to $0$ when exceeding the boundary. 
The internal offset is an index quantity fully determined by the reference temporal counts and the gap: it starts from $f_{\text{off}}^{(1)} = 0$ and accumulates as $f_{\text{off}}^{(i)} = \sum\limits_{j=1}^{i-1}\bigl(t_{\text{ref}}^{(j)} + f_o\bigr)$ per reference before any reset, and it resets to $0$ whenever the accumulated offset would reach or exceed the configurable offset $f_o$:
$$
\begin{cases}
t_x^{(i)} = \begin{cases}
t_{x,o}^{(i)} & \text{if } f_{\text{stack}} = \text{False},\\[4pt]
f_{\text{off}}^{(i)} + t_{x,o}^{(i)}, & \text{if } f_{\text{stack}} = \text{True},
\end{cases}\\[12pt]
h_x^{(i)} = h_{x,o}^{(i)} + h_{\mathrm{tar}} + f_o + \displaystyle\sum_{j=1}^{i-1}\bigl(h_{\text{ref}}^{(j)} + f_o\bigr) ,\\[8pt]
w_x^{(i)} = w_{x,o}^{(i)} + w_{\mathrm{tar}} + f_o + \displaystyle\sum_{j=1}^{i-1}\bigl(w_{\text{ref}}^{(j)} + f_o\bigr) .
\end{cases}
$$

\item 3D RoPE configuration (vii).
The reference elements are placed first. Their temporal index resets to $0$ ($t_x^{(i)} = t_{x,o}^{(i)}$), and their height and width indices start at $f_o$, accumulate across references, and wrap back to $f_o$ when out of bounds. Once all references have been processed, the target is placed immediately after the accumulated reference region; its temporal band begins at $f_o$ ($t_x = f_o + t_{x,o}$), and its spatial indices follow the same $f_o$-anchored accumulation:
$$
\begin{cases}
t_x^{(i)} = t_{x,o}^{(i)},\\[4pt]
h_x^{(i)} = h_{x,o}^{(i)} + f_o + \displaystyle\sum_{j=1}^{i-1} h_{\text{ref}}^{(j)} ,\\[8pt]
w_x^{(i)} = w_{x,o}^{(i)} + f_o + \displaystyle\sum_{j=1}^{i-1} w_{\text{ref}}^{(j)} ,
\end{cases}
$$
with the target placed after the reference cluster in the same table. We experimented with $f_o = 1$.

\item 3D RoPE configuration (viii).
The target commences at the origin $(0,0,0)$. 
The references employ a fixed temporal offset $f_o$ alone (without incorporating the target's temporal length), while their spatial stacking follows that of configuration (i):
$$
\begin{cases}
t_x^{(i)} = f_o + t_{x,o}^{(i)},\\[4pt]
h_x^{(i)} = h_{x,o}^{(i)} + h_{\text{tar}} + \displaystyle\sum_{j=1}^{i-1} h_{\text{ref}}^{(j)} ,\\[8pt]
w_x^{(i)} = w_{x,o}^{(i)} + w_{\text{tar}} + \displaystyle\sum_{j=1}^{i-1} w_{\text{ref}}^{(j)} .
\end{cases}
$$
We experimented with $f_o = 5$.

\item 3D RoPE configuration (ix).
We devised an asymmetric two-sided design in which the reference elements and the target element are encoded differently. 
Each reference element independently applies 3D RoPE with all three coordinates starting from $0$, and its self-attention is performed element-wise, ensuring that the references are not assigned any artificial temporal ordering relative to one another. 
On the target side, the entire sequence is re-indexed using the video-style 3D RoPE—the target element occupies temporal indices $[0, t_{\mathrm{tar}})$ and the $i$-th reference element occupies the subsequent $t_{\mathrm{ref}}^{(i)}$ temporal indices—after which the target queries attend to the full sample over this concatenated sequence. 
This configuration was investigated under the reference-inject self-attention mask, alongside the single-use absolute positional encoding described in \cref{para:Configurations of Element Identity Encoding.}.

\item 3D RoPE configuration (x).
The target's temporal band begins at the configurable offset $f_o$ and its spatial indices start at $1$, i.e., $t_x = f_o + t_{x,o}$, $h_x = 1 + h_{x,o}$, $w_x = 1 + w_{x,o}$. 
For the reference elements, let $i = 1,2,\dots,n$ enumerate them. 
The height and width indices commence immediately after the target's extent ($h_{\mathrm{tar}} + 1$, $w_{\mathrm{tar}} + 1$), accumulate across references, and wrap back to $0$ when out of bounds. 
The temporal indexing has two modes controlled by the stacking flag $f_{\text{stack}}$: if $f_{\text{stack}}$ is false, each reference restarts from $0$ ($t_x^{(i)} = t_{x,o}^{(i)}$); if $f_{\text{stack}}$ is true, an internal offset starts from $f_{\text{off}}^{(1)} = 0$ and accumulates as $f_{\text{off}}^{(i)} = \sum_{j=1}^{i-1}\bigl(t_{\text{ref}}^{(j)} + 1\bigr)$ per reference before any reset, resetting to $0$ when it would overlap the target's reserved band $[f_o, f_o + t_{\text{tar}})$, giving $t_x^{(i)} = f_{\text{off}}^{(i)} + t_{x,o}^{(i)}$. 
Formally:
$$
\begin{cases}
t_x^{(i)} = \begin{cases}
t_{x,o}^{(i)} & \text{if } f_{\text{stack}} = \text{False},\\
f_{\text{off}}^{(i)} + t_{x,o}^{(i)}, & \text{if } f_{\text{stack}} = \text{True},
\end{cases}\\[10pt]
h_x^{(i)} = h_{x,o}^{(i)} + h_{\mathrm{tar}} + 1 + \displaystyle\sum_{j=1}^{i-1} h_{\text{ref}}^{(j)} ,\\[8pt]
w_x^{(i)} = w_{x,o}^{(i)} + w_{\mathrm{tar}} + 1 + \displaystyle\sum_{j=1}^{i-1} w_{\text{ref}}^{(j)} .
\end{cases}
$$
We experimented with $f_o = 1$.

\item 3D RoPE configuration (xi).
In addition to the configurations described above, we also considered a natural variant in which the target element starts at the origin $(0,0,0)$, the temporal indices of the reference elements begin at a fixed offset $f_o$ (by default, $f_o$ is set to $-1$), and the spatial indices of the references accumulate from the origin in element order. 
Formally, for a visual token $x^{(i)}$ of the $i$-th reference element, this variant is given by:
$$
\begin{cases}
t_x^{(i)} = f_o + t_{x,o}^{(i)},\\[4pt]
h_x^{(i)} = h_{x,o}^{(i)} + \displaystyle\sum_{j=1}^{i-1} h_{\text{ref}}^{(j)},\\[8pt]
w_x^{(i)} = w_{x,o}^{(i)} + \displaystyle\sum_{j=1}^{i-1} w_{\text{ref}}^{(j)}.
\end{cases}
$$

\item Task-conditioned RoPE for local and global editing.
Inspired by the task-aware RoPE design in UNIC~\cite{ye2025unic}, which learns a task-conditioned positional layout alongside a task-conditioned embedding, we employ a task-conditioned RoPE for local and global editing. 
Each multi-reference sample is assigned a task tag drawn from one of three categories—``local edit", ``global edit", and ``unknown" (tags that could not be determined manually)—which is mapped to a binary task identifier: the global-editing and unknown classes are assigned $\textit{task} = 0$, whereas the local-editing and video-inpainting classes receive a non-zero identifier. 
The RoPE coordinates depend on this identifier, and a task-conditioned learnable embedding is attached to the reference elements and added to their index embeddings described in \cref{para:Element Identity Encoding.}, thereby allowing the model to discriminate between the two RoPE layouts. 
For $\textit{task} = 0$, the reference and target elements are encoded in distinct spatial spaces: the target element occupies the origin of its own spatial band, all references share a common spatial basis positioned at the end of the target's extent, and the temporal index follows the video order, with the target leading and the references continuing the temporal band in element order. 
Formally, for a visual token with local coordinates $(t_{x,o}, h_{x,o}, w_{x,o})$ within element $i$:
$$
\begin{cases}
t_x^{(i)} = t_{\mathrm{tar}} + \displaystyle\sum_{j=1}^{i-1} t_{\text{ref}}^{(j)} + t_{x,o}^{(i)},\\[8pt]
h_x^{(i)} = h_{\mathrm{tar}} + h_{x,o}^{(i)},\\[6pt]
w_x^{(i)} = w_{\mathrm{tar}} + w_{x,o}^{(i)},
\end{cases}
\quad (\textit{task} = 0)
$$
where the summation runs over the reference elements preceding element $i$, and the target element adopts the origin $(0,0,0)$. 
For the local-editing class, all elements are encoded within the same space originating from the common origin, i.e., $t_x^{(i)} = t_{x,o}^{(i)}$, $h_x^{(i)} = h_{x,o}^{(i)}$, $w_x^{(i)} = w_{x,o}^{(i)}$ for the $i$-th reference element and $t_x = t_{x,o}$, $h_x = h_{x,o}$, $w_x = w_{x,o}$ for the target element; this uniform arrangement suits the local-editing regime. 
The task-conditioned absolute positional encoding is a zero-initialized learnable embedding vector associated with each task identifier. It operates under the same mechanism as the absolute positional encoding described in \cref{para:Element Identity Encoding.}, but is added to the reference visual tokens only once, before the tokens enter the DiT blocks.
\end{itemize}

\paragraph{Masking Strategies for Conditional Generation and Editing Tasks.}
\label{para:Masking Strategies for Conditional Generation and Editing Tasks.}
We split the original T2V samples at the centre by frame count and pre-computed the VAE latents offline for the first-half and second-half sub-samples separately. Building upon these existing training samples, in early September 2025, we investigated the addition of masks to the reference elements across samples, either in the image space or in the latent space. In the image-space variant, blocks of a sampled size were masked on the decoded frames, which were then re-encoded into latents by the VAE. In the latent-space variant, the masked latent was obtained directly by $\tilde{z}_{\text{ref}} = z_{\text{ref}} \odot (1 - M)$, with a sampled block size and a sampled mask ratio. In the strategies below, the masked latent sequence serves as the reference element, and the unmasked counterpart serves as the target element the model is trained to reconstruct. The strategies we examined are as follows:
\begin{itemize}
    \item Random patch masking with shared positions across latent frames. A block size is sampled uniformly from $\{4,8,16\}$, and a random grid of blocks is masked once and reused for every latent frame, so that all latent frames share the same masked positions; the masked sequence of the first-half sub-latents is the reference element, and the unmasked first-half sub-latents are the target element.
    \item Random patch masking with per-frame positions. The same block-size distribution is used, but the block grid is re-sampled independently for each latent frame, so that masked positions differ across latent frames; the target element is likewise the unmasked first-half sub-latents.
    \item First-frame-to-video (I2V). Only the first latent frame remains visible as the reference element, while all subsequent latent frames are masked; the unmasked first-half sub-sample is the target element, turning the sample into an I2V task.
    \item Last-frames-to-video (L2V). Only the first three latents of the second half remain visible as the reference element, while all latents before this retained tail (the first three latents of the second half) are masked; the target element is the first half concatenated with the retained tail.
    \item First-last frame-to-video (FLF2V). The first latent frame and the retained trailing tail (the first three latents of the second half) remain visible as the reference element, while all latent frames between them are masked; the target element is the first half concatenated with the retained tail.
    \item Second-half completion (Video Extension). The first half remains visible as the reference element and the second half is masked; the target element is the complete two-half concatenation, training the model to extend the first half into the second.
    \item First-half completion (Backward Continuation). The second half remains visible as the reference element and the first half is masked; the target element is the complete two-half concatenation, training backward continuation.
    \item Moving mask patches. A rectangular patch of random size and position is masked on either all latent frames of the reference element or a random subset of them; the target element is the unmasked first-half sub-sample.
\end{itemize}
Across trials, the mask ratio was drawn uniformly from several ranges, e.g., $\mathcal{U}(0.3, 0.6)$, $\mathcal{U}(0.3, 0.75)$, and $\mathcal{U}(0.3, 0.8)$.

For the image editing samples, we similarly applied latent-space masks to the reference elements: the latent of each reference image was masked with a mask ratio drawn from $\mathcal{U}(0.75, 0.9)$, following the random patch masking strategy described above, while the target element was the edited output latent. This strategy improved editing responsiveness to some extent, and we further experimented with and observed the effects of different mask ratio ranges on editing responsiveness.

\paragraph{Masking Strategies for Editing OPSD.}
\label{para:Masking Strategies for Editing OPSD.}
As described in \cref{subsec: On-policy Self-distillation of Flow Matching}, the modality mimic editing loss introduced in \cref{para:Modality Mimic Editing Loss.} demonstrates that OPSD for flow matching models is also feasible for editing tasks. 
The masking strategies described in \cref{para:Masking Strategies for Conditional Generation and Editing Tasks.} can be generalized to editing OPSD for image models: for an image editing sample, the teacher sample retains all original reference elements fully visible, while the student sample is constructed by randomly masking the reference elements of the teacher sample in the latent space, with the mask ratio sampled from a predefined interval (e.g., $\mathcal{U}(0.3, 0.8)$). 
Both samples share the same noises, timesteps, and caption text embeddings; in addition, the target elements of the two samples share identical original latents and noisy latents. The two samples differ only in the reference elements: fully visible for the teacher sample and masked for the student sample. 
The editing OPSD loss is then computed through the same procedure as the modality mimic editing loss, adopting the KL-divergence or MSE measure, where this editing OPSD sample pair corresponds to $\tilde{z}^1_{\text{v2v,t}}$ and $\tilde{z}^1_{\text{i2i,t}}$ in the notation of \cref{para:Modality Mimic Editing Loss.}, with every element of an image editing sample being a single-frame latent as in the I2I samples.
The modality mimicry generation method variants described in \cref{sec:Other Attempts During the Exploration} can also be generalized to editing tasks in the same manner.

\paragraph{Task Variants for Conditional Video Generation.}
\label{para:Task Variants for Conditional Video Generation.}
In August 2025, we instantiated several task variants for conditional video generation, built on the T2V latent samples and the pre-segmented sub-samples described in ``Masking Strategies for Conditional Generation and Editing Tasks.''. These variants differ in how the reference elements are extracted and what the target element is expected to generate:
\begin{itemize}
    \item Reference-frame-to-video. From the original T2V latent samples, several single latent frames are sampled from the video as static reference elements, and the target element is the entire video. The model is thus trained to generate the video conditioned on static frames.
    \item Reference-clip-to-video. From the original T2V latent samples, several non-overlapping sub-clips are sampled from the first half of the video as dynamic reference elements carrying temporal structure. Again, the target element is the entire video, training the model to condition on dynamic snippets, with editing instructions enumerating the reference shots, e.g., ``Refer to shot 1, shot 2, \dots, generate the next shot''.
    \item Video extension. Building on the pre-segmented sub-samples, the first segment is taken as the reference element and the last segment as the target. The editing instructions enumerate the reference shots by index and ask for the next one, e.g., ``Extend this video''.
    \item Backward continuation. The segment order of the pre-segmented sub-samples is reversed—the later segment serves as the reference and the earlier segment as the target—thereby training a backward continuation capability.
    \item Video reconstruction. The first segment of a pre-segmented sub-sample is duplicated to serve as both the reference and the target, forming a video-reconstruction sample whose captions ask the model to ``Keep this video unchanged''.
\end{itemize}

Finally, we kindly ask for the readers' understanding regarding any possible insufficiently detailed descriptions, descriptive errors, or symbolic errors that may cause confusion. Interested readers are most welcome to share their valuable corrections.